\documentclass{article}

\usepackage{arxiv}

\usepackage[utf8]{inputenc} 
\usepackage[T1]{fontenc}    
\usepackage{hyperref}       
\usepackage{url}            
\usepackage{booktabs}       
\usepackage{amsfonts}       
\usepackage{nicefrac}       
\usepackage{microtype}      
\usepackage{graphicx}
\usepackage{natbib}
\usepackage{doi}
\usepackage[dvipsnames]{xcolor}  
\usepackage{amsmath}

\usepackage{savesym}
\savesymbol{lesssim}
\savesymbol{gtrsim}
\usepackage{amssymb}

\usepackage{longtable,booktabs}
\usepackage{tipa}
\usepackage{bm}
\usepackage{lscape}
\usepackage{pdflscape} 
\usepackage{rotating}
\usepackage[russian,english]{babel}
\usepackage{enumitem}

\ifpdf%
\usepackage{epstopdf}%
\else%
\fi

\title{Deception detection in text and its relation to the cultural dimension of individualism/collectivism}

\author{Katerina Papantoniou\\
	Department of Computer Science\\
	University of Crete\\ 
	Greece\\
	\texttt{kpapantoniou@csd.uoc.gr} 
	\And
	\href{https://orcid.org/0000-0001-8926-4229}{\includegraphics[scale=0.06]{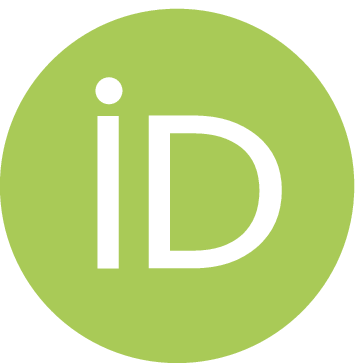}\hspace{1mm}Papagiotis Papadakos} 
	\\
	Institute of Computer Science\\
FORTH\\
Greece\\
	\texttt{papadako@ics.forth.gr} 
		\And
	Theodore Patkos \\
	Institute of Computer Science \\
FORTH\\
Greece\\
	\texttt{patkos@ics.forth.gr} \\
	\And

	\href{https://orcid.org/0000-0002-8937-4118}{\includegraphics[scale=0.06]{images/logo-orcid.eps}\hspace{1mm}Giorgos Flouris}
	\\
	Institute of Computer Science\\
FORTH\\
Greece\\
	\texttt{fgeo@ics.forth.gr} \\

			\And
			Ion Androutsopoulos \\
	Department of Informatics\\
Athens University of Economics and Business\\
Greece\\
	\texttt{ion@aueb.gr} \\

			\And
			
			\href{https://orcid.org/0000-0002-0863-8266}{\includegraphics[scale=0.06]{images/logo-orcid.eps}\hspace{1mm}Dimitris Plexousakis} 
	\\
	Institute of Computer Science\\
FORTH\\
Greece\\
	\texttt{dp@ics.forth.gr} \\

}

\renewcommand{\shorttitle}{\textit{} }

\hypersetup{
pdftitle={A template for the arxiv style},
pdfsubject={q-bio.NC, q-bio.QM},
pdfauthor={David S.~Hippocampus, Elias D.~Striatum},
pdfkeywords={First keyword, Second keyword, More},
}

\begin{document}
\maketitle

\begin{abstract}
Automatic deception detection is a crucial task that has many applications both 
in direct physical and in computer-mediated human communication. Our focus is  on automatic deception detection in text across cultures.  In this context, we view culture through the prism of the individualism/collectivism dimension and we approximate culture by using country as a proxy. Having as a starting point recent conclusions drawn from the social psychology discipline, we explore if differences in the usage of specific linguistic features of deception across cultures can be confirmed and attributed to cultural norms in respect to the individualism/collectivism divide. In addition, we investigate if a universal feature set for cross-cultural text deception detection tasks exists. We evaluate the predictive power of different feature sets and approaches. We create culture/language-aware classifiers by experimenting with a wide range  of  n-gram features  from  several  levels  of  linguistic  analysis, namely  phonology, morphology and syntax, other linguistic cues like word and phoneme counts, pronouns use, etc., and token embeddings. We conducted our experiments over eleven datasets from five languages (English, Dutch, Russian, Spanish and Romanian), from six countries (United States of America, Belgium, India, Russia, Mexico and Romania), and we applied two classification methods, namely logistic regression and fine-tuned BERT models. 
The results showed that the undertaken task is fairly complex and demanding. Furthermore, there are indications that some linguistic cues of deception have cultural origins, and are consistent in the context of diverse domains and dataset settings for the same language. This is more evident for the usage of pronouns and the expression of sentiment in deceptive language.
The results of this work show that the automatic deception detection across cultures and languages cannot be handled in a unified manner, and that such approaches should be augmented with knowledge about cultural differences and the domains of interest.
\end{abstract}


\section{Introduction}
Automated deception detection builds on
years of research in interpersonal psychology, philosophy, sociology, 
communication studies and computational models of
deception detection \citep{bookvrij1,bookvrij2}. Textual data of
any form, such as consumer reviews, news articles, social media comments, political
speeches, witnesses' reports, etc., are currently
in the spotlight of deception research \citep{bookvrij2}. What 
contributed to this vivid interest is the enormous production of
textual data and the advances in computational linguistics.
In many cases, text is either the only available source for extracting deception cues, or the most affordable and less intrusive one, 
compared to  approaches based on Magnetic Resonance Imaging (MRI) \citep{1973Natur.242..190L} and Electrodermal 
Activity (EDA) \citep{Critchley2013}. In this work, we 
exploit natural language processing (NLP) techniques and tools for automated text-based deception detection, and focus on the relevant cultural and language factors.

As many studies suggest, deception is an act that depends on many factors such
as personality \citep{conf/eisic/FornaciariCP13,focus},
age \citep{10.3389/fpsyg.2014.00590}, gender \citep{Tilley2005GenderDI, 
toma2008separating,Fu2008LyingIT}, or culture \citep{1890, Taylor2017CultureMC, Leal2018Culture}. 
All these factors
affect the
way and the means one uses to deceive. The vast majority of works in automatic 
deception
detection take an ``one-size-fits-all'' approach, failing to adapt the
techniques
based on
such factors. Only recently, research efforts that take into account such 
parameters started to appear \citep{perezrosas-mihalcea:2014:P14-2,miha}. 

\textit{Culture} and \textit{language} are tightly interconnected since 
language is a means of expression, embodiment and symbolization of cultural 
reality \citep{kramsch_2011} and as such, differences among cultures are 
reflected in language usage. According to previous studies \citep{liekeh.rotman2012,1890, Taylor2017CultureMC, Leal2018Culture}, this also applies to the expression of \textit{deception} among people belonging to 
different cultures (a detailed analysis related to this point is provided in 
Section~\ref{sec:cl}). The 
examination of the influence of cultural properties in deception detection is 
extremely important since 
differences in social norms may lead to misjudgments and misconceptions and 
consequently can impede fair treatment and justice \citep{jones,1890}. The globalization of criminal activities that employ face-to-face communication, (e.g., when illegally trafficking people across borders) or digital communication (e.g., phishing in e-mail or social media), as well as the increasing 
number of people passing interviews in customs and borders all over the world are only some scenarios that make the incorporation of cultural aspects in the research of deception 
detection a necessity. Since the implicit assumption made about the uniformity of linguistic indicators 
of deception comes
in conflict with prior work from psychological and sociological 
disciplines, our three research goals are:
\begin{enumerate}[label=(\alph*).]
\item Can we verify the prior body of work which states that  linguistic cues of deception are expressed differently, e.g., are milder or stronger, across cultures due to different cultural norms? More specifically, we want to explore how the individualism/collectivism divide defines the usage of specific linguistic cues \citep{1890,Taylor2017CultureMC}. Individualism and collectivism constitute  a well-known division of cultures, and concern the degree in which members of a culture value more individual over 
group goals and vice versa \citep{Triandis1988IndividualismAC}.
Since cultural boundaries are difficult to define precisely when collecting data, we use datasets from different countries assuming that they reflect at an aggregate level the dominant cultural aspects that relate to deception in each country. In other words, we use countries as proxies for cultures, following in that respect \citet{Hofstede:2001}. We also experiment with datasets originating from different text genres (e.g., reviews about hotels and electronics, opinions about controversial topics, transcripts from radio programs etc.).
\item Explore which language indicators and cues are more effective to detect deception given a piece of text, and identify if a \textit{universal feature set}, that we could rely on for detection deception tasks exists. On top of that, we investigate the volatility of cues across different domains by keeping the individualism/collectivism and language factors steady, whenever we have appropriate datasets at our disposal.
\item In conjunction with the previous research goal, we create and evaluate the performance of a wide range of binary classifiers for predicting the truthfulness and deceptiveness of text.
\end{enumerate}

These three research goals have not been addressed before, at least from this 
point of view. Regarding the 
first goal, it is particularly useful to confirm some of the previously reported conclusions 
about deception and culture under the prism of individualism/collectivism with a larger number 
of samples and from populations beyond the closed
environments of university campuses and small communities used by the original studies. For the other two research goals, we aim at providing an efficient methodology for the deception detection task, exploring the boundaries and limitations of the options and tools currently available for different languages.

To answer our first and second research goals, we performed statistical tests on a set of linguistic cues of deception already proposed in bibliography, placing emphasis on those reported to differentiate across the individualism/collectivism divide. We conducted our analysis on datasets originating from six countries,  namely United States of America, Belgium, India, Russia, Romania, and Mexico, which are seen as proxies of cultural features at an aggregate level.  
Regarding the third research goal, the intuition is to explore different approaches for deception detection, 
ranging from methodologies that require minimal linguistics tools for each language (such as word n-grams), to approaches that require deeper feature extraction (e.g., syntactic features obtained via 
language-specific parsers) or language models that  require training on large corpora, either in separation or in combination.
One of our challenges is the difficulty to collect and 
produce massive and representative deception detection datasets.
This problem is amplified by the diversity of languages and cultures, combined with the limited linguistic tools for  under-researched 
languages despite recent advances \citep{conneau-etal-2018-xnli,alyafeai2020survey,Hu2020XTREME, hedderich2020survey}.
To this end, we exploit various widely available related datasets for languages with adequate linguistic tools. We also create a new dataset based on transcriptions from a radio game.
For each language under research, we created classifiers  using a wide range of n-gram features from several levels of linguistic
analysis namely, phonological, morphological and syntactic, along with other linguistic 
cues
of deception and token embeddings. We provide the results of the experiments from logistic regression classifiers, as well as fine-tuned BERT models. Regarding BERT, we
have experimented with settings specific to each particular language, based on the corresponding monolingual models, as well as with a cross-language setting using the multilingual model \citep{devlin-etal-2019-bert}.

In the remainder of this paper, we first present the relevant background
(Section~\ref{sec:bg}),
including both theoretical work and computational work relevant to deception and deception
detection, with
emphasis on the aspects of culture and language. We then proceed with 
the presentation of the datasets that we utilized (Section~\ref{sec:datasets}),
the feature extraction process (Section~\ref{sec:features}), and the statistical evaluation of linguistic cues (Section~\ref{sec:significance}). Subsequently, we present and discuss the classification schemes and the evaluation results, comparing them with related studies (Section~\ref{sec:classification}).
Finally, we conclude and
provide some future directions for this work (Section~\ref{sec:conc}).
 
\section{Background}
\label{sec:bg}
\subsection{Deception in psychology and communication}
Several theories back up the observation that people speak, write and behave
differently when they are lying than when they are telling the truth. Freud was 
the first who observed that the subconscious feelings of people about someone or
something are reflected in how they behave and the word choices they make 
\citep{freud}. The
most influential theory that connects specific linguistic cues with
the truthfulness of a statement is the Undeutsch hypothesis 
\citep{hypo,Undeutsch1989}. This 
hypothesis asserts that 
statements of real-life experiences derived from memory differ
significantly in content and quality from fabricated ones,
since the invention of a fictitious memory requires more cognitive creativity and
control than remembering an actually experienced event.

On this basis, a great volume of research work examines which linguistic
features are more suitable to distinguish a truthful from a deceptive
statement.
These linguistic features can be classified roughly into four categories:
word counts, pronoun use, emotion words and markers of cognitive
complexity. The results for these dimensions have been
contradictory and researchers seem to agree that cues are heavily 
\textit{context-dependent}.
More specifically, the importance of specific linguistic features tends to 
change
based
on many parameters such as the type of text, e.g., dialogue, narrative
\citep{LESLI2}, the
medium of the communication, e.g., face-to-face, computer-mediated
\citep{Zhou2004,doi:10.1080/01638530701739181,Zhou:2008:FLF:1378727.1389972,
Rubin:2010:DDD:1920331.1920377}, 
deception type \citep{stake},
how motivated the deceiver is \citep{stake} etc. There is also a volume of work that examines how the conditions that the experiments were performed in, e.g., sanctioned, unsanctioned, influence the accuracy results and the behaviour of the participants \citep{Feeley1998,sanctioned,10.3389/fpsyg.2015.01965}.

Given the volatility of the results within even the context of a specific
language, the implicit assumption made about the universality of deception cues
can lead to false alarms or misses. Differences in social norms and
etiquette, anxiety and awkwardness that may stem from the language barrier (when speakers do not use their native languages) can
distort judgments. A reasonable argument is that, since the world's languages differ in many ways, the linguistic cues which might have been 
identified as deceptive in one language might not been applicable to 
another. For example, a decrease in first person personal pronoun use is an
indicator of deception in English \citep{Hauch20151stperson}.  What happens though in languages where personal
pronoun use is not always overt such as in Italian, Spanish, Greek and  Romanian (i.e., null subject languages)? In addition, modifiers (i.e., adjectives and adverbs), prepositions, verbs are also commonly examined cues. But not all languages use the same grammatical grammatical categories; for example, Russian and Polish have no articles \citep{newman,Zhou2004,Spence2012}.

All psychology and
communication studies that involve participants from different cultural
groups, asking them to identify truth and fabrications within the same and
different cultural group,
conclude to the same result about the accuracy rate of predictions. More
specifically, as Table~\ref{studies} indicates, the accuracy rate in all
the studies dropped to chance when judgments were made across cultures, whereas 
for within culture judgments it was in line with the rest of the bibliography,
that places accuracy to be typically slightly better than chance
\citep{depaulo1985deceiving}. Indeed, deception detection turns out to be a
very
challenging task for humans. It is indicative that even in studies that involve 
people who have
worked for years at jobs that require training in deception detection, such as
investigators or customs
inspectors, the results are not significantly better \citep{citeulike:3836625}. 
These results are usually attributed to \textit{truth bias}, i.e., the tendency 
of 
humans
to actively believe or passively presume that another person is honest, despite 
even evidence to the contrary 
\citep{depaulo1985deceiving,vrijbook}. The further impairment in accuracy in across culture studies is attributed to
the \textit{norm
violation model}. According to this model, people infer deception whenever
the communicator violates what the receiver anticipates as being normative
behaviour, and this is evident in both verbal and non-verbal
communication \citep{1890}.
\begin{table}[ht!]
  \caption{Social psychology studies on within and across culture deception 
detection.} 
 {\begin{minipage}{\textwidth}
    \begin{tabular}{@{\extracolsep{\fill}}p{3cm}p{6cm}p{6cm}}
    \hline
Reference & Description & Within \& Across culture accuracy (\%) \\
    \hline
\citet{Bond1990} & Jordanian and US undergraduate students were videotaped
while telling lies and truths for the examination of a deceiver's nonverbal
behaviour.&  56 \hspace{1.5cm} 49\\ 
\citet{doi:10.1177/0146167200265010} & American, Jordanian and Indian
students, as well as an illiterate Indian sample were videotaped similarly to
\citet{Bond1990}.& 54 \hspace{1.5cm} 51\\
\citet{Lewis} & Study on deceptive computer-mediated communication between
Spanish and US participants.& 59\hspace{1.5cm} 51\\  \hline
 \end{tabular}
\end{minipage}}
\label{studies}
\end{table}

\subsection{Culture \& language}\label{sec:cl}
The correlation and interrelation between cultural differences and language
usage has been extensively studied in the past.
The most influential theory is the Sapir-Whorf hypothesis that is also known
as the theory of the linguistic relativity  \citep{sapir,wholf}. This theory
suggests that language influences cognition. Thus every human views the world
by his/her own language. Although influential, the strong version of the Sapir-Whorf 
hypothesis has been heavily challenged \citep{deutscherguy2010}. However,
neo-Whorfianism that is a milder strain of the Sapir-Whorf 
hypothesis is now an active research topic 
\citep{Newsapir,doi:10.1002/0470018860.s00567}, stating 
that language influences a speaker's view of the world 
but does not inescapably determine it.

Another view of the relationship between language and culture is 
the notion of \textit{linguaculture} (or \textit{languaculture}). The term 
was introduced by linguistic anthropologists 
Paul Friedrich \citep{friedrich} and Michael Agar \citep{agar}. 
The central idea is that a language is culture bound and much more than a code 
to label objects found in the world \citep{Shaules2019}.

Early studies \citep{haire1966managerial,doi:10.1111/j.1744-6570.1980.tb02165.x} 
support that language and cultural values are correlated 
in the sense that the cross-cultural
interactions that account for similarity in cultural beliefs (geographic
proximity, migration, colonization) also produce linguistic similarity. 
\citet{haire1966managerial} found Belgian-French and Flemish-speakers held
values similar to the countries (France and the Netherlands) with which they shared
language, religion and other aspects of
cultural heritage. In such cases, parallel similarities of language and values
can be seen because they are part of a common cultural heritage transmitted over
several
centuries.
\subsection{Deception \& culture}\label{sec:dc}
The \textit{individualism/collectivism} dipole is one of the
most viable constructs to differentiate cultures and express
the degree to which people in a society are integrated
into groups. In individualism,
ties between individuals are loose and individuals are expected to take care of
only themselves and their
immediate families, whereas in collectivism ties in society are stronger. The individualism/collectivism construct strongly correlates with the distinction between high and low-context communication styles \citep{hall1976}. The low context communication style, that is linked with more individualist cultures, states that messages are more explicit, direct and the transmitter is more open and expresses true intentions. In contrast, in a high context communication messages are more implicit and indirect, so context and word choices are crucial in order for messages to be communicated correctly. The transmitter in this case tries to minimize the content of the verbal message and is reserved in order to maintain social harmony \citep{Wurtz05}. Some 
studies from the discipline of psychology 
examine the behaviour of verbal and non-verbal cues of deception across 
different cultural groups based on these constructs \citep{1890,Taylor2017CultureMC,Leal2018Culture}.

In the discipline of psychology, there is a recent work from \citet{1890,Taylor2017CultureMC} that comparatively 
examines deceptive lexical indicators among diverse cultural groups. More
specifically,
\citet{1890} conducted some preliminary experiments 
over 60 participants from four ethnicities, namely White British,
Arabian, North African and Pakistani.
In \citet{Taylor2017CultureMC} the authors present an extended research work, over 320 
individuals from four ethnic groups,
namely Black African, South Asian, White European and White British, who were
examined for estimating how the
degree of the individualism and collectivism of each culture, influences the
usage of specific linguistic indicators in deceptive and truthful verbal
behaviour. The participants were recruited from community and religious centres across North West England,
and were \textit{self-assigned} to one of the groups. 
The task was to
write one truthful and one deceptive statement about a personal experience, or
an opinion and counter-opinion in English. 
In the study, the collectivist group (Black African and South Asian) decreased 
the \textit{usage of 
pronouns} when lying, and used more first-person and fewer third-person pronouns 
to distance the social group from the deceit. In contrast,
the individualistic group (White European and White British)
used fewer first-person and
more third-person pronouns, to distance themselves
from the deceit.

In these works, Taylor stated the hypothesis that affect in deception is related to cultural differences. This hypothesis was based on previous related work that explored the relation between sentiment and deception across cultures, which is briefly summarized in Table~\ref{sentimentstudies}. 
The results though refute the original hypothesis, showing that
the use of \textit{positive affect} while lying was
consistent among all the cultural groups. More specifically, participants used more positive affect words and fewer words with negative sentiment when they were lying, compared to when they were truthful. 
Based on his findings, emotive language during deception may be a strategy for deceivers to maintain  social harmony.

\begin{table}[ht!]
  \caption{Studies from social psychology discipline on the expression of sentiment in individualism and collectivism.}
 {\begin{minipage}{25pc}
    \begin{tabular}{@{\extracolsep{\fill}}p{3.5cm}p{10cm}}
    \hline
Work & Description   \\
    \hline
\citet{vrij2010pitfalls}  &Deception acts might emerge feelings of guilt, fear, or delight.\\ 
\citet{markus1991}  & Collectivists, in order to avoid conflict and to protect social harmony, may be more engaged to friendly emotions rather than to more unattached emotions like anger.\\
\citet{seiter2002} & Collectivists consider lying more socially acceptable behaviour. To this end, emotions, as proposed by \citep{vrij2010pitfalls} might not emerge in the first place.\\ 
\citet{matsumoto} & Individualism is connected with emotional expression, whereas collectivists are more probable to restraint their emotional expression.\\  \hline
 \end{tabular}
\end{minipage}}
\label{sentimentstudies}
\end{table}

According to the same study, the use of \textit{negations}  is a 
linguistic indicator of deception in the collectivist group, but is unimportant for the
individualist group. 
Negations have been studied a lot with respect to differences among cultures and the emotions they express. \citet{negations} conclude that Asian languages speakers are more likely to use negations than English speakers, due to preference to the indirect style of communication. Moreover, \citet{southafrica} states that for South African languages
the indirect style of communication leads to the usage of negation constructs for the expression of positive meanings. 

\textit{Contextual details}  is a cognition factor
also examined in Taylor's works.
According to the related literature, 
contextual details such as the spatial arrangement of people or objects, occur
naturally 
when people describe existing events from their memory.
The key finding of this
study suggests that this is actually true for the relatively more individualistic
participants, e.g., European. For the collectivist 
groups though, spatial details were less important while experiencing the event at
the first place and subsequently, during recall. 
As a result, individualist cultures tend to provide fewer perceptual details and more social details when they are lying, a trend that changes in collectivist cultures.  Table~\ref{taylorresults}  summarizes all the above findings.   

It is important to mention that the discrepancies on linguistic cues between individualist and collectivist groups were not confirmed for all types of examined lies, namely lies about opinions and experiences. In more details, the analysis showed
that \textit{pronoun use} and \textit{contextual embedding} (e.g., the 
``circumstances'') varied when participants lied about
experiences, but not when
they lied about opinions. By contrast, the affect-related language of the 
participants varied when they lied about
opinions, but not experiences. 
All the above findings indicate that it does not suffice to conceptualize liars as people  motivated ``to not get caught'', since additional factors influence the way they lie, what they do not conceal, what they have to make up, who they want to protect, etc.

\begin{table}[ht!]
  \caption{Summary of differences in language use between truthful and deceptive 
statements across the four cultural groups examined in the work of \citet{1890,Taylor2017CultureMC}. Differences in pronoun usage and perceptual details were confirmed when participants lied about experiences, whereas  affective language differences were confirmed when participants lied about opinions.}
 {\begin{minipage}{25pc}
    \begin{tabular}{@{\extracolsep{\fill}}p{5cm}p{2cm}p{2.3cm}p{2.1cm}p{2cm}}
    \hline
Language indicator&White British (I)&White European (I)&Black African (C)&South 
Asian (C)\\
    \hline
Positive affect &$\upuparrows$ & $\upuparrows$ & 
$\upuparrows$ &   $\upuparrows$ \\
Negations & - & $\uparrow$ & - &$\uparrow$ \\
Perceptual details information&$\downdownarrows$  & $\downarrow$ & $\uparrow$ 
&$\upuparrows$ \\
1st pers. pronouns&$\downdownarrows$  & $\downarrow$ & $\uparrow$ 
&$\upuparrows$ \\
3rd pers. pronouns&$\upuparrows$  & $\uparrow$ & $\downarrow$ 
&$\downdownarrows$ \\
    \hline
    \end{tabular}
  \end{minipage}}
\\($\uparrow$) more in deceptive, ($\downarrow$) more in truthful, (--) 
no difference, (I) individualism, (C) collectivism, ($\uparrow\uparrow$, $\downarrow\downarrow$) suggest larger differences between truthful and deceptive statements.
\label{taylorresults}
\end{table}
\citet{Leal2018Culture} investigate if differences in low and high context culture communication styles can be incorrectly interpreted as cues of deceit in verbal communication. 
Through collective interviews, they studied British interviewees as a representatives of low-context cultures, and Chinese and Arabs as representatives of high-context cultures. The key findings of this work revealed that indeed differences between cultures are more prominent than differences between truth tellers and liars, and this can lead to communication errors. 

\subsection{Automated text-based deception detection}\label{sec:add}
From a computational perspective, the task of deception detection that focuses 
on pursuing linguistic indicators in text is mainly approached  as a classification
task that exploits a wide range of features. In this respect, most research work 
combines psycholinguistic indicators drawn from prior work on deception 
\citep{depaulo1985deceiving,Porter1996,newman} along with n-gram features 
(mainly word n-grams), in order to enhance
predictive performance in a specific context. As already
stated, the psycholinguistic indicators seem to have a strong discriminating 
power in most of the studies, although the quantitative predominance in truthful 
or deceptive texts is extremely sensitive to 
parameters,
such as how motivated the deceiver is, the medium of communication and the 
overall
context.
The number of words that express negative and positive emotions, the number of pronouns,
verbs, and adjectives, and the sentence length  are among the most frequently used features.

\citet{hirschberg} obtain \textit{psycholinguistic} indicators by using the 
lexical categorization program LIWC \citep{liwc} along with other features
to distinguish between deceptive and non-deceptive speech. In the work of 
\citet{GirleaGA16} psycholinguistic deception and
persuasion features were used for the identification of deceptive dialogues
using as a dataset dialogues taken from the party game
Werewolf (also known as 
Mafia)\footnote{\url{https://en.wikipedia.org/wiki/Mafia\_(party\_game)}}. For
the extraction of the psycholinguistic features, the MPQA subjectivity
lexicon\footnote{\url{http://mpqa.cs.pitt.edu/lexicons/subj\_lexicon}} was used,
as well as manually created lists. Various LIWC
psycholinguistic, morphological and n-gram features for tackling the problem of the automatic detection
of deceptive opinion spam\footnote{Any fictitious opinion that has been deliberately
written to sound authentic.} are examined by \citet{Ott:2011:FDO:2002472.2002512,Ott13negativedeceptive}. These 
feature sets 
were tested in a linear Support Vector Machine (SVM) \citep{Cortes95support-vectornetworks}. In these two works \citet{Ott:2011:FDO:2002472.2002512,Ott13negativedeceptive} provide two datasets with deceptive and truthful
opinions, one with positive sentiment reviews \citep{Ott:2011:FDO:2002472.2002512} and one with negative sentiment \citep{Ott13negativedeceptive}. These datasets, either in isolation or combined, have been used as a gold standard in many works. 
\citet{74f0bf0478b44618b8e71218b3a362f3} examined the hypothesis that the number of
named entities is higher in truthful than in deceptive statements,
by comparing the discriminative ability of named entities 
with a lexicon word count approach (LIWC) and a measure of sentence specificity. The results suggest that named entities may be a useful addition to existing approaches.

\citet{Feng:2012:SSD:2390665.2390708} investigated how 
\textit{syntactic} stylometry can help in text deception detection.
The features were obtained from Context Free
Grammar (CFG) parse trees and were tested over four different datasets, spanning from 
product reviews to essays. The results showed improved performance compared to several baselines that were based on shallower lexico-syntactic features. 

\textit{Discourse and pragmatics} have also been used for the task of deception
detection. Rhetorical Structure Theory (RST) and Vector Space Modeling (VSM) are
the two
theoretical components that have been applied by \citet{discourse} in order to 
set apart deceptive and truthful stories. The
authors proposed a two-step approach: in the first step, they analyzed 
rhetorical
structures, discourse constituent parts and their coherence relations, whereas in
the second, they applied a vector space model to cluster the stories by
discourse feature similarity. \citet{Pisarevskaya:2019:AL:3308560.3316604} also 
explored the 
hypothesis that deception in text should be visible from its 
discourse structure. They formulated the task of deception detection 
as a classification task using discourse trees, based on RST. For evaluation reasons, 
they created a dataset containing 2,746 truthful and deceptive complaints about banks in English, where the proposed solution achieved a classification accuracy of 70 per cent.

The motivation of \citet{journals/soco/Hernandez-Castaneda17}
was to build a \textit{domain-independent classifier} using SVM. 
The authors experimented with different feature sets: a continuous semantic 
space model represented by 
\textit{Latent Dirichlet Allocation} (LDA) topics \citep{Blei:2003:LDA:944919.944937}, a 
binary word-space model \citep{wsm}, and dictionary-based 
features in five diverse domains (reviews for books and hotels; opinions about abortion, death penalty and best friend). The results revealed the difficulties of 
building a robust cross-domain classifier. More specifically, the average accuracy of 86 per cent in the 
one-domain setting dropped to a range
of 52 to 64 per cent in a cross-domain setting, where a dataset is kept for testing 
and the rest are used for training. LDA was also used by \citet{Jia2018LDA} along with term frequency and word2vec \citep{Mikolov2013word2vec} for  the feature extraction step in a supervised approach to distinguish between fake and non-fake hotel and restaurant reviews. These different features types were examined both separately and in combination, while three classifiers were trained, namely logistic regression, SVM and multilayer perceptron (MLP) \citep{Rumelhart1987MLP}. The evaluation was performed using the Yelp filter dataset\footnote{\url{http://odds.cs.stonybrook.edu/yelpchi-dataset/}} \citep{Mukherjee2013Yelp} and the experimental results showed that the combinations of LDA with logistic regression and LDA with MLP performed better with 81 per cent accuracy. The work of \citet{Torres2019hotels} focuses on how features may change influenced by the nature of the text in terms of  content and polarity. The proposed method examines three different features types based on a bag of words representation. The first type uses all the words in a vocabulary (after a preprocessing step), the second one selects word features that are uniquely  associated with each class (deceptive, truthful) while the third one further extends the classes to four, also adding the sentiment polarity factor. The dataset of  \citet{Ott:2011:FDO:2002472.2002512,Ott13negativedeceptive}  was used for the evaluation of the six classifiers (i.e., k-NN, logistic regression, SVM, random forest, gradient boosting and MLP) that were employed.

\citet{Fontanarava2017FakeReviews} proposed combining a large number of reviews along with reviewer features for the detection of fake reviews. Some of the features were newly introduced for the task inspired by relevant research in fake news. Features were fed to a random forest classifier which was evaluated on the Yelp filter dataset. The results show that the combined features were beneficiary for the task studied.

Finally, various kinds of \textit{embeddings} (e.g., token, node, character, document etc.) and \textit{deep learning} approaches have been applied to the deception detection task. One of the first works is that of \citet{Ren2017BiLSTM} that employs a Bidirectional Long Short-Term Memory network (BiLSTM) \citep{Graves2013BiLSTM} to learn document level representations. A semi-supervised approach is employed in \citet{Cennet2018SPR2EP} for the detection of spam reviews, by using a combination of doc2vec \citep{Lee2014doc2vec} and node2vec \citep{Grover2016node2vec} embeddings. These embeddings are then fed into a logistic regression classifier to identify opinion spam. 
\citet{ZHANG2018576} proposed a deceptive review identification method that uses recurrent convolutional neural networks \citep{Liang2015RCNN} for opinion spam detection. 
The basic idea is that since truthful reviews have been written by people in the context of the real experience,
while the deceptive ones are not, this contextual information can be exploited by the model. \cite{Aghakhani2018FakeGAN} adopted  Generative  Adversarial  Networks  (GANs) \citep{Goodfellow2014GAN} for  the detection of   deceptive   reviews.

\noindent\\ 
\textbf{Non-English and multi-language research}\\

Without a doubt, the English language engrosses the majority of the research
interest for the task of deception detection, due to the bigger pool of English 
speaking researchers, the interest of
industry for commercial exploitation and the abundance of linguistic resources. 
However, analogous approaches have been 
utilized also in \textit{other languages}. 

In the work of \citet{L14-1001}, the 
task of deception detection from text for the
Dutch language is explored by using an SVM with unigram features. In the 
absence of any related dataset, the authors proceeded with the construction of 
their own dataset.  SVMs have also been used for 
deception detection in opinions written in Spanish with the use of the 
Spanish version of the LIWC \citep{Almela:2012:STD:2388616.2388619}. 

Similarly, in the work of \citet{Tsunomori2015} a dialogue 
corpus for the 
Japanese
language is presented and subsequently a
binary classification based on decision trees over this corpus is performed 
using acoustic/prosodic,
lexical and
subject-dependent features.
The comparison with a
similar English corpus has shown interesting results. More specifically, while
in the
prosodic/acoustic features there were no differences between the two languages, 
in
lexical features the results were greatly different. In English, noise, third
person pronoun, and features indicating the presence of ``Yes'' or ``No'' were 
effective. In Japanese the lexical features used in this research were largely ineffective; and only one lexical feature, the one that indicated the presence of a verb base form, 
proved effective. 

For the 
Chinese language, one of the first studies is that of \citet{4438849} who 
examined the
computer-mediated communication of Chinese players engaged in the Werewolf
game. Having as starting point prior research for English, they ended up with a 
list of
features, (e.g., number of words, number of messages, average sentence
length,
average word length, total number of first-person and third person singular/plural pronouns) and they performed 
statistical
analysis. Results revealed that, consistent with some studies for English
speakers,
the use of third person pronouns increased during deception. In Chinese though, 
there were no significant differences between the proportional use
of first pronouns.

For spam detection in Arabic opinion texts an ensemble approach has been proposed by \citet{Saeed2019Arabic}. A stacking ensemble classifier, that combines a k-means classifier with a rule-based classifier, outperforms the rest of the examined approaches. Both classifiers use content-based features, like n-grams. Given the lack of datasets for fake reviews in Arabic, the authors use for evaluation purposes the translated version of the dataset of  \citet{Ott:2011:FDO:2002472.2002512,Ott13negativedeceptive}. They also use this dataset for the automatic labelling of a large dataset of hotel reviews in Arabic \citep{Elnagar2018arabiccorpus}. A supervised approach is also utilized for deceptive review detection in Persian \citep{Basiri2019persian}. In this work, POS tags, sentiment based features, and metadata (e.g., number of positive/negative feedback, overall product score, review length, etc.) are exploited to construct and compare various classifiers (e.g., naive Bayes, SVMs and decision trees). A dataset with 3,000 deceptive and truthful mobile reviews was gathered using customers reviews published in \url{digikala.com}. The labelling of the latter dataset was performed by using a majority voting on the answers of 11 questions previously designed for spam detection by  human  annotators.

Last but not least, to the best of our knowledge, the only work towards the 
creation 
of \textit{cross-cultural deception detection classifiers} is the work of 
Perez-Rosas et al. \citep{miha,perezrosas-mihalcea:2014:P14-2}. Similar to our work, country is used as a proxy for culture. Using 
crowdsourcing, the 
authors collected four deception datasets.
Two of them are in
English, originating from the United States and from India, one in
Spanish
obtained from speakers from Mexico, 
and one in Romanian from people from Romania. 
Next, they built classifiers for each
language
using unigrams and psycholinguistic (based on LIWC) features. Then, 
they explored the detection of deception using
training data originating from a different culture. To achieve this, they investigated two approaches.
The first one is based on the translation of unigrams features, while the second one is based on the  equivalent LIWC semantic categories. 
The performance, as expected, dropped in comparison with the within-culture classification and was similar for both approaches. The analysis for the psycholinguistic features showed that there are word classes in LIWC that only
appear in some of
the cultures, e.g., classes related to time appear in English texts written by 
Indian people
and in
Spanish texts but not in the US dataset. Lastly, they observed that deceivers in all cultures make use of negation, negative emotions, and
references to
others and that truth tellers use more optimism and friendship words, as
well as
references to themselves.

\section{Datasets}\label{sec:datasets}
We experimented with eleven datasets from six countries, namely United States, 
Belgium, India, Russia, Romania, and Mexico. We provide a detailed description of each dataset below, while Table~\ref{average} provides 
some statistics and summarizes important information for each dataset.
We put much effort
on the collection and the creation of the appropriate datasets. We wanted to 
experiment
with fairly diverse cultures in terms of the degree of  individualism/collectivism,  
having at the same time at
our disposal
basic linguistic tools and resources for the linguistic features
extraction step.
 
\begin{figure}[t]
  \includegraphics[scale=0.6]{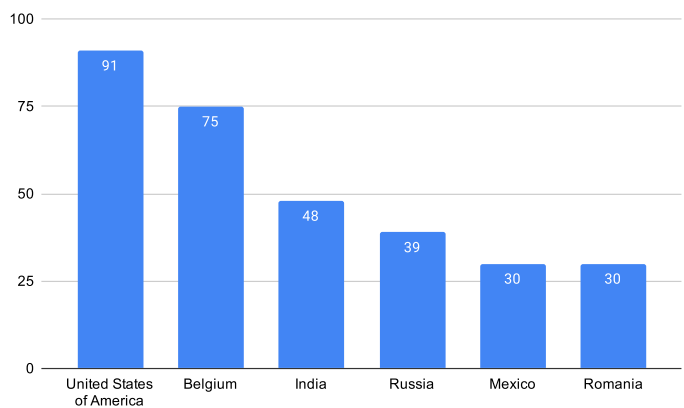}\\
  \caption{Differences between cultures along Hofstede's individualism dimension \tt{(source: 
https://www.hofstede-insights.com/product/compare-countries/)}.}
  \label{hofstede}
\end{figure}

In terms of the quantification of cultural diversity, we based our work 
on Hofstede's long-standing research
on cultural differences \citep{Hofstede:2001}. Hofstede defined a framework that 
distinguishes six 
dimensions (power distance, individualism/collectivism, uncertainty avoidance, masculinity/femininity, long-term/short-term orientation and indulgence/restraint) along which cultures can
be characterized. In his study, as in our work, country has been used as a proxy for culture. For each dimension, Hofstede's provides a score for each culture. Figure~\ref{hofstede} depicts the cultural differences for the six aforementioned countries for the  individualism dimension, which is the focus of our work. The individualism scores vary 
significantly, with 
United States possessing the highest one, and both Mexico and Romania the lowest. We acknowledge that treating entire countries as single points along the individualism/collectivism dimension may be an over-simplification, especially for large countries. In the United States, for example, there is heterogeneity and diversity between regions (e.g., between Deep South and Mountain West) and even in the same region there may be different cultural backgrounds. However, the United States can be considered individualistic at an aggregate level, although there is a measurable variation on the value of this dimension \citep{usinsideculture,Taras2016}.

\begin{table}[ht!]
\caption{Overview of the used datasets. The corresponding columns are: a) dataset, b) culture, c) language, d) type, e) origin, f) collection process, g) number of total, truthful, and deceptive documents, and h) average length of words in truthful and deceptive documents. (T) stands for truthful and (D) for deceptive, while (I) stands for individualist cultures and (C) for collectivist cultures. Truthful documents tend to be longer than deceptive ones, except in Bluff and Russian collections.}
 {\begin{minipage}{25pc}
    \begin{tabular}{@{\extracolsep{\fill}}p{1.8cm}p{0.4cm}lp{1.2cm}lllllll}
    \hline
  Dataset & Cult. & Lang.&Type&Origin&Process&
      \multicolumn{3}{c}{\#Docs}&
      \multicolumn{2}{c}{Average Length}  \\
  &&&&&&  {T+D} & {T} & {D} & {T} & {D} \\
    \hline    
OpSpam & I &en &reviews&written&turkers&1,600 &800&800& 151 0 &146 7 \\
Boulder& I &en&reviews&written&turkers&1,582  &480&1,102& 114 9 &93 9 \\
DeRev& I&en &reviews&written&unsanctioned& 236 &118&118& 128 5 &125  \\
Bluff& I &en&oral stories&transcript&sanctioned&  267    &89&178& 173 4 &207 0  \\
EnglishUS& I&en&essays&written&turkers&600  &300&300&   79 5 & 62 5 \\
CLiPS& I &nl&reviews &written&sanctioned&1,298  &649&649 &143 8&133 7 \\
EnglishIndia& C&en&essays&written&turkers&600  &300&300&   76 1 &66 4 \\
Russian& C&ru&essays&written &sanctioned& 226  &113&113& 160 1 &161 6\\
SpanishMexico& C&es&essays& written  &sanctioned&346  &172&174&  95 6 &64 8 \\
Romanian& C&ro&essays&written &sanctioned& 870  &435&435&  91 5 &68 3\\
NativeEnglish&I&en&multi&multi &multi& 4,285  &1,787&2,498&128 9 &116 5\\
   \hline
    \end{tabular}
  \end{minipage}}
\label{average}
\end{table}

The creation of reliable and realistic ground truth dataset for the deception detection task is considered a difficult
task on its own \citep{Fitzpatrick2012BuildingAD}. In our 
case, the selected corpora have been created using the traditional techniques 
for
obtaining corpora for
deception detection research, namely sanctioned and unsanctioned deception.
Briefly, a sanctioned lie is a lie
to satisfy the experimenter's instructions, e.g., participants are given a 
topic, while an unsanctioned lie is a lie
that is told without any explicit instruction or permission from the researcher, e.g., diary studies and surveys in which participants recall lies already uttered.
Crowdsourcing platforms, e.g., Amazon Mechanical
Turk\footnote{\url{https://www.mturk.com}}, have also been used
for the
production of sanctioned content. In all sanctioned cases, 
a reward (e.g., a small payment) was given as a motivation.
In addition, apart from the already existing datasets in 
the bibliography, we created a new dataset (see 
Section~\ref{datasetbluff}) that concerns spoken text from 
transcripts of a radio game show. 

\subsection{English - Deceptive Opinion Spam (OpSpam)}\label{datasetopspam}
The OpSpam corpus\footnote{\url{http://myleott.com/op-spam.html}} 
\citep{Ott:2011:FDO:2002472.2002512,Ott13negativedeceptive} was created with
the
aim to constitute a benchmark for deceptive opinion spam detection and has been
extensively used as such in subsequent research
efforts. The authors approached the creation of the deceptive and truthful
opinions in two distinct ways.
Firstly, they chose hotel reviews as their domain, due to the
abundance of such opinions on the Web, and focused on 
the 20 most popular hotels in Chicago and positive sentiment reviews. Deceptive
opinions were collected by using Amazon Mechanical Turk. Quality 
was ensured by applying a number of filters, such as using highly rated turkers, located in the Unites States, and allowing only one submission per turker.
Based on these 
restrictions, 400 deceptive
positive sentiment opinions were collected. Secondly, the truthful opinions were collected from
TripAdvisor\footnote{\url{https://www.tripadvisor.com}} for the same 20
hotels as thoroughly described in \cite{Ott:2011:FDO:2002472.2002512}.
Only 5-star reviews were kept to collect reviews with positive sentiment, 
eliminating all non-English reviews, all reviews 
with less than 150 characters, and reviews of
authors with no other reviews. This was  an effort to
eliminate possible spam from the online
data. Then, 400 truthful comments were sampled to create a balanced dataset. 
The same procedure was followed for negative sentiment reviews, by collecting 400 more deceptive opinions with negative sentiment through Amazon Mechanical Turk, and 400 truthful with 1 or 2 star reviews from various online sites. For more details, see \citet{Ott13negativedeceptive}. 

Human performance was assessed with the help of volunteers. They asked three untrained undergraduate university students
to read and judge the truthfulness  and deceptiveness of a subset of the 
acquired datasets. An observation from the results is that human deception 
detection performance is greater for negative (61 per cent) rather than positive 
deceptive opinion spam (57 per cent). But in both cases, automated classifiers 
outperform human performance.

In this work we proceeded with the unification of these two datasets. The corpus contains:
 \begin{itemize}
  \item 400 truthful positive reviews from TripAdvisor \citep{Ott:2011:FDO:2002472.2002512},
  \item 400 deceptive positive reviews from Mechanical Turk \citep{Ott:2011:FDO:2002472.2002512},
  \item 400 truthful negative reviews from Expedia, Hotels.com, Orbitz,
Priceline, TripAdvisor and Yelp \citep{Ott13negativedeceptive},
    \item 400 deceptive negative reviews from Mechanical Turk \citep{Ott13negativedeceptive}.
\end{itemize}

\subsection{English - Boulder Lies and Truth Corpus 
(Boulder)}\label{datasetboulder}
Boulder Lies and Truth
corpus\footnote{\url{https://catalog.ldc.upenn.edu/LDC2014T24}}
\citep{SALVETTI16.1203} was developed at the
University of
Colorado Boulder and contains approximately 1,500 elicited English
reviews of
hotels and electronics for the purpose of studying deception in written
language. Reviews were collected by crowdsourcing with Amazon Mechanical Turk. During data collection, a filter was used to accept US - only submissions \citep{salvetti2014phd}. The
original corpus divides the
reviews in three categories:
\begin{itemize}
\item Truthful: a review about an object known by the writer, reflecting the
real sentiment of the writer towards the object of the review.
\item Opposition: a review about an object known by the writer, reflecting the
opposite sentiment of the writer towards the object of the review (i.e., if the
writers liked the object they were asked to write a negative review, and the opposite if they didn't like the object).
\item Deceptive (i.e., fabricated): a review written about an object unknown
to the writer, either positive or negative in sentiment.
\end{itemize}
This is one of the few available datasets that
distinguish different types of deception (fabrications and lies).
Since the dataset was constructed via turkers, the creators of the dataset took extra care to
minimize the inherent risks, mainly
the tendency of turkers to speed up their work and maximize their economic
benefit through
cheating. More specifically, the creators implemented several methods to
validate the elicited reviews, checking for plagiarism efforts and the intrinsic
quality of the reviews. We unified the two subcategories 
of deception (fabrication and lie), since the focus of this 
work is to investigate deceptive cues without regard to the specific type of 
deception. 

\subsection{English - DeRev}\label{datasetderev}
The DeRev dataset \citep{articlederev} comprises deceptive and truthful 
opinions about books. The opinions have been posted on Amazon.com. 
This is a dataset that provides ``real life'' examples on how 
language is used to express deceptive and genuine opinions, i.e., this is an example of a corpus of unsanctioned deception.
Without a doubt, manually detecting deceptive posts in this case is a very challenging task, since it is 
impossible to find definite proof that a review is truthful or not. 
For that reason a lot of heuristic criteria were employed and only a small subset 
of the collected dataset that had 
high degree of confidence was accepted to be included in the gold standard 
dataset. In more details, only  236  out of the 6,819 reviews that were collected (118 deceptive and 118 truthful) constituted the final dataset.
The starting point for identifying the deceptive and genuine clues that 
define the heuristic criteria was a series of 
articles\footnote{\url{http://www.guardian.co.uk/books/2012/sep/04/sock-puppetry-publish-be-damned}}\textsuperscript{,}\footnote{\url{https://www.moneytalksnews.com/3-tips-for-spotting-fake-product-reviews-\%E2\%80\%93-from-someone-who-wrote-them}}\textsuperscript{,}\footnote{\url{http://www.nytimes.com/2011/08/20/technology/finding-fake-reviews-online.html}}\textsuperscript{,}\footnote{\url{http://www.nytimes.com/2012/08/26/business/book-reviewers-for-hire-meet-a-demand
-for-online-raves.html}}
 with suggestions and advice about how
to unmask a deceptive review in the Web, as
well as specific incidents of fake reviews that have been disclosed. 
Such clues are the absence of information about the 
purchase of the reviewed book, the use of nicknames, reviews that have been 
posted for the same book in a short period of time, and a reference 
to a suspicious book (i.e., a book whose authors have been accused of purchasing reviews, or have admitted that they have done so). The truthfulness of the reviews was identified in a similar manner by reversing the cues. We performed a manual inspection, which confirmed that all of the 113 reviewers of the 236 reviews we used (excluding 8 reviewers whose accounts were no longer valid) had submitted at least one review marked by the platform as having been submitted in the United States. Hence, it is reasonable to assume that the vast majority of the reviewers were US-based.

\subsection{English - Bluff The Listener (Bluff)}\label{datasetbluff}
The ``Wait Wait... Don’t Tell Me!'' is an hour-long weekly radio news panel
game show
produced by Chicago Public Media and National Public Radio
(NPR)\footnote{\url{http://www.npr.org/programs/wait-wait-dont-tell-me}} that 
airs
since 1998. One of the segments of this show is called ``Bluff the Listener'' in which a contestant listens to three
thematically linked news reports from 
three panelists, one of which is truthful and the
rest are fictitious. Most of the stories are humorous and somewhat beyond 
belief, e.g., a class to teach your dog Yiddish. The listener must determine 
the truthful story in
order to win a prize, whereas at the same time the panelist that is picked is
awarded with a point to ensure the motivation for all the participants.
An archive of transcripts of this show is available since 2007 in the official
web page of the show. We used these transcripts and we managed to retrieve and
annotate 178 deceptive and 89 truthful stories. Consequently, we collected 
the participant's replies to
calculate the human success rate. Interestingly, the calculated rate was about 
68 per cent,  which is quite high since in experimental
studies of detecting deception, the accuracy of humans is typically only 
slightly
better than chance, mainly due to {\em truth bias} as previously mentioned.
This might be attributed to the fact that the panelists 
of the show have remained almost the same, and as a result the listeners might have
learned their patterns of deception over time. In addition, we have to stress that the intent of the panelists to deceive is intertwined with their intent to entertain and amuse their audience. Hence, it is interesting to examine if the linguistic cues of deception can be distorted by this double intent, and if they still suffice to discriminate between truth and deception even in this setting.

\subsection{English/Spanish/Romanian - Cross-Cultural Deception}\label{datasetspanish}
To the best of the authors' knowledge, this is the only available multicultural 
dataset constructed
for cross-cultural deception
detection\footnote{\url{http://web.eecs.umich.edu/~mihalcea/downloads.html\#CrossCulturalDeception}}
\citep{perezrosas-mihalcea:2014:P14-2,miha}. It covers four different languages, EnglishUS (English spoken in the US), EnglishIndia (English spoken by Indian 
people), SpanishMexico
(Spanish spoken in Mexico), and Romanian, approximating culture with the country of origin of the dataset. Each dataset
consists of short deceptive and
truthful essays for three topics: opinions on abortion, opinions on death
penalty,
and feelings about a best friend. The two English datasets were collected from
English speakers using Amazon Mechanical Turk with a location restriction to 
ensure 
that the contributors are from the country of interest (United States and 
India). The
Spanish and Romanian datasets were collected from native Spanish and Romanian
speakers using a web interface. The participants for Spanish and Romanian have 
been recruited through contacts of the paper’s authors. For all datasets, the 
participants were asked first to provide their truthful responses, and then 
their deceptive ones. 
In this work, we use all the available individual datasets.
We detected a number of spelling errors and some systematic punctuation problems in both English datasets, with the spelling problems to be more prevalent in the EnglishIndia dataset. To this end, we decided to correct the punctuation errors, e.g., ``kill it.The person'', in a preprocessing step in both datasets. Regarding  the spelling errors, we found no correlation between the errors 
and the type of text (deceptive, truthful), and since the misspelled words were almost evenly distributed among both types of text, we did not proceed to any correction. 

\subsection{Dutch - CLiPS Stylometry Investigation (CLiPS)}\label{datasetclips}
CLiPS Stylometry Investigation (CSI)
corpus\footnote{\url{https://www.clips.uantwerpen.be/bibliography/csi-corpus}}
\citep{L14-1001} is a Dutch corpus containing documents of two genres namely 
essays and reviews. All documents were written by students of Linguistics \& 
Literature at the University of Antwerp\footnote{The city of Antwerp is the capital 
of Antwerp province in the Flemish Region of the Kingdom of Belgium.}, taking 
Dutch proficiency
courses for native speakers, between 2012 and 2014. It is a multi-purpose corpus that serves in many
stylometry tasks such as detection of age, gender,
authorship, personality, sentiment, deception, genre. The place that authors grew up is provided in the metadata. On this basis, it is known that only 11.2 per cent of the participants grew up outside Belgium, with the majority of them (9.7 per cent of the total authors) grown up in the neighbouring country of the Netherlands. 

This corpus, which concerns the review genre, contains 1,298 (649 truthful and 649 deceptive) texts. 
All review texts in the corpus are written by the participants 
as a special assignment for their course. Notice that the participants did not know the purpose of the review task. For the collection of the reviews students were 
asked to write a  convincing  review, positive  or  negative,  about  a
\textit{fictional} product while the truthful reviews reflect the author’s real opinion 
on an existing product. All the reviews were written about products from the 
same five categories: smartphones, musicians, food chains, books, and movies.

\subsection{Russian - Russian Deception Bank (Russian)}\label{datasetrussian}
For the Russian language, we used the corpus of the 
rusProfilingLab\footnote{\url{
http://en.rusprofilinglab.ru/korpus-tekstov/russian-deception-bank}} \citep{
LitvinovaSLL17}. It contains
truthful and deceptive narratives written by the same individuals on the same 
topic (``How I spent yesterday'' etc.). 
To minimize the effect of the \textit{observers paradox}\footnote{According to Labov (1972) ``the aim of linguistic research in the community must be to find out how 
people 
talk when they are not being systematically observed; yet we can only obtain 
these data by systematic observation.''}, researchers did not explain the 
 aim of the research to the
participants. Participants that managed to deceive the trained  
psychologist who evaluated their responses, were rewarded with a cinema ticket voucher. The corpus consists of 113 
deceptive  and 113 truthful texts, written by 113 individuals (46 males and 67 females)  
who were university students and native Russian speakers. Each corpus text 
is accompanied by various metadata such as gender, age, results of a
psychological test, etc.

\subsection{English - Native English (NativeEnglish)}\label{nativeEnglish}
Finally, we combined all the datasets that were created from native English speakers (i.e., OpSpam, Boulder, DeRev, Bluff, and EnglishUS) in one dataset.
The idea is to create one multi-domain dataset, big enough for training, where the input is provided by native speakers.

\section{Features}\label{sec:features}
In this section, we detail the 
feature selection and extraction processes. Furthermore, we explicitly define the features 
that we exploited for pinpointing differences between cultures.

\subsection{Features extraction}
We have experimented with three feature types along with their combinations, 
namely a plethora of linguistic cues (e.g., word counts, sentiment, etc.),
various types of n-grams,
and token embeddings.
Linguistic 
indicators are extracted based on prior work, as already analyzed in 
Sections~\ref{sec:dc} and~\ref{sec:add}. Further, we have evaluated  
various types of n-grams in order to identify the most discriminative ones. 
The use of n-grams is among the earliest and more 
effective approaches for the task of deception detection. 
\citet{Ott:2011:FDO:2002472.2002512} and \citet{conf/eisic/FornaciariCP13} were 
among the first to use word n-grams for deception detection, while 
character n-grams and syntactic n-grams (defined below) have been used by 
\citet{10.1007/978-3-319-18117-2_21} and \citet{Feng:2012:SSD:2390665.2390708} 
respectively. Lastly, due to the absence of a large training 
corpus, we tried to combine feature engineering and statistical models, 
in order to enhance the overall performance and get the best of 
both worlds. This approach is in line with recent
research on deception detection that tries to leverage various types of features 
\citep{Bhatt:2018:CNS:3184558.3191577,krishnamurthy2018deep,10.1145/3349536}.   

\LTcapwidth=\textwidth

\begin{longtable}[!ht]{lccccc}
\caption{The list of used features. Features that are examined in Taylor's work \citep{1890, Taylor2017CultureMC} are marked with an asterisk (*). The  dot ($\bullet$)  marks non-normalized features. Absence of a tick marks the inability to extract this specific feature for this particular language. The N/A indicates that this feature is not applicable for this particular language.} 
\label{table:features} \\
\hline Features &English & Dutch & Russian & Spanish & Romanian \\  \midrule 
\multicolumn{6}{c}{\textbf{1. Linguistic cues}} \\ \midrule
\multicolumn{6}{c}{\it{a. Word counts}}\\\midrule
average word length  $\bullet$	&	\checkmark	&	\checkmark	&	\checkmark	&	\checkmark	&	\checkmark          \\
\#adjectives \& \#adverbs	&	\checkmark	&	\checkmark	&	\checkmark	&	\checkmark	&	\checkmark    \\
\#articles	&	\checkmark	&	\checkmark	&	N/A	&	\checkmark	&	\checkmark       \\
\#boosters	&	\checkmark	&		&		&		&	\\
\#filled pauses	&	\checkmark	&		&		&		&	\\
\#function words	&	\checkmark	&	\checkmark	&	\checkmark	&	\checkmark	&	\checkmark               \\
\#hedges	&	\checkmark	&		&		&		&	\\
\#lemmas    $\bullet$	&	\checkmark	&	\checkmark	&	\checkmark	&	\checkmark	&	\checkmark             \\
\#negations*	&	\checkmark	&	\checkmark	&	\checkmark	&	\checkmark	&	\checkmark    \\
\#prepositions	&	\checkmark	&	\checkmark	&	\checkmark	&	\checkmark	&	\checkmark       \\
\#punctuation marks  $\bullet$	&	\checkmark	&	\checkmark	&	\checkmark	&	\checkmark	&	\checkmark        \\
\#vague words	&	\checkmark	&		&		&		&	\\
\#verbs	&	\checkmark	&	\checkmark	&	\checkmark	&	\checkmark	&	\checkmark      \\ 
\#words      $\bullet$	&	\checkmark	&	\checkmark	&	\checkmark	&	\checkmark	&	\checkmark         \\
\multicolumn{6}{c}{\it{b. Phoneme counts}}\\\midrule
\#fricatives	&	\checkmark	&	\checkmark	&	\checkmark	&	\checkmark	&	\checkmark      \\
\#nasals	&	\checkmark	&	\checkmark	&	\checkmark	&	\checkmark	&	\checkmark             \\
\#plosives	&	\checkmark	&	\checkmark	&	\checkmark	&	\checkmark	&	\checkmark             \\
\multicolumn{6}{c}{\it{c. Pronoun use}}\\ \midrule	&		&		&		&		&	\\
\#total pronouns*	&	\checkmark	&	\checkmark	&	\checkmark	&	\checkmark	&	\checkmark           \\
\#1st person pronouns*	&	\checkmark	&	\checkmark	&	\checkmark	&	\checkmark	&	\checkmark            \\
\#1st person pronouns (singular)	&	\checkmark	&	\checkmark	&	\checkmark	&	\checkmark	&	\checkmark            \\
\#1st person pronouns (plural)	&	\checkmark	&	\checkmark	&	\checkmark	&	\checkmark	&	\checkmark            \\
\#3rd person pronouns*	&	\checkmark	&	\checkmark	&	\checkmark	&	\checkmark	&	\checkmark            \\
\#demonstrative pronouns	&	\checkmark	&	\checkmark	&	\checkmark	&	\checkmark	&	\checkmark            \\
\#indefinite pronouns	&	\checkmark	&	\checkmark	&	\checkmark	&	\checkmark	&	\checkmark            \\
\multicolumn{6}{c}{\it{d. Sentiment}}\\ \midrule 
positive$\mid$negative -SentiWordNet*	&	\checkmark	&		&		&		&	\\
positive$\mid$negative -MPQA*	&	\checkmark	&		&		&		&	\\ 
positive$\mid$negative -FBS*	&	\checkmark	&		&		&		&	\\
sentiment-ANEW*	&	\checkmark	&		&		&		&	\\
positive$\mid$negative -Spanish lexicon*	&		&		&		&	\checkmark	&	\\
positive$\mid$negative -VU-sentiment lexicon*	&		&	\checkmark	&		&		&	\\
positive$\mid$negative -RuSentiLex*	&		&		&	\checkmark	&		&	\\
positive$\mid$negative -RoSentiLex*	&		&		&		&		&	\checkmark  \\
\multicolumn{6}{c}{\it{e. Cognitive complexity}}\\   \midrule
mean sentence length   $\bullet$	&	\checkmark	&	\checkmark	&	\checkmark	&	\checkmark	&	\checkmark            \\
mean preverb length   $\bullet$	&	\checkmark	&		&		&		&	\\
\#conjunction words	&	\checkmark	&	\checkmark	&	\checkmark	&	\checkmark	&	\checkmark\\
\#subordinate clauses	&	\checkmark	&		&		&		&	\\
\multicolumn{6}{c}{\it{f. Relativity}}\\ \midrule
\#exclusion words	&	\checkmark	&		&		&		&	\\
\#modal verbs	&	\checkmark	&		&		&		&	\\
\#motion verbs	&	\checkmark	&	\checkmark	&	\checkmark     &	      \\	
\#spatial words*	&	\checkmark	&	\checkmark	&	\checkmark	&	\checkmark	&	\checkmark    \\
\#verbs in future tense	&	\checkmark	&		&	\checkmark	&	\checkmark	&	\\
\#verbs in past tense	&	\checkmark	&	\checkmark	&	\checkmark	&	\checkmark	&	\checkmark         \\
\#verbs in present tense	&	\checkmark	&	\checkmark	&	\checkmark	&	\checkmark	&	\checkmark     \\ \midrule
\multicolumn{6}{c}{\textbf{2. N-grams \textit{(N=uni,bi,tri,uni+bi,bi+tri,un+bi+tri)}}}\\ \midrule
Phoneme n-grams &    \checkmark      &     \checkmark &      \checkmark    &    \checkmark    &    \checkmark \\
Character n-grams    &    \checkmark      &     \checkmark  &      \checkmark    &    \checkmark    &  \checkmark  \\
Word n-grams   &    \checkmark      &     \checkmark   &      \checkmark    &    \checkmark &    \checkmark       \\
POS n-grams    &    \checkmark      &     \checkmark   &      \checkmark    &    \checkmark      &    \checkmark \\
Syntactic n-grams   &   \checkmark       & \\  \midrule
\multicolumn{6}{c}{\textbf{3. BERT Embeddings}} \\ \midrule
BERT embeddings&    \checkmark   &  
 \checkmark    &     
\checkmark    &    \checkmark  &    \checkmark    \\ 
\multicolumn{6}{c}{\textbf{4. Mixture}} \\ \midrule
N-grams \& linguistic cues&    \checkmark   &   \checkmark    &     
\checkmark    &    \checkmark   &    \checkmark   \\
BERT embeddings \& linguistic cues&    \checkmark   &  
 \checkmark    &     
\checkmark    &    \checkmark &    \checkmark     \\ 
 \bottomrule 
\end{longtable}

\subsection{Linguistic cues}\label{linguistic_cues}
Table~\ref{table:features} presents the complete list of features for each language explored in this work. These features count specific cues in text, aiming to capture characteristics of deceptive and 
truthful language. These indicators have been conceptually divided into six 
categories, namely word counts, phoneme counts, pronoun use, sentiment, cognitive 
complexity, and relativity. 
The absence of a tick in Table~\ref{table:features} marks the 
inability to extract the specific feature, given the available linguistic tools 
and resources for each language while the ``N/A'' marks the non-existence of the particular feature in the specific language, i.e., articles in Russian.

Although we believe that most feature names are self-explanatory, we have to describe further the \#hedges and \#boosters features. Hedges is a term coined by the cognitive linguist George Lakoff \citep{Lakoff1973} to describe words expressing some feeling of doubt or hesitancy (e.g., guess, wonder, reckon etc.). On the contrary, boosters are words that express confidence (e.g., certainly, apparently, apparent, always). Both are believed to correlate either positively or negatively with deception and thus are frequently used in related research work \citep{bachenko-etal-2008-verification}.
Regarding the important feature of pronouns, we consider first person pronouns in singular and plural form, e.g., I vs. we, mine vs. ours, etc., third person pronouns, e.g., they,  indefinite pronouns, e.g., someone, anyone, etc., demonstrative pronouns (e.g., this, that, etc.), and the total number of pronouns.
The linguistic
tools used for the extraction
of the features, e.g., POS taggers, named entity recognition tools, etc., are shown in Table~\ref{table:linguistic_tools}. Some of the features were extracted with handcrafted lists authored or modified by us. Such features include filled pauses (e.g., ah, hmm etc.), motion verbs, hedge words, boosters. etc. 

Table~\ref{table:senti_tools} lists the sentiment analysis tools used for each language. 
We exploited, whenever possible, language-specific sentiment lexicons
used in the bibliography and avoided the simple solution of automatically 
translating sentiment lexicons from American English. Related research 
\citep{arabiclex2016} has
shown that mistranslation (e.g., positive words translated as having neutral 
sentiment in the target language), cultural differences, and different sense 
distributions may lead to errors and may insert noise when translating sentiment lexicons.  
Analogously, we maintained the same practice for the rest of the features.
When this was not
feasible, we proceeded with just the translation of linguistic resources (mostly for the Russian language). For 
the 
\textit{\#spatial words} feature, that counts the number of spatial 
references in 
text, we followed a two-step process. We employed a combination of a named entity 
recognizer (NER) tool (see Table~\ref{table:linguistic_tools}) and spatial
lexicons for each language. The lexicons, principally gathered by us, contain spatially
related words (e.g., under, nearby, etc.) for each language, 
while the named entity recognizer extracts location related entities from the
corpora (e.g., Chicago, etc.). In the case of the English language, the existence of a spatial word in the text was computed 
using a dependency parse, in order to reduce false positives. 
The final value of this feature is the sum of the two values (spatial words and location named entities). For Romanian, we had to train our own classifier based on  Conditional Random Fields (CRFs) \citep{CRF,Finkel2005stanfordCRF} by using as training corpus the RONEC \citep{dumitrescu-avram-2020-introducing} corpus, a free, open-source resource that contains annotated named entities for 5,127 sentences in Romanian.

The values of the features were normalized depending on 
their type. For example, the \emph{\#nasals} feature was normalized by dividing with the total number of characters in the 
document, while the \emph{\#prepositions} with the number of tokens in 
the document. The features \emph{\#words}, \emph{\#lemmas}, \emph{\#punctuaction marks}, \emph{average word length}, \emph{mean sentence length} and \emph{mean preverb length} were left non-normalized.
For each sentiment lexicon, except for ANEW, we computed the score by applying the following formula to 
each document $d$ of $|d|$ tokens and each sentiment $s$ (positive$\mid$negative):
\begin{align*}
\textrm{sentiment\_score}(d, s) =
\frac{\sum_{w \in d} \textrm{sentiment\_strength}(w, s)}
     {|d|}
\end{align*}
The \textit{sentiment\_strength} for SentiWordNet is a value in the interval [0,1] while for the rest sentiment resources the values are either 0 or 1.

For the ANEW lingustic resource \citep{anew} that 
rates words in terms of pleasure (affective 
valence), arousal, and dominance with values from 0 to 10, 
we only considered the normalized valence rating that expresses 
the degree of positivity or negativity of a word. The applying formula in this case is:
\begin{align*}
\textrm{ANEW-sentiment\_score}(d) =
\frac{\sum_{w \in d} \textrm{(ANEW\_valence}(w)-5)}
     {|d|\cdot5}
\end{align*}

Lastly, we included phoneme-related features in our analysis. Our hypothesis was that phonological features, captured by phonemes for text, will be more  discriminative in spoken datasets, since the deceiver will put extra care to sound more truthful to the receiver, even subconsciously. This hypothesis is in line with 
an increasing volume of work that investigates the existence of non-arbitrary 
relations between phonological representation  and  semantics. This   
phenomenon  is known  as
\textit{phonological iconicity} and links a  word's form with the emotion it  
expresses \citep{nastase, Schmidtke2014}. Table~\ref{nasalstudies} summarizes such representative works.

\begin{table}[ht!]
  \caption{Phoneme connection to sentiment in phonological iconicity studies.}
 {\begin{minipage}{25pc}
    \begin{tabular}{@{\extracolsep{\fill}}p{5cm}p{10cm}}
    \hline
Work & Description   \\
    \hline
\citet{fonagy} &This early study on Hungarian poems showed that sonorants (e.g., /l/, /m/) occur more often in tender, but plosives (e.g., /k/, /t/) more often in aggressive poems. \\ 
\citet{taylor_another_1965} & Evidence that pleasantness relations are language specific.  \\ 
\citet{zajonc} & Passages   about   Hell   from Miltons ``Paradise Lost'' were found to contain significantly more  front  vowels  and  hard consonants than passages about  Eden  while  the  latter contained more medium back vowels. \\
\citet{whissell} & The analysis of phonemes in different sources (song lyrics, poetry, word lists, advertisements) show that plosives correlate   with   unpleasant words.  \\ 
\citet{doi:10.1080/01638531003674894}  & Multilingual analysis on poems found that plosive sounds are more likely to express a pleasant mood, whereas a relatively high frequency of nasal sounds indicates an unpleasant mood. Universality is claimed since the authors found the same trend independently of the language.\\
\citet{10.3389/fpsyg.2016.01779} &  This work failed to reproduce the results of 
\citet{doi:10.1080/01638531003674894}. \\ 
\citet{papantoniou-konstantopoulos-2016-unravelling} & The analysis of names of movie fictional characters showed among other findings the connection of the nasals with negative sentiment. 
\\ \hline

\end{tabular}
\end{minipage}}
\label{nasalstudies}
\end{table}

\begin{table}[ht!]
  \caption{Sentiment lexicons used for each language.}

 {\begin{minipage}{25pc}
    \begin{tabular}{@{\extracolsep{\fill}}lp{6cm}p{8cm}}
    \hline
en &ANEW \citep{anew}& Normative emotional ratings for 3,188 words. It provides 
values in respect to pleasure, arousal and dominance of each term.\\ 
&FBS\footnote{\url{http://www.cs.uic.edu/~liub/FBS/opinion-lexicon-English.rar
}} \citep{Hu:2004:MSC:1014052.1014073} &  6,786 words (2,006 positive and 4,783
negative).\\
&MPQA\footnote{\url{http://mpqa.cs.pitt.edu/\#subj_lexicon}} 
\citep{Riloff:2003:LEP:1119355.1119369,Wilson:2005:RCP:1220575.1220619}
& Each word annotated for intensity (strong, weak). 6,885 words (2,718 positive 
and
4,912 negative).\\ 
&SentiWordNet\footnote{\url{https://sentiwordnet.isti.cnr.it}}
\citep{BACCIANELLA10.769}    & All   
WordNet synsets automatically    annotated    for    degrees    of   
positivity, negativity,    and    neutrality/objectiveness.\\ 
nl&VU-sentiment-lexicon\footnote{\url{
https://github.com/opener-project/VU-sentiment-lexicon}}\citep{maks-etal-2014-generating} &
9,237 words (3,314 positive and 5,923 negative).\\
ru&RuSentiLex\footnote{\url{http://www.labinform.ru/pub/rusentilex/index.htm}
}
\citep{LOUKACHEVITCH16.285}  & Lexicon generated through semi-automatic techniques, which contains 16,057 words (10,227 negative, 3,770 positive, 1,747 neutral and 291 either positive or negative based on context).  \\  
es& Spanish Sentiment Lexicon \citep{PEREZROSAS12.1081.L12-1645}&It provides
two polarity lexicons: a. an automatically generated with 2,496 concepts and b. a
semi-automatically generated with 1,347 concepts. We employed the semi-automatically generated lexicon since it is the one with largest reported accuracy of approximately 90 per cent.\\ 

ro&RoSentiLex& We translated the MPQA lexicon by using a bilingual Romanian-English dictionary \citep{11372/LRT-493}.  \\   \hline
    \end{tabular}
  \end{minipage}}
  \label{table:senti_tools}
\end{table}

\subsection{N-grams}\label{subsec:phomen}
We have evaluated several variations of n-grams from various levels of 
linguistic analysis to encode linguistic information. 
Given the diversity of the datasets,
we used different types of n-grams 
to identify those that are more  effective in discriminating   
deceptive and truthful content.  
For each n-gram type and for each dataset we extracted  unigrams, bigrams, trigrams, unigrams+bigrams, bigrams+trigrams, and
unigrams+bigrams+trigrams.
Some examples are  shown in
Table~\ref{ngrams}.

\begin{table}[ht!]
\caption{Linguistic tools used on each language for the extraction of 
features.}
{\begin{minipage}{25pc}
    \begin{tabular}{@{\extracolsep{\fill}}llllll}
    \hline
 Tool & English &Dutch& Russian & Spanish & Romanian\\ \hline
Phonemes	&	Espeak-ng	&	Espeak-ng	&	Espeak-ng	&	Espeak-ng	&	Espeak-ng	\\
Lemmatizer	&	Stanford  CoreNLP	&	-	&	OpenNLP	&	-	&	  - 	\\
Stemmer	&	Stanford CoreNLP	&	snowball	&	snowball	&	Stanford CoreNLP	&	snowball 	\\
POS Tagger	&	Stanford CoreNLP	&	OpenNLP	&	TreeTagger	&	Stanford CoreNLP	&	TreeTagger	\\
NER	&	Stanford CoreNLP	&	OpenNLP	&	DBPedia Spotlight	&	Stanford CoreNLP	&	custom	\\
Syntactic Parser	&	Stanford CoreNLP	&	-	&	-	&	-	&	 -	\\ \hline
    \end{tabular}
  \end{minipage}}
\label{table:linguistic_tools}
 \end{table}

\begin{itemize}
\item \textit{Phoneme n-grams:} These features were extracted 
from the phonetic representation of texts derived 
by applying the spelling-to-phoneme module of the espeak-ng speech synthesizer 
(see Table~\ref{table:linguistic_tools}). We examined  phoneme n-grams at the level of words.\\
\item \textit{Character n-grams:} 
Consecutive characters that can also belong to different words.\\
 \item \textit{Word n-grams:} 
We examined versions with and without stemming
and stopword removal. \\
 \item \textit{POS n-grams:} POS n-grams are contiguous part-of-speech tag sequences such 
as adjective-noun-verb, noun-verb-adverb, and so on, that provide shallow 
grammatical information.
We extracted POS n-grams 
using the appropriate  POS-tagger for 
each language (see Table~\ref{table:linguistic_tools}).\\
 \item \textit{Syntactic n-grams:} syntactic n-grams (sn-grams) are constructed 
by following all the possible paths in dependency trees and keeping the labels of the dependencies (arcs) along the paths.
We used Stanford's 
CoreNLP syntactic parser for the construction of dependency trees
for the English datasets (see Table~\ref{table:linguistic_tools}).
\end{itemize}

\begin{table}[!ht!]
  \caption{Examples of n-gram features.}
{\begin{minipage}{25pc}
\begin{tabular}{@{\extracolsep{\fill}}llllll}
\hline
&Phoneme&Character&Word&POS&Syntactic \\ \hline
unigram&\textipa{l" Ev@l} [level]&h, o, t, e, l& abortion, new & EX, IN, NNP, NNPS & ccomp, xcomp \\
bigram& \textipa{nj" u: l" Ev@l} [new level]&ho, ot, te, el & new hotel & VBN DT, VBG RP &  root-nsubj, root-aux\\
trigram& \textipa{nj" u: l" Ev@l at} [new level at]& hot, ote, tel & a business trip & NN NN VBZ& aux-advmod-root\\ \hline
    \end{tabular}
  \end{minipage}}
  \label{ngrams}
\end{table}

\subsection{BERT embeddings}\label{embeddings}
Regarding token embeddings, we used the contextualized embeddings from the BERT \citep{devlin-etal-2019-bert} model. BERT, which stands for Bidirectional Encoder Representations from
Transformers, is a language model based on a stack of Transformer encoder layers pretrained on a large unlabelled cross-domain corpus using masked language modeling and next-sentence prediction objectives. Since its introduction, BERT has achieved state-of-the-art results in many NLP tasks. In most cases, the best results are obtained by adding a shallow task-specific layer (e.g., a linear classifier) on top of a pretrained BERT model, and \textit{fine-tuning} (further training) the pretrained BERT model jointly with the task-specific layer on a labelled task-specific dataset. In effect, each encoder layer of BERT builds token embeddings (dense vectors, each representing a particular token of the input text). The token embeddings of each encoder layer are revised by the next stacked encoder layer. A special classification embedding ([CLS]) is also included in the output of each layer, to represent the entire input text. In classification tasks, typically the [CLS] embedding of the top-most encoder layer is passed on to the task-specific classifier, which in our case decides if the input text is deceptive or not. We explore this approach in Section 6.2. We note that BERT uses a WordPiece tokenizer\footnote{Consult also https://huggingface.co/transformers/master/tokenizer\_summary.html} \citep{wordpiece}, which segments the input text in tokens corresponding to character sequences (possibly entire words, but also subwords or even single characters) that are frequent in the large corpus BERT is pretrained on. We also note that BERT's token embeddings are context-aware, i.e., different occurrences of the same token receive different embeddings when surrounded by different contexts.
In Table~\ref{bert} we provide details about the used BERT models. We exploit pretrained models on each language, as well as the multilingual BERT model, which is pretrained over Wikipedia in 104 languages. 

\begin{table}[ht!]
 \caption{BERT pretrained models used for each language.}
 {\begin{minipage}{25pc}
    \begin{tabular}{@{\extracolsep{\fill}}lp{2.5cm}p{9cm}}
    \hline
Lang.&Name&Description\\ \hline
en & bert-base-uncased\footnote{\label{googlebert}\url{https://github.com/google-research/bert}} \citep{devlin-etal-2019-bert}&12-layer, 768-hidden, 12-heads, 110M parameters, 30K wordpieces\\
nl &BERTje cased\footnote{\url{https://github.com/wietsedv/bertje}} \citep{vries2019bertje}&12-layer, 768-hidden, 12-heads, 110M parameters, 30K wordpieces\\
ru &RuBERT cased\footnote{\url{http://docs.deeppavlov.ai/en/master/features/models/bert.html}} \citep{DBLP:journals/corr/abs-1905-07213}&12-layer, 768-hidden, 12-heads, 180M parameters, 120K wordpieces\\
es &BETO uncased\footnote{\url{https://github.com/dccuchile/beto}} \citep{CaneteCFP2020}&12-layer, 1024-hidden, 16-heads, 110M parameters, 30K wordpieces\\
ro &bert-base-romanian-cased\footnote{\url{https://github.com/dumitrescustefan/Romanian-Transformers}} \citep{roBERT}&12-layer, 768-hidden, 12-heads, 1M parameters, 50K wordpieces\\
multi & bert-base-multilingual-cased&12-layer, 768-hidden, 12-heads, 110M
parameters, 110K wordpieces\\      \hline
    \end{tabular}
  \end{minipage}}
  \label{bert}
\end{table}

\section{Statistical evaluation of linguistic cues}
\label{sec:significance}
In this section we conduct a statistical analysis of the linguistic cues (see Section~\ref{linguistic_cues}) per dataset. In more details, we conduct a Mann-Whitney U test to identify the statistically significant linguistic
features of each dataset (the NativeEnglish dataset is the unified dataset of all native English speakers datasets). Afterwards, we apply  a multiple logistic regression (MLR) analysis over the statistically important features of each dataset. This test shows the distinguishing strength of the important linguistic features. We discuss the results for each dataset/culture, and try to provide some cross-cultural observations.

\subsection{Statistical significance analysis}
Since we cannot make any assumption about the distribution of the feature values in each dataset, 
we performed the non-parametric Mann–Whitney U test (two-tailed) with
a 99 per cent confidence interval and 
$\alpha  =  0.01$.
The null hypothesis (H0) to be refuted is that there is no statistically significant difference between the mean rank
of a feature for texts belonging to the deceptive class and the mean rank of the same feature for texts belonging to the truthful class.
The results are available in the Appendix Tables \ref{u_test_us} and \ref{u_test_culture}.
Below we summarize the main observations.
\begin{enumerate}
\item No statistically significant features were found in the Russian collection and as a result we ignore this dataset in the rest of this analysis. This is probably due to the inappropriateness of the selected features and/or the shortage of language resources for the Russian language, or even because of the intrinsic properties and peculiarities of the dataset itself. This suggests that we cannot come to any conclusion about how the linguistic features are used in this dataset and compare it with the rest.
\item Statistically significant differences were found in most of the datasets for the features: {\em \#lemmas}, {\em \#words}, and {\em \#punctuation}. 
In more details:
\begin{itemize}
\item The importance of {\em \#lemmas} is observed in most of the  datasets. A large number of lemmas seems to be a signal for truthful texts in most of the examined datasets, with the exception of the DevRev and Bluff datasets, where a large number of lemmas is a signal for deceptive texts. These two datasets are quite distinct from the rest, since the former is an example of unsanctioned deception, while the latter concerns transcriptions of spoken data with notably stylistic elements like humour and paralogism.  Although, we cannot characterize it as a universal feature, since it is not observed in the Russian dataset, it is a {\em language-agnostic} cue that seems to be employed across most cultures.
\item The same observations hold also for the feature {\em \#words}, with the exception that it is not statistically significant for the OpSpam dataset.  
\item Regarding the {\em \#punctuation} feature, it is rather important for all datasets except for Bluff and DeRev. Since Bluff is a dataset created from transcripts, the transcription process  might shadow the intonation and emotional status of the original agent with the idiosyncrasies of the transcriber/s, e.g., there are almost zero exclamations. Furthermore, the use of punctuation,
except in DeRev and Bluff, is an indication of  truthful text.
\end{itemize}
\item An observation of possibly cultural origin is the fact that {\em sentiment related features}, positive or negative, are notably important for the individualist cultures (US and Dutch). The expression of more positive sentiment vocabulary is linked with the deceptive texts, while negative sentiment is linked to truthful text, except 
in the EnglishUS case, where the negative sentiment is related to the deceitful texts.
For the collectivistic cultures that are more engaged in the high context communication style, sentiment related features are not distinguishing. As explained earlier, the effort to restrain sentiment and keep generally friendly feelings towards the others in order to protect social harmony, might be responsible for this difference. Our findings contradict Taylor's results and are in agreement with his original hypothesis and related studies like \citet{seiter2002} (see Section \ref{sec:dc}).
\item Another important finding of our experiments is that in almost all datasets the formulation of sentences in {\em past tense} is correlated with truth, while in {\em present tense} with deception, independently of the individualistic score of the corresponding culture. This can be attributed to the process of recalling information in the case of truthful reviews or opinions. In the case of deception, present tense might be used due to preference to simpler forms, 
since the deceiver is in an already overloaded mental state. In the US datasets the only exceptions are the Bluff and the OpSpam datasets, where we observe the opposite. However, in the OpSpam dataset these two features are not statistically significant.
\item Furthermore, the {\em \#modal verbs} is important in the US datasets. Specifically, an increased usage of modal verbs usually denotes a deceptive text.
\item Another cross-cultural observation correlated with the degree of individualism is the {\em \#spatial words} feature. Specifically, for the datasets where this feature is important, we observe a difference in the frequency of spatial details for the deceptive texts in the collectivist datasets and the truthful texts in the individualistic ones. In detail, more spatial features are linked with deception for the Romanian and SpanishMexico datasets, while their frequency is balanced in the case of Dutch, and diverges to truthful text for the NativeEnglish dataset. These observations are in agreement with Taylor  (see Table \ref{taylorresults}). On top of that, discrepancies in the quantity of spatial details have also been 
found in different modalities \citep{1385277}.  More specifically, deceivers had  significantly fewer spatial details than truth-tellers in audio but more in 
text. This signifies how sensitive this linguistic cue is not only across cultures but also when other parameters  such  as context or modality vary.
\item Regarding the {\em \#pronouns}, our results show mixed indications about their usage that do not fully agree with Taylor. 
Notice though, 
that we had only limited tool functionality for pronoun extraction (i.e., no tools for Dutch and SpanishMexico).
As a result, we created our own lists for English and used translations for the other languages.
Generally, pronouns in various forms seem to be important in most datasets. {\em Third person pronouns} are correlated with deceptive texts mainly in EnglishUS and less in the  Romanian and EnglishIndia datasets, all of which belong to the same cross-cultural opinion deception detection dataset, and to truthful ones in the Boulder dataset (with a rather small difference though). This is in partial agreement with Taylor's results, where third person pronouns are linked with deception in collectivist languages. 
Regarding {\em first person pronouns}, the observations show mixed results. They are linked with both truthful and deceptive text, in the latter case though only for individualistic datasets (i.e., Bluff and OpSpam). Exploring the use of singular and plural forms sheds a bit more light, since the plural form is linked with truthful text in both collectivistic and individualistic cultures, except in Dutch where the plural form slightly prevails for deceptive. Finally, {\em indefinite} and {\em demonstrative} pronouns are rarely important.
\item The {\em \#nasal} feature, that counts the occurrences of /\textipa{m}/, /\textipa{n}/ and in some languages /\textipa{N}/ in texts, is rather important for the highly collective SpanishMexico and Romanian datasets. It prevails in truthful texts while we observe the opposite in the individualistic NativeEnglish. 
This is an interesting observation that enriches the relevant research around nasals. Generally, there are various studies (see Table \ref{nasalstudies}) that claim a relation between the occurrence of consonants and the emotion of words based on the physiology of articulation for various languages. Most of the studies link nasals with sadness and plosives with happiness, although there are other studies contradicting these results (see Table \ref{nasalstudies}). Furthermore, nasals have been connected with different semantic classes like iconic mappings, size and affect as shown by \citet{Schmidtke2014}. Finally, notice that plosives are not statistically significant in our results. We believe that this is a direction that needs further research with larger datasets and more languages.
\item Finally,  the {\em \#filled pauses} feature, that was incorporated to showcase differences between written and oral deception cues, does not provide any remarkable insight.
\end{enumerate}

A collateral resulting observation is that most of the distinguishing features do not require complex extraction processes but only surface processing like counts on the token level.

\noindent\\ 
\textbf{EnglishUS and EnglishIndia datasets comparison}\\
The EnglishUS and EnglishIndia datasets are ideal candidates to 
examine individualism-based discrepancies in linguistic deception cues by 
keeping the language factor the same.
These  two datasets are part of the Cross-Cultural Deception dataset (see 
Section~\ref{datasetspanish}) and were created using the  same methodology. 
Both contain opinions on the same topics (death penalty, abortion, best friend), in the same 
language, and come from two cultures with a large difference in terms of the individualism index score (91 vs 48). 
Initially, to explore differences in the authors writing competence we computed the Flesch reading-ease score\footnote{This reading-ease score is based on the average length of the words and sentences.} \citep{Kincaid1975DerivationON} on both datasets. The scores are similar (63.0 for the EnglishIndia dataset and 63.6 for EnglishUS dataset) and correspond to the same grade level.
Notice though, that based on  Tables \ref{u_test_us} and \ref{u_test_culture} in the Appendix, the native speakers use larger sentences and more subordinate clauses. A possible explanations is that since Indians are not native speakers of English, they might lack in language expressivity and use English similarly whether they are telling the truth or lying. 

A crucial observation though, is the limited number of statistically important features in the case of the EnglishIndia (only 3) compared to the EnglishUS (15). 
Furthermore, the pronoun usage differs a lot between the two datasets. In more details, the individualist group employs more {\em \#1st person pronouns} in truthful text, while in the case of the EnglishIndia first person pronouns are not important. In the case of {\em \#3rd person pronouns} both datasets use the same amount of pronouns with a similar behaviour.
As already mentioned, this might be a difference of cultural origin, since individualist group deceivers try to distance themselves from the deceit, while in the collectivist group deceivers aim to distance their group from the deceit. 
Finally, we notice again the importance of the sentiment cues for the native English speakers and their insignificance in the  EnglishIndia dataset, which correlates to our previous observations. For the remaining features, it is risky to make any concluding statements in relation to cultural discrepancies. 

\subsection{Multiple Logistic Regression (MLR) analysis}

To further examine the discriminative ability of the linguistic features and explore their relationship, we conducted a \textit{multiple logistic regression} (MLR) analysis on the resulting significant features from the  Mann–Whitney U test. The null hypothesis is that there is no relationship between the features and the probability of a text to be deceptive. In other words, all the coefficients of the features are considered equal to zero for the dependent variable.

Since MLR presupposes uncorrelated independent variables, we keep only the most significant feature for any set of correlated and dependent features for each dataset and manually filter out the rest. For example, we keep only the single most important positive or negative sentiment feature per dataset (e.g., in English where we use various lexicons). Also in the case of features that are compositions
of more refined features, we keep the most refined ones when all of them are important, e.g.,   
we keep the feature pair {\em\#first person pronouns (singular)} and {\em\#first person pronouns (plural)} instead of the more general {\em\#first person pronouns}. Overall, we cannot guarantee that there is no correlation between the features.

In Tables~\ref{MLR_english} and~\ref{table:MLR_cross} we present the results of the MLR analysis, reporting the features with \textit{p}-value \textless 0.1, for the native English and the cross language cases respectively.
For each feature in the table we report the corresponding coefficient, the standard error, the z-statistic (Wald z-statistic), and the \textit{p}-value. 
Higher coefficient values increase the odds of having a deceptive text in the presence of this specific feature,
while lower values increase the odds of having a truthful text. The Wald (or z-value) is the regression coefficient divided by its standard error. The larger magnitude (i.e., either too positive or too negative)  indicates that the corresponding regression coefficient is not 0 and the corresponding feature matters. 
Generally, there are no features participating in all functions both in the context of the native English datasets and across different languages and cultures, an indication of how distinct the feature sets are both within and across cultures. 
Among different languages is difficult to conclude how the characteristics of each language (e.g., pronoun-drop languages) and/or the different extraction processes (e.g., sentiment lexicons) affect the analysis.
A more thorough analysis is safer to be performed in the context of the same language though.
Below, we report some observations from this analysis. 

\noindent\\ 
\textbf{Native English dataset observations}\\
For the native English language datasets shown in Table~\ref{MLR_english}, 
we observe  high coefficients for the various types of pronouns (especially of the first person ones).
Although there is no clear indication about their direction, most of the times
they are associated with truthful text. The only exceptions are the {\em\#1st person pronouns (singular)} in the case of OpSpam, the {\em\#demonstrative pronouns}
in the case of Boulder, and the {\em\#3rd person pronouns} in the case of EnglishUS. 
Additionally, we can observe the importance of sentiment, as already noted in the statistical analysis, especially of positive sentiment as captured by MPQA,
which highly discriminates deceptive texts in the OpSpam and Boulder datasets.
Finally, the {\em\#punctuation marks} feature is correlated with truthful text in many datasets, although with a lower coefficient.

Notice that in the results there are a number of features with high coefficients that appear to be strong only in one dataset, e.g., the \#boosters, \#nasals and \#hedges features that are extremely distinguishing  only in the OpSpam collection.
This observation indicates differences and variations among various datasets/domains, in accordance with previous considerations in the literature on how some features can capture the idiosyncrasies of a whole domain or only of a particular use case. In the case of the OpSpam, such features might be representative  of the online reviews domain or might reflect how mechanical turkers fabricate sanctioned lies \citep{Mukherjee2013FakeRD}. 

Regarding the \textit{\#spatial details} feature, for which we made some interesting observations in the previous statistical analysis, we observe that they are important for discriminating truthful text only in OpSpam. The observation that fake reviews in OpSpam include less spatial language has already been pointed out by \citet{Ott:2011:FDO:2002472.2002512, Ott13negativedeceptive} for reviews with both positive and negative sentiment. This is not the case  in situations where owners bribe customers in return for positive reviews or 
when owners ask their employees to write reviews \citep{P14-1147}.

Finally, regarding the {\em\#lemmas} and {\em\#words}, they were not found to be important in this analysis.
The same holds for the used tenses in most datasets with no clear direction.

{\fontsize{10}{10}\selectfont
\begin{longtable}{lrrrrr}
\caption{Multiple logistic regression analysis on linguistic features for each US dataset. SE stands for Standard Error. We show in bold \textit{p}-values \textless 0.01. Positive or negative estimate values indicate features associated with deceptive or truthful text, respectively.}\label{MLR_english} \\
\hline

\textbf{OpSpam}&Estimate&SE&Wald&\textit{p}-value \\ \midrule 
\#1st person pronouns (singular)&24 411&4 116&5 931&\bm{$\sim 0$}\\ 
\#boosters&19 456&9 293&2 094&0 036\\ 
\#hedges&12 461&4 76&2 618&\textbf{0 009}\\ 
positive sentiment-MPQA&11 056&1 684&6 564&\bm{$\sim 0$}\\ 
\#nasals&9 773&5 747&1 7&0 089\\ 
\#fricatives&8 432&4 611&1 829&0 067\\ 
\#verbs&8 324&2 814&2 958&\textbf{0 003}\\ 
mean preverb length&0 192&0 034&5 628&\bm{$\sim 0$}\\ 
mean sentence length&0 121&0 018&6 546&\bm{$\sim 0$}\\ 
\#punctuation marks&-1 155&0 109&-10 594&\bm{$\sim 0$}\\ 
\#spatial words&-2 138&0 794&-2 691&\textbf{0 007}\\ 
intercept&-4 749&1 189&-3 995&\bm{$\sim 0$}\\  
\midrule

\textbf{Boulder}&Estimate&SE&Wald&\textit{p}-value  \\ \midrule 
\#modal verbs&8 84&2 984&2 963&\textbf{0 003}\\ 
positive sentiment-MPQA&5 088&1 366&3 723&\bm{$\sim 0$}\\ 
\#lemmas&-0 014&0 009&-1 665&0 096\\ 
\#punctuation marks&-0 16&0 081&-1 979&0 048\\ 
\#articles&-6 273&1 946&-3 224&\textbf{0 001}\\ 
\midrule

\textbf{DeRev}&Estimate&SE&Wald&\textit{p}-value  \\ \midrule 
\#lemmas&0 05&0 021&2 433&0 015\\ 
\#words&-0 025&0 01&-2 453&0 014\\ 
\#pronouns&-13 319&4 077&-3 267&\textbf{0 001}\\ 
\#articles&-18 203&5 177&-3 516&\bm{$\sim 0$}\\ 
\midrule

\textbf{EnglishUS}&Estimate&SE&Wald&\textit{p}-value \\ \midrule 
\#3rd person pronouns&13 5&3 421&3 946&\bm{$\sim 0$}\\ 
\#verbs&6 532&3 38&1 933&0 053\\ 
intercept&2 838&1 686&1 683&0 092\\ 
\#words&-0 026&0 012&-2 08&0 038\\ 
\#punctuation marks&-0 462&0 182&-2 533&0 011\\ 
\#1st person pronouns (plural)&-8 493&4 207&-2 019&0 043\\ 
\#prepositions&-9 008&3 276&-2 75&\textbf{0 006}\\ 
\#indefinite pronouns&-10 642&3 762&-2 829&\textbf{0 005}\\ 
\#1st person pronouns (singular)&-22 886&3 982&-5 748&\bm{$\sim 0$}\\

sentiment-ANEW&16 621&9 272&1 793&0 073\\ 
\#function words&-13 126&4 825&-2 72&\textbf{0 007}\\ 
\#demonstrative pronouns&-26 514&13 569&-1 954&0 051\\  \midrule 

\textbf{Bluff}&Estimate&SE&Wald&\textit{p}-value \\ \midrule  
sentiment-ANEW&16 621&9 272&1 793&0 073\\ 
\#function words&-13 126&4 825&-2 72&\textbf{0 007}\\ 
\#demonstrative pronouns&-26 514&13 569&-1 954&0 051\\ 
\hline

\end{longtable}}

\noindent\\ 
\textbf{Per culture and cross-cultural  observations}\\
Table~\ref{table:MLR_cross} reports the per culture and cross-cultural observations.
Although the resulting feature sets are quite distinct, we observe some similarities around the usage of pronouns. Again pronouns have very large coefficients in most datasets, and the usage of {\em \#1st person pronouns} is correlated with truthful text for the individualistic native English and collectivist Romanian speakers, while the usage of {\em \#3rd person pronouns} is correlated with deception in collectivist EnglishIndia and Romanian datasets. Positive sentiment, as already discussed, prevails in native English speakers for deceptive text, while sentiment features do not play any major role in other cultures. Additionally, the  {\em \#lemmas} and {\em \#words} do not seem to discriminate between the different classes of text, and the usage of tenses plays a mixed and not significant role.
{\em \#nasals} appear to correlate with deceptive text in native English speakers, while it is the most discriminative feature for truthful text for Spanish.
For the EnglishIndia dataset, by far the most distinguishing feature is the {\em \#negations}. This is a finding that agrees with the relevant bibliography in relation to the significance of negations in South Asian languages (see also Section~\ref{sec:dc}). 
A final observation is the absence of features correlated with truthful and deceptive text in the similarly created EnglishIndia and SpanishMexico datasets respectively.

{\fontsize{10}{10}\selectfont
\begin{longtable}{lrrrrr}
\caption{Multiple logistic regression analysis for each dataset across cultures. The Russian dataset is absent since no significant features were found in the Mann-Whitney test. SE stands for Standard Error. We show in bold \textit{p}-values \textless 0.01. Positive or negative estimate values indicate features associated with deceptive or truthful text, respectively.} \label{table:MLR_cross}\\
\toprule
\textbf{USA - NativeEnglish}&Estimate&SE&Wald&\textit{p}-value  \\ \midrule 
\#boosters&8 338&3 86&2 16&0 031\\ 
\#plosives&5 024&1 916&2 623&\textbf{0 009}\\ 
positive sentiment-MPQA&3 747&0 864&4 338&\bm{$\sim 0$}\\ 
\#verbs&3 351&1 181&2 837&\textbf{0 005}\\ 
mean preverb length&0 104&0 015&6 749&\bm{$\sim 0$}\\ 
\#lemmas&0 009&0 004&2 122&0 034\\ 
\#words&-0 005&0 002&-2 479&0 013\\ 
\#punctuation marks&-0 329&0 042&-7 755&\bm{$\sim 0$}\\ 
\#articles&-2 406&1 197&-2 01&0 044\\ 
\#1st person pronouns (plural)&-11 399&2 154&-5 293&\bm{$\sim 0$}\\ 
\#exclusion words&-13 679&4 115&-3 324&\bm{$\sim 0$}\\ 
\midrule

\textbf{Belgium - Dutch}&Estimate&SE&Wald&\textit{p}-value  \\ \midrule 
\#verbs&2 614&1 202&2 175&0 031\\ 
average word length&0 321&0 113&2 853&\textbf{0 004}\\ 
\#lemmas&0 029&0 009&3 28&\textbf{0 001}\\ 
\#words&-0 016&0 005&-3 542&\bm{$\sim 0$}\\ 
mean sentence length&-0 052&0 017&-3 055&\textbf{0 002}\\ 
\#verbs in present tense&-0 848&0 307&-2 765&\textbf{0 006}\\ 
intercept&-2 657&1 114&-2 385&0 017\\

\midrule

\textbf{India - EnglishIndia}&Estimate&SE&Wald&\textit{p}-value \\ \midrule 
\#negations&15 602&4 21&3 706&\bm{$\sim 0$}\\ 
\#3rd person pronouns&6 917&2 224&3 11&\textbf{0 002}\\ 
\midrule
\textbf{Mexico - SpanishMexico}&Estimate&SE&Wald&\textit{p}-value \\ \midrule 
intercept&7 728&2 259&3 421&\bm{$\sim 0$}\\ 
\#verbs in past tense&-4 775&2 361&-2 022&0 043\\ 
\#conjunction words&-10 95&4 465&-2 453&0 014\\ 
\#prepositions&-12 67&3 939&-3 216&\textbf{0 001}\\ 
\#nasals&-23 77&7 4&-3 213&\textbf{0 001}\\ 
\midrule
\textbf{Romania - Romanian}&Estimate&SE&Wald&\textit{p}-value \\ \midrule 
\#3rd person pronouns&19 107&5 094&3 75&\bm{$\sim 0$}\\ 
\#fricatives&7 056&3 904&1 807&0 071\\ 
intercept&2 48&0 628&3 947&\bm{$\sim 0$}\\ 
\#spatial words&-1 529&0 743&-2 06&0 039\\ 
\#1st person pronouns (plural)&-12 692&6 205&-2 046&0 041\\ 
\#1st person pronouns (singular)&-21 014&8 111&-2 591&\textbf{0 01}\\ 

\bottomrule
\end{longtable}}

\section{Classification}\label{sec:classification}
In this section, we evaluate the predictive performance of different feature sets and approaches for the deception detection task. Firstly we present and discuss the results of logistic regression, then the results of
fine-tuning a neural network approach based on the state-of-the-art BERT model, 
and finally we provide a comparison with other related works.  
As a general principle, and given the 
plethora of different types of neural networks and machine learning algorithms
in general, this work does not focus on optimizing
the performance of the machine learning algorithms to the specific datasets. 
Our focus is to explore, given the limited size of training data, 
which are the most discriminative types 
of features in each domain and language, and, in succession, if the
combination of features is beneficial to the task of deception detection.

We split the datasets into training, testing, and validation 
subsets with a 70-20-10 ratio. We report the results on test sets, while 
validation subsets were used for  fine-tuning the hyper-parameters of the algorithms. In 
all cases, we report \textit{Recall}, \textit{Precision}, \textit{F-measure} and 
\textit{Accuracy}. These statistics were calculated 
according to the following definitions:
\begin{align*}
Precision\ (P): \frac{tp}{tp+fp} && F1: 2\cdot \frac{ Precision\cdot Recall}{Precision+Recall}\\
Recall\ (R): \frac{tp}{tp+fn} && Accuracy \ (Accu.): \frac{tp+tn}{tp + tn + fp + fn} 
\end{align*}
where a true positive (tp) and a true negative (tn) occurs when the model correctly predicts the positive 
class or negative class respectively, while a false positive (fp) and
a false negative (fn) when the model incorrectly predicts the positive and negative 
class respectively.

\subsection{Logistic regression experiments}
Logistic regression has been widely applied in numerous NLP tasks, among 
which deception detection from text 
\citep{Fuller:2009:DSD:1501042.1501420,W17-3608}. 
We experimented with several logistic regression models, including one based on linguistic features (i.e., {\em linguistic}), various n-grams features ({\em phoneme-gram}, {\em character-gram}, {\em word-gram}, {\em POS-gram}, and {\em syntactic-gram}), and the
{\em linguistic+} model that represents the most performant model that combines linguistic features with any of the n-gram features.
For our experiments we used two implementations of logistic regression of Weka \citep{hall09:_weka_data_minin_softw} \textit{simple logistic} \citep{Landwehr2005, Sumner2005} and \textit{logistic} \citep{leCessie1992}. The \textit{simple logistic} has a built-in attribute selection mechanism based on LogitBoost \citep{friedman2000}, while the \textit{logistic} aims to fit a model that uses all attributes. 
In all cases, we have two mutually exclusive classes (deceptive, truthful) and we use a classification threshold of 0.5.
In the case of n-grams, a preprocessing step selects the highest occurring 1000 n-gram features, while when the attribute selection is set on the  \textit{CfsSubsetEval} evaluator of Weka is used. The \textit{CfsSubsetEval} evaluator estimates the predictive power of \emph{subsets} of features.

In the following tables (Tables~\ref{opspamresults} -~\ref{all-culture}) we present the logistic regression results.
We group the native English datasets, 
and seek for differences across them, since they are written in the same language and we assume the same culture for the authors (see Section~\ref{lrseparate} and Tables \ref{opspamresults} - \ref{bluffresults}). Then we proceeded with cross-domain dataset experiments for native English datasets, by iteratively keeping each native English dataset as a testing set and using the rest as
the training set (see Section~\ref{cross_domain_paragraph} and Table~\ref{all-usdatasets}). Lastly, in Section~\ref{cross_culture_lr_experiments} 
we present cross-culture experiments.
We report only the best performed experimental setup in the test set based on the accuracy value for each feature type. The measures \textit{Precision}, \textit{Recall} and \textit{F1} refer to the deceptive class while in all cases we report a majority baseline that classifies all instances in the most frequent class. We also report AUC (surface area under a ROC curve) measure \citep{auc}. AUC value shows the probability that a randomly chosen positive instance will be ranked higher than a randomly chosen negative instance. Consequently, the closer the AUC is to 1, the better the performance of the classifier \citep{AUC_inbalance}.
The description of the experimental setup uses the following notation:\\
\begin{tabular}{rl}
$(a,b)$:& all n-grams of size in $[a,b]$, with $a \geq b$ and $a$, $b$ $\in$ $[1,3]$\\
        & (e.g., (1,2) denotes all unigrams and bigrams).\\
$stem$:&word stemming.\\
$attrsel$:&attribute selection.\\
$stop$:&stopwords removal.\\
$lowercase$:&lowercase conversion.\\
$SimpLog|Log$:&Weka algorithm (\textit{simple logistic} or \textit{logistic}).\\
\end{tabular}

\subsubsection{Native English dataset experiments}\label{lrseparate}
Tables \ref{opspamresults}-\ref{boulderresults} present the 
results
for the US datasets that concern the online reviews domains (i.e., datasets OpSpam, DeRev, and Boulder). Each dataset consists of reviews about a particular product category or 
service, with the exception of the Boulder dataset, which covers the wider domain of
hotels and electronic products.
	
\begin{table}[ht!]
 \caption{Results for the OpSpam dataset. 
 }
 {\begin{minipage}{25pc}
    \begin{tabular}{@{\extracolsep{\fill}}lp{6cm}lllll}
    \hline	
Type&Best Setup&R&P&F1&AUC&Accu.\\ 
 \hline	
Linguistic	    &	${SimpLog}$ 	            &	0 67 	&	0 73	&	0 70	&	0 80	&	0 71	\\
Phoneme-gram	&	${(1,3),SimpLog}$ 	&	0 76	&	0 79	    &	0 77	&	0 86	&	0 78	\\
Character-gram	&	${(1,3),SimpLog}$ 	        &	0 78	&	0 78	&	0 78 	&	0 86	&	0 78	\\
Word-gram	    &	${(1,2),SimpLog,stem}$ &	0 81 &	0 82	&0 81 	& 0 90	&0 82	\\
POS-gram	    &	${(3,3),Log,attrsel}$ 	        &	0 72	&	0 70	&	0 71 	&	0 78	&	0 71	\\
Syntactic-gram	&	${(1,2),SimpLog}$ 	&	0 76	&	0 70 	&	0 73	&	 0 77	&	0 72	\\
Linguistic+	    &	${Word,(1,1),SimpLog,stop,lowercase}$ 	    &	\textbf{0 88} 	&	\textbf{0 85}	&	\textbf{0 86} 	&	 	\textbf{0 91}	&	\textbf{0 86}	\\

Majority baseline& && &&&0 50 \\

    \hline
    \end{tabular}
  \end{minipage}}
\label{opspamresults}
\end{table}

\begin{table}[ht!]
  \caption{Results on the test set for the DeRev dataset.
 }
 {\begin{minipage}{25pc}
    \begin{tabular}{@{\extracolsep{\fill}}lp{6cm}lllll}
    \hline	
Type&Best Setup&R&P&F1&AUC&Accu. \\ 
 \hline	
Linguistic          &${Log,attrsel}$        &0 72&0 74&0 71&0 84&0 72\\
Phoneme-gram        &${(1,2),Log}$          &0 83&0 79&0 81&0 88&0 80 \\
Character-gram      &${(1,2),Log}$  &0 91&0 84&0 87&0 90&0 87\\
Word-gram           &${(1,1),Log,stop,stem}$&\textbf{1 00}&\textbf{1 00}&\textbf{1 00}&\textbf{1 00}&\textbf{1 00}\\
POS-gram            &${(3,3),Log}$       &0 83&0 68&0 75&0 76&0 72\\
Syntactic-gram      &${(3,3),Log,attrsel}$  &0 65&0 60&0 62&0 64&0 61\\
Linguistic+         &${Word,(1,1),Log,stop,stem}$&1 00&0 92&0 96&0 98&0 96\\
Majority baseline & && &&&0 49 \\

    \hline
    \end{tabular}
  \end{minipage}}
\label{derevresults}
\end{table}

\begin{table}[ht!]
  \caption{Results on the test set for the Boulder dataset.
  }
  {\begin{minipage}{25pc}
    \begin{tabular}{@{\extracolsep{\fill}}lp{6cm}lllll}
    \hline
Type&Best Setup&R&P&F1&AUC&Accu. \\ 
 \hline	
Linguistic&${Log,attrsel}$&0 71&0 69&0 62&0 60 &0 70\\
Phoneme-gram&${(2,3),SimpLog}$ &0 97 &0 70&0 81 &0 50&0 68 \\
Character-gram&${(3,3),SimpLog}$ &\textbf{0 99} &0 70 &\textbf{0 82}&0 53&0 70\\
Word-gram&${(1,3),Log,stop,stem,attrsel}$ &0 92 &0 73&0 81 &0 65&0 71\\
POS-gram&${(2,3),Log,attrsel}$ &0 90 &\textbf{0 75} &\textbf{0 82} &\textbf{0 67}&\textbf{0 73}\\
Syntactic-gram&${(1,3),SimpLog}$ &0 98 &0 70 &\textbf{0 82}&0 65&0 70\\
Linguistic+&${SN,(1,2),Log,attrsel}$&0 87 &0 73 &0 79&0 60&0 68\\
Majority baseline&&&&&&0 70 \\
    \hline
    \end{tabular}
  \end{minipage}}
\label{boulderresults}
\end{table}

In the OpSpam dataset (see Section~\ref{datasetopspam}) the best performance is achieved with the combination of linguistic cues with the word-gram (unigram) configuration (86 per cent accuracy). The other configurations, although not as  performant, managed to overshadow the majority baseline (see Table \ref{opspamresults}). Additionally, the second best performance of the word unigram approach showcases the importance of the word textual content in this collection.

In the DeRev dataset (see Section~\ref{datasetderev}), the word unigram configuration offers exceptional performance (accuracy of 1.00 per cent). The rest configurations achieve much lower performances. However, as in the case of the OpSpam dataset, the performance in all the configurations is a lot better than the majority baseline. 
Since we were puzzled with the 1.00 value in all measures for the unigram configuration,  we ran some additional experiments. Our results show that in this specific dataset, there are words that appear only in one class. For example, the word ``Stephen'' is connected only with the truthful class and the words ``thriller'', ``Marshall'', ``faith'', and ``Alan'' only with the deceptive class. After thoroughly checking how this collection was created, we found that the above observation is a result of how this dataset was constructed.
Specifically, the authors have used different items (i.e., books) for the deceptive and truthful cases, and as a result the classifiers learn to identify the different items.
To this end, the performance of the linguistic, POS-gram and syntactic-gram configurations are more representative for this dataset, since they are more resilient to this issue.

The Boulder dataset is quite challenging, since it includes  two domains under the generic genre of online reviews (hotels and electronics), and two types of deception i.e., lies and fabrications, (see Section~\ref{datasetboulder}).
Given the above, we observe that the performance of all classifiers 
is much lower and close to the majority baseline (as shown in Table~\ref{boulderresults}). The best accuracy is provided by the POS (bigrams+trigrams) configuration that achieves with a value of 73 per cent,  followed closely by the rest. Notice also the poor performance of the AUC measures, which is an important observation since the dataset is not balanced.

\begin{table}[ht!]
  \caption{Results on the test set for the EnglishUS dataset. 
  }
 {\begin{minipage}{25pc}
    \begin{tabular}{@{\extracolsep{\fill}}lp{6cm}lllll}
    \hline	
Type&Best Setup&R&P&F1&AUC&Accu.\\ 
 \hline	
Linguistic&${SimpLog}$&\textbf{0 62}&\textbf{0 74}&\textbf{0 67}&\textbf{0 74}&\textbf{0 70}\\
Phoneme-gram&${(1,1),SimpLog}$&0 67&0 63&0 65&0 70&0 64\\
Character-gram&${(1,1),Log,attrsel}$&0 57&0 59&0 58&0 60&0 58\\
Word-gram&${(1,3),Log}$&0 67&0 60&0 63&0 66&0 61\\
POS-gram&${(1,1),Log,attrsel}$&0 68&0 66&0 67&0 70&0 67\\
Syntactic-gram&${(1,1),Log,attrsel}$&0 70&0 60&0 65&0 70&0 62\\
Linguistic+&${Word,(1,1),SimpLog.stem}$&0 65&0 68&0 67&0 71&0 68\\
Majority baseline&&&&&&0 50 \\
    \hline
    \end{tabular}
  \end{minipage}}
\label{englishusresults}
\end{table}

The results for the EnglishUS dataset, which 
is based on deceptive and truthful essays about opinions and feelings 
(see Section~\ref{datasetspanish}), 
are presented in Table~\ref{englishusresults}. 
In this dataset, the linguistic model offers the best performance (71 per cent accuracy). The combination of linguistic cues with word unigrams, the POS-gram (unigrams) and the phoneme-gram (unigrams) configurations provide lower but relative close performance.

\begin{table}[ht!]
  \caption{Results on the test set for the Bluff dataset.
  }
 {\begin{minipage}{25pc}
    \begin{tabular}{@{\extracolsep{\fill}}lp{6cm}lllll}
    \hline	
Type&Best Setup&R&P&F1&AUC&Accu.\\ 
 \hline	
Linguistic&${Log}$&0 60&0 75 &0 67&0 60&0 60 \\
Phoneme-gram&${(1,1),SimpLog}$&\textbf{0 74}&0 65&0 69& 0 49&0 56 \\
Character-gram&${(1,1),SimpLog}$&0 69&0 71&0 70&0 56&0 60\\
Word-gram&${(1,2),Log}$&0 46&0 64&0 53& 0 58&0 46\\
POS-gram&${(1,2),SimpLog}$&0 71&0 71&0 71&0 63&0 61 \\
Syntactic-gram&${(1,2),SimpLog}$&0 63&0 73&0 68&\textbf{0 64}&0 60\\
Linguistic+&${POS,(1,3),Log}$&0 69&\textbf{0 77}&\textbf{0 73}&0 62&\textbf{0 65} \\
Majority baseline&&&&&&0 69\\
    \hline
    \end{tabular}
  \end{minipage}}
\label{bluffresults}
\end{table}

Lastly, Table~\ref{bluffresults} contains the results for the Bluff dataset, which is the only dataset that originates from spoken data and is multi-domain (see Section~\ref{datasetbluff}). All the configurations are equal or below the major baseline which is 69 per cent. Notice that this is a small unbalanced dataset with most configurations having a low AUC performance. 
The inclusion of features that elicit humorous patterns, could possibly improve the performance of the classifiers, since an integral characteristic of this dataset is humour, a feature that we do not examine in this work.

\begin{table}[ht!]
  \caption{A list with top ten discriminating deceptive features for each native English dataset. The features are listed by decreasing estimate value as calculated by the logistic regression algorithm.}
 {\begin{minipage}{25pc}
    \begin{tabular}{@{\extracolsep{\fill}}lllll}
    \hline
\textbf{OpSpam}&\textbf{DeRev}&\textbf{Boulder}&\textbf{EnglishUS}&\textbf{Bluff} \\
word-gram+ling.&word-gram&POS-gram&ling.&POS-gram+ling.
\\ \hline
\#1st person pron.(sing.) 	&	gettingbookreview	&intercept	&	\#boosters	&	\#1st person pron. (pl.) 	 \\
\#boosters	&	intrigu	&NNS IN RB	&	\#3rd person pron.&	negative -FB		\\
\#hedges	&	thriller	&	VB PRP MD	&	\#nasals	& 	\#conjuctions	\\
positive sent.-MPQA 	&	money	&TO VB PRP\$	& negative -FB		&	sent.-ANEW	 \\
forever&	bibl	&VB VBG	& \#spatial words		& \#1st person pron. \\
millennium 	&	advic	&	PRP MD RB	&intercept		&	\#negations		\\
lacking&	amaz	& NN IN	&	& \#1st person pron. (sing.) 		\\
regency	&	approach	&	RB DT	&		&	\#plosives	\\
grand &	humor	&	NN MD	&		&	\#demonstrative pron.	\\
luxury	&	addict	& DT NN	& 	& \#verbs in future tense		\\ \hline
    \end{tabular}
  \end{minipage}}
  \label{discriminative_native_deceptive}
\end{table}

\begin{table}[ht!]
  \caption{A list with top ten discriminating truthful features for each native English dataset. The features are listed by decreasing estimate value as calculated by the logistic regression algorithm.}
 {\begin{minipage}{25pc}
    \begin{tabular}{@{\extracolsep{\fill}}lllll}
    \hline
\textbf{OpSpam}&\textbf{DeRev}&\textbf{Boulder}&\textbf{EnglishUS}&\textbf{Bluff}\\
word-gram+ling.&word-gram&POS-gram&ling.&POS-gram+ling.
\\ \hline
priceline	&	intercept &NNS PRP RB	&\#1st person pron. (sing.)	&\#boosters	\\
\#spatial words&	still	& CD NNP &\#prepositions	&	\#hedges
\\
separate	&	told	&IN DT DT	&	\#conjunction words		& \#indefinite pron.\\
returned &	stand	&\$ CD	&		\#indefinite pron.	&	\#nasals 	\\
doormen	&	cour	&	RB VBG DT	&\#motion verbs	&	positive -MPQA\\
updated	&	sad	&IN JJ IN	&\#words	&		negative -FBS \\
renovated	&	pros	&NNS RB RB 
	&& \#modal verbs	\\
fault	&	master	&VBG NN IN	&	&	\#exclusion words	\\
based	&	free	&NN LRB 	&		&\#total pronouns	\\
note	&	unlik	&VB CD	&		&	positive -SentiWordNet		\\ \hline
    \end{tabular}
  \end{minipage}}
  \label{discriminative_native_truthful}
\end{table}
Tables~\ref{discriminative_native_deceptive} and~\ref{discriminative_native_truthful} present the top ten features in terms of their estimate value for each class, for the configuration with the best performance. We observe that in one-domain datasets, the content in the form of word-grams is prevalent and implicitly express deceptive patterns. This is the case for the OpSpam and DeRev datasets. For example, spatial details and verbs in past tense (i.e., told, renovated, updated, based, returned) are associated with the truthful class  while positive words (e.g., amazing, luxury, intriguing) are related to deceptive class. In the rest datasets that consist of different topics  (i.e., two in the Boulder, three in the EnglishUS and multiple in the Bluff dataset), the best performance is achieved with the use of linguistic cues, more abstract types of n-grams such as POS-grams or with the combination of linguistic cues with n-grams. We also observe the existence of the feature ``priceline'' in the OpSpam list. This refers to one of the sites from which the truthful reviews were collected (e.g., Yelp, Priceline, TripAdvisor, Expedia etc.). However, since this resembles the problem in the DeRev dataset  in which particular features mostly are associated with one class we checked that a very small percentage of truthful reviews contain such reference.
As a closing remark, we would like to showcase the rather stable performance of the linguistic models in all datasets (except maybe in the case of the Bluff dataset in which the performance of all models is hindered). As a result, the linguistic cues can be considered as a valuable information for such classification models, that 
in many cases can provide complementary information and improve the performance of other content or non-content based models.

\subsubsection{Cross dataset experiments for US datasets}\label{cross_domain_paragraph}
In this part, we examine the performance of the classifiers when they are trained on different datasets
than those on which they are evaluated. In more details, we used every native English dataset  once as a testing set for evaluating a model trained over 
the rest native English datasets.

{\fontsize{9}{9}\selectfont
\begin{longtable}{lp{4.5cm}lllll}
\caption{Cross-dataset results for US datasets
\label{all-usdatasets}} \\
\toprule
Type&Best Setup&R&P&F1&AUC&Accu. \\ 
\midrule
\multicolumn{2}{l}{\textbf{All-OpSpam}} \\ \midrule
Linguistic	&	${SimpLog}$	&	\textbf{0 91} 	&	0 51	&	0 65 	&	 0 52	&	0 52	\\
Phoneme-gram	&	${(1,1),Log}$ 	&	0 67 	&	0 55 	&	0 60	&	 0 58	&	0 56 	\\
Character-gram	&	${(1,1),Log}$	&	0 87 	&	\textbf{0 58} 	&	\textbf{0 70} 	&	0 68	&	\textbf{0 62}	\\
Word-gram	&	${(1,2),SimpLog,stem}$ 	&0 86 	&	0 57 	&	0 69 	&	0 68	&	0 61 	\\
POS-gram	&	${(1,1),SimpLog}$ 	&	0 90 	&	\textbf{0 58} 	&	\textbf{0 70} 	&	\textbf{0 70}	&	\textbf{0 62}	\\
Syntactic-gram	&	${(1,1),Log}$ 	&	0 90 	&	0 53 	&	0 67 	&	0 62	&	0 55 	\\
Linguistic+	&	${Word,(1,3),Log,lowercase}$ 	&	0 68	&	0 55 	&	0 61 	&	0 56	&	0 56	\\ 
\midrule[0.000001pt]
Majority baseline&&&&&&0 50 \\
\midrule
\multicolumn{6}{l}{\textbf{All-DeRev}} \\\midrule
Linguistic	&	${SimpLog}$ 	&	0 83 	&		\textbf{0 56} 	&	0 67 	&	0 55	&		0 58	\\
Phoneme-gram	&	${(2,3),Log}$ 	&	0 79 	&		\textbf{0 56} 	&	0 65 	&	0 61	&	0 58 	\\
Character-gram	&	${(1,2),SimpLog}$ 	&	0 89	&	0 53 	&	0 66 	&	 0 59	&	0 55	\\
Word-gram	&	${(2,2),Log,stem}$ 	&	0 78 	&	\textbf{0 56} 	&	0 65 	&	\textbf{0 64}	&	\textbf{0 59} 	\\
POS-gram	&	${(1,3),SimpLog}$ 	&	\textbf{0 90}	&	0 55 	&		\textbf{0 68}	&		\textbf{0 64}	&	0 58 	\\
Syntactic-gram	&	${(1,3),Log}$ 	&	0 66	&	0 55	&	0 60	&	0 57	&	0 56 	\\
Linguistic+	&	${Word,(1,1),Log,attrsel,lowercase}$	&	0 80	&	 0 55 	&	0 65	&	 0 52	&	0 57 	\\ 
\midrule[0.000001pt]
Majority baseline&&&&&&0 50 \\
 \midrule
\multicolumn{6}{l}{\textbf{All-Boulder}} \\\midrule
Linguistic	&	${Log,attrsel}$ 	&	0 70 	&	0 70 	&	0 70	&	0 53	&	0  58 	\\
Phoneme-gram	&	${(3,3),Log,attrsel}$ 	&	\textbf{0 90} 	&	0 70 	&	\textbf{0 79}	&	0 52	&	\textbf{0 67} 	\\
Character-gram	&	${(1,1),Log,attrsel}$ 	&	0 69 	&	\textbf{0 74} 	&	0 71	&	\textbf{0 58}	&	0 61 	\\
Word-gram	&	${(2,3),Log,attrsel,lowercase}$	&	0 79 	&	0 72 	&	0 75 	&	0 54&0 64 \\
POS-gram	&	${(1,2),Log,attrsel}$ 	&	0 62 	&	0 72 	&	0 67 	&	0 54	&	0 57	\\
Syntactic-gram	&	${(1,2),SimpLog}$ 	&	0 64 	&	0 73 	&	0 68 	&	 0 56	&	0 58 	\\
Linguistic+	&	${Phoneme,(1,1),Log}$	&	0 61	&	0 70	&	0 65 	&	 0 49	&	0 54 	\\
\midrule[0.000001pt]
Majority baseline&&&&&&0 70 \\
\midrule
\multicolumn{6}{l}{\textbf{All-EnglishUS}} \\\midrule
Linguistic	    &	${SimpLog}$ 	&	\textbf{0 98} 	&	0 50 	&	0 66 	&	 0 57	&	0 51\\
Phoneme-gram	&	${(1,1),SimpLog}$ 	&	0 95	&	0 52	&	\textbf{0 67} 	&	0 54	&	0 53\\
Character-gram	&	${(1,1),Log}$ 	&	0 89 	&	\textbf{0 54}	&	\textbf{0 67} 	&	0 58	&	0 56\\
Word-gram	    &	${(1,3),Log,stop}$	&	 0 86 	&0 53 	&0 66	&0 52	&0 56\\
POS-gram	    &	${(1,2),Log}$ 	&	0 73	&	0 53 	&	0 61 	&	0 55	&	0 54\\
Syntactic-gram	&	${(1,2),Log}$ 	&	0 76 	&	0 53 	&	0 62 	&	0 52	&	0 54\\
Linguistic+	    &	${Character,(1,2),Log}$ 	&	0 76	&	0 55 	&	0 64 	&	\textbf{0 59}	&	\textbf{0 57}\\ 
\midrule[0.000001pt]
Majority baseline&&&&&&0 50\\
\midrule
\multicolumn{6}{l}{\textbf{All-Bluff}} \\\midrule
Linguistic	    &	${SimpLog}$ 	            &	0 13 	&	0 50 	&	0 20 	&	 0 43	&	0 33\\
Phoneme-gram	&	${(3,3),Log,attrsel}$ 	&	0 88 	&	0 66	&	0 75	&	0 48	&	0 61\\
Character-gram	&	${(1,1),Log,attrsel}$ 	    &	0 63	&	0 66 	&	0 65	&	 0 45	&	0 54\\
Word-gram	    &	${(3,3),Log,stop,stem,attrsel}$ &	\textbf{0 96} 	&	0 67 	&	\textbf{0 79} 	&		0 51&	0 66\\ 
POS-gram	    &	${(3,3),Log,attrsel}$ 	&	0 60	&	0 69 	&	0 64 	&	 0 55	&	0 54\\
Syntactic-gram	&	${(1,2),Log}$ 	&	0 59	&	0 68	&	0 63 	&		0 51	&	0 54 \\
Linguistic+	    &	${POS,(1,3),SimpLog}$ &	0 84 	&	\textbf{0 71}	&0 77 	&	\textbf{0 59}&\textbf{0 67}\\
\midrule[0.000001pt]
Majority baseline&&&&&&0 67 \\
\bottomrule
\end{longtable}}	

The setting of these experiments results in highly heterogeneous datasets not only in terms of thematic but also in terms of the collection processes, the type of text (e.g., review, essay), the deception type etc. These discrepancies seem to be reflected to the results (see Table~\ref{all-usdatasets}). Overall, the results show that the increased training size, with instances that are connected with the notion of deception but in different context, and without sharing many other properties, are not beneficial for the task. Note also that the configuration in these experiments results in unbalanced datasets both in training and testing sets so the comparison is fairly demanding.

The performance for the linguistic-only setting has an average accuracy of 50 per cent. These result show that there is no overlap between the distinguishing features across the datasets that can lead to an effective feature set, 
as has already been revealed in the MLR analysis (see Table~\ref{table:MLR_cross}). In the case of the All -Bluff dataset the linguistic cues only configuration has the lowest accuracy of 33 per cent which is quite below the random chance. 
After a closer inspection, we observed that the classifier identifies only the truthful texts (recall that the Bluff dataset has an analogy 2:1 in favour of the deceptive class). 
This could be explained by the reversed direction of  important features such as the {\em\#words} compared to the rest of the datasets. 
Moreover, there are features that are statistically significant only in this dataset and not on the training collection,  e.g., the negative sentiment FBS, and vice versa, e.g., {\em\#demonstrative pronouns}.
The interested reader can find the details  in Table~\ref{u_test_us} in the Appendix and in Table~\ref{MLR_english}.

Similarly, n-gram configurations are close to randomness in most of the cases. However, topic relatedness seems to have a small positive impact on the results for the All -OpSpam and the All -Boulder datasets, as expected, since the Boulder dataset contains hotels and electronics reviews, and the OpSpam dataset also concerns hotels. The high recall values for the deceptive class in some of the classifiers depicts the low coverage and the differences between the datasets.

The POS-grams and the syntactic-grams settings that are less content dependent, fail to detect morphological and syntactical patterns of deception respectively across the datasets. This could be attributed to the fact that such n-gram patterns might not be discriminating across different datasets and due to the fact that such types of n-grams can be implicitly influenced from the unrelated content. Overall and as future work, we plan to remove strongly domain specific attributes from the feature space, in order for the training model to rely more on function words and  content independent notions. In this direction, a hint for a possible improvement is given in Tables~\ref{discriminative_cross_deceptive} and~\ref{discriminative_cross_truthful}, where the most performant models include functions words, auxiliary verbs and so on.

\begin{table}[ht!]
  \caption{A list with top ten discriminating deceptive features for each of the five cross-dataset cases. The features are listed by decreasing estimate value as calculated by the logistic regression algorithm.}
 {\begin{minipage}{25pc}
    \begin{tabular}{@{\extracolsep{\fill}}lllll}
    \hline
\textbf{All -OpSpam}&\textbf{All -DeRev}&\textbf{All -Boulder}&\textbf{All -EnglishUS}&\textbf{All -Bluff} \\

POS&word-gram&phoneme-gram&character+ling.&POS-gram+ling.
\\ \hline
\#	&	a luxur	&\textipa{wEn aI askt}	&	\#boosters	&	\#total pronouns		\\
	&	\textit{[a luxury]}	& \textit{[when we asked]}		&	&	 		\\
intercept	&	rock hotel	&\textipa{jO: f3:st bUk}	&	\#3rd person pronouns&	positive -MPQA		\\
	&	\textit{[Rock hotel]}	& \textit{[your first book]}	&	&			\\
EX	&	chicago hotel	&	\textipa{It w6z an}
	&	\#nasals	& 	\#verbs	\\
		&	\textit{[Chicago hotel]}	& \textit{[it was an]}	&	&
\\
JJR	& chicago and &	\textipa{wi: faIn@li g6t} & negative -FB		&	NN CC PRP\$	\\
	&	\textit{[Chicago and]}	& \textit{[we finally got]}	& 	&		
\\
RP&	from our	&\textipa{maI \*r u:m aI}	& \#spatial words		& NNP NNS	\\
	&	&\textit{[my room I]}	&	&		
\\
PDT 	&	husband and	&\textipa{nEkst taIm wi:}	&intercept		&	DT NNP IN		\\
&	&\textit{[next time we]}		&			&				
\\
PRP\$&	you would	&\textipa{n6t bi: steIIN}	&	& IN NNP NNP		\\
	&	& \textit{[not be staying]}	&		&	\\
VB	&	care of		&	\textipa{maI steI} at	&	\#indefinite pronouns	&	NNP CD	\\
	&	&	\textit{[may stay at]}	&	& 	\\
WRB &	with our	& \textipa{D@ dZeImz In}	& \#negations		&	NNP POS	\\
	&	& \textit{[the check in]}	&		&		\\
VBG	&	though	& \textipa{and aI \*r i:s@ntli}& \#prepositions	& NNP VBZ		\\
 	&	& \textit{[and I recently]}	&	& 	\\ \hline
    \end{tabular}
  \end{minipage}}
  \label{discriminative_cross_deceptive}
\end{table}

\begin{table}[ht!]
  \caption{A list with top ten discriminating truthful features for each of the five cross-dataset cases. The features are listed by decreasing estimate value as calculated by the logistic regression algorithm.}
 {\begin{minipage}{25pc}
    \begin{tabular}{@{\extracolsep{\fill}}lllll}
    \hline
\textbf{All -OpSpam}&\textbf{All -DeRev}&\textbf{All -Boulder}&\textbf{All -EnglishUS}&\textbf{All -Bluff}\\
POS&word-gram&POS-gram&character+ling.&POS-gram+ling.
\\ \hline
WP\$	&	next day &\textipa{@k\*r 6s D@ st\*r i:t}	&\#exclusion words	&PRP VBP IN \\
	& 	& \textit{[across the street]}	& 			& 		\\

LS&	michigan avenu	& \textipa{wi: hav steId}
	&\#3rd person pronouns	&DT NNP NNP
	\\
	&	\textit{[Michigan avenue]}&	\textit{[we have stayed]} &		&		\\
LRB	&	the swissotel	&\textipa{wi: w3: 6n}	&	\#1st person pronouns (pl.)	&\#punctuation marks\\
	&	& \textit{[we were on]}	&	& 
\\
\$ &	michigan ave	&\textipa{wi: steId hi@}	&		\#articles 	&NN MD	\\
	&	\textit{[Michigan Ave.]}	&	\textit{[we stayed here]}	&	&
\\
JJS	&	bed and	& \textipa{5 kO:n@ \*r u:m}	&\#conjuctions&		NNP IN NNP 	\\
	&	& \textit{[a corner room]}&		&
\\
FW	&	was look	&\textipa{and w6z t@Uld}	&intercept	&	PRP VBP NN	\\
	&	&\textit{[and was told]}	&		&
\\ 
NNS RB RB 	&the elev& \textipa{naIts at D@} & \#demonstrative pronouns	 & IN NN VBZ	\\
	&	\textit{[the elevator]}	& \textit{[nights at the]}		&	&	\\
WDT	&	a coupl	&\textipa{dZ\textturnv st f\*r 6m 5}	& \#1st person pronouns (sing.) 	&	PRP\$ JJS	\\
	& 	\textit{[a couple]}	&\textit{[just from a]}	&&			\\
VBP	&	was excel	&\textipa{g6t 5 g\*r eIt}	&	 \#function words			&intercept	\\
	&	\textit{[was excellent]}	&  \textit{[got a great]} &		&	 		\\
RBR	&	never sta	&\textipa{k6fi In D@}	& positive -FBS		&	\#1st person pron. (pl.) 		\\
	&	\textit{[never stayed]}	&  \textit{[coffee in the]}	& 		&		\\ \hline
    \end{tabular}
  \end{minipage}}
  \label{discriminative_cross_truthful}
\end{table}

\subsubsection{Per culture experiments}\label{cross_culture_lr_experiments}
For this series of experiments we grouped datasets based on the culture of the participants. Specifically, we experiment with individualistic datasets from the US and Belgium (Hofstede's 
individualistic scores of 91 and 75) and the collectivist datasets from India, Russia, Mexico and Romania
(individualistic scores of 48, 39, 30 and 30 respectively).
For the United States culture we used the unified NativeEnglish dataset. This dataset is unbalanced in favour of the deceptive class due to the Boulder and Bluff datasets and consists of 4,285 texts in total (2,498 deceptive and 1,787 truthful). The results are presented in Table~\ref{all-culture}. We also measured the accuracy of pairs of of the available n-gram feature types, to check if different types of n-grams can provide different signals of deception. The results show only minor improvements for some languages. We provide the results in the Appendix (see Table \ref{ngrampairs}). 

Generally and despite the fact that it is safer to examine results in a per dataset basis, it is evident that the word and phoneme-grams setups prevail in comparison with the rest of the setups. Even when the best accuracy is achieved through a combination of feature types, word and phoneme n-grams belong to the combination. This is the case for the native English dataset and the Romanian dataset (see Tables~\ref{all-culture} and~\ref{ngrampairs} respectively). Overall, for all the examined datasets the classifiers surpass the baseline by a lot.

The most perplexing result was the performance of the linguistic cues 
 in the EnglishIndia and EnglishUS datasets (results presented in Tables~\ref{englishusresults} and \ref{all-culture}) that are part of the Cross-Cultural dataset (see Section~\ref{datasetspanish}). These datasets have similar sizes, cover the same domains and were created through an almost identical process. However, we observe that while the feature sets of the EnglishUS  achieve accuracy of 71 per cent, the accuracy  drops to 54 per cent in the EnglishIndia. This is surprising, especially for same genre datasets that use the same language (i.e., EnglishUS and EnglishIndia).
 To ensure that this difference is not a product of the somewhat poor quality of text in the EnglishIndia dataset (due to the orthographic problems), we made corrections in both datasets and we repeated the experiments.  However, since the differences in the results were minor, it is difficult to identify the cause of this behaviour. One hypothesis is that this difference in the performance of the feature sets may be attributed to the different expression of deception between these two cultures, given the fact that almost all other factors are stable.
 The second hypothesis is that since most Indians are non-native speakers of English, they  use the language in the same way while being truthful or deceptive. This hypothesis is also supported by the fact that there are very few statistically important features for EnglishIndia, e.g., {\em \#negations} and the {\em \#3rd person pronoun}.
 As a result, the classifiers cannot identify the two classes,
 and exhibit a behaviour closer to randomness. 
 Notice that we might be noticing implications from both hypotheses, since the {\em \#3rd person pronoun} is also important while deceiving for the collectivistic  Romanian.  

{\fontsize{10}{10}\selectfont
\begin{longtable}{lp{6cm}lllll}
\caption{Per culture results. 
\label{all-culture}} \\
\toprule
Type&Best Setup&R&P&F1&AUC&Accu. \\ 
\midrule
\multicolumn{6}{c}{\textit{Individualistic}}\\
\midrule
\multicolumn{6}{l}{\textbf{USA-NativeEnglish}} \\ \midrule
Linguistic	    &	${SimpLog}$ 	            &	0 79 	    &	0 64 	    &	0 71 	&	0 65	&	0 62	\\
Phoneme-gram	&	${(1,1),SimpLog}$ 	        &	0 81	    &	0 67 	    &	0 73 	&	 0 68	&	0 65	\\
Character-gram	&	${(1,3),SimpLog}$ 	&	0 82	    &	0 70 	    &	0 76 	&	0 71	&	0 69	\\
Word-gram	    &	${(1,2),SimpLog,stop,lowercase}$	&	0  84 	    &	\textbf{0 73} 	&	\textbf{0  78} 	&	\textbf{0 79}	&	\textbf{0 72}\\
POS-gram	    &	${(1,1),Log}$	&	0 85	    &	0 65 	    &	0 73	&	 0 66	&	0 64	\\
Syntactic-gram	&	${(2,3),Log,attrsel}$ 	        &	0 86 	    &	0 67 	    &	0 75 	&	 0 71	&	0 67	\\
Linguistic+	    &	${Word,(1,2),SimpLog,stop,lowercase}$	&\textbf{0 82} 	&	\textbf{0 73}	    &	0 77 	&	 \textbf{0 79}	&	\textbf{0 72}	\\ \midrule[0.000001pt]
Majority baseline &&&&&&0 58 \\
\midrule
\multicolumn{6}{l}{\textbf{Belgium-CLiPS}} \\ \midrule
Linguistic	    &	${Log}$ 	            &	0 64 	    &	0 60 	    &	0 61 	&	0 69	&	0 60	\\
Phoneme-gram	&	${(1,1),SimpLog}$ 	        &	0 70 	    &	0 75	    &	0 72 	&	0 81	&	0 73	\\
Character-gram	&	${(1,3),Log,attrsel}$ 	&	0 71 	&	0 74 	&	0 73	&	 0 80	&	0 73	\\
Word-gram	    &	${(1,1),Log,stop,attrsel}$	&	\textbf{0 74}	    &	\textbf{0 80} 	&	\textbf{0 77} 	&	 \textbf{0 83}	& \textbf{0 78} 	\\
POS-gram    	&	${(3,3),Log}$	&	0 48 	    &	0 50	    &	0 49 	&	0 51	&	0 50\\
Linguistic+	    &	${Word,(1,1),Log,stem,stop,attrsel}$ 	        &	\textbf{0 74} 	    &	0 77 	    &	0 76 	&	\textbf{0 83}	&	0 76	\\
 \midrule[0.000001pt]
Majority baseline &&&&&&0 50 \\
\midrule
\multicolumn{6}{c}{\textit{Collectivistic}}\\
\midrule
\multicolumn{6}{l}{\textbf{India-EnglishIndia}} \\ 
\midrule
Linguistic	    &	${SimpLog}$ 	            &	0 60 	        &	0 54	&	0 57 	&	0 60	&	0 54 	\\
Phoneme-gram	&	${(2,3),SimpLog}$ 	        &	0 70 	&	0 53 	&	0 60 	&	0 57	&	0 60 	\\
Character-gram	&	${(1,2),Log,attrsel}$ 	    &	\textbf{0 72}	        &	0 59 	&	\textbf{0 65} 	&	0 61	&	\textbf{0 61}	\\
Word-gram	    &	${(1,2),Log,stem}$ 	        &	0 67 	       &	\textbf{0 60} 	&	0 63 	&	 \textbf{0 63}	&	\textbf{0 61}	\\
POS-gram	    &	${(2,2),SimpLog}$ 	    &	0 60 	          &	\textbf{0 60} 	&	0 60	&	0 62	&	0 60 	\\
Syntactic-gram	&	${(2,3),Log}$ 	            &	0 67	        &	0 57 	&	0 61 	&	0 59	&	0 58	\\
Linguistic+	    &	${Word,(3,3),SimpLog,lowercase}$ &	0 58 	&	0 56	&	0 57 	&	 0 58	&	0 56 	\\
\midrule[0.000001pt]
Majority baseline &&&&&&0 50\\
\midrule
\multicolumn{6}{l}{\textbf{Russia-Russian}} \\ \midrule
Linguistic	    &	${Log,attrsel}$ 	    &	0 50 	&	0 50 	&	0 50 	&	0 50	&	0 50 	\\
Phoneme-gram	&	${(1,2),Log,attrsel}$ 	            &	0 82 	&	\textbf{0 60}	&	\textbf{0 70}	&	\textbf{0 68}	&	\textbf{0 64}	\\
Character-gram	&	${(2,2),Log,attrsel}$ 	        &	0 55 	&	0 50 	&	0 52 	&	0 49	&	0 50	\\
Word-gram	    &	${(1,2),SimpLog}$	    &	\textbf{0 86} 	&	0 59 	&	\textbf{0 70} 	&	 0 63	&	\textbf{0 64} 	\\
POS-gram	    &	${(2,2),SimpLog}$ 	        &	0 68 	&	\textbf{0 60}	&	0 64	&	0 55	&	0 61 	\\
Linguistic+	    &	${POS,(2,2),SimpLog}$	        &0 45&	0 59&0 51&0 54&0 57\\
\midrule[0.000001pt]
Majority baseline &&&&&&0 50 \\
\midrule
\multicolumn{6}{l}{\textbf{Mexico-SpanishMexico}} \\ \midrule
Linguistic	    &	${Log}$ 	        &	0 62	&	0 60 	&	0 61	&	0 67&	0 60	\\
Phoneme-gram	&	${(1,3),Log}$ 	        &		\textbf{0 82} 	&		\textbf{0 70} 	&	\textbf{0 76}	&	\textbf{0 79}	&	\textbf{0 74}	\\
Character-gram	&	${(1,1),Log}$ 	    &	0 79 	&	0 60	&	0 68 	&	0 62	&	0 63	\\
Word-gram	    &	${(1,3),SimpLog,lowercase}$ &	\textbf{0 82} 	&		\textbf{0 70} 	&	\textbf{0 76}	&	\textbf{0 79}	&	\textbf{0 74}	\\
POS-gram	    &	${(1,3),SimpLog}$ 	    &	0 65 	&	0 63 	&	0 64 	&	0 62	&	0 63	\\
Linguistic	    &	${Word,(1,1),Log,stem,stop,attrsel}$	&0 62 	&	0 64	&	0 63 	&	0 65 	&0 63	\\
\midrule[0.000001pt]
Majority baseline &&&&&&0 50 \\
\midrule
\multicolumn{6}{l}{\textbf{Romania-Romanian}} \\ \midrule
Linguistic	    &	${SimpLog}$ 	                &	0 62 	&	0 64 	        &	0 63	&	0 67	&0 64	\\
Phoneme-gram	&	${(1,2),Log,attrsel}$ 	&	0 59	&	0 68 	        &	0 63 	&	0 69	        &	0 66	\\
Character-gram	&	${(1,2),SimpLog}$ 	        &\textbf{0 66}	&0 59	&	0 62 	&	0 60      &0 62	\\
Word-gram	    &	${(1,3),Log,stem,attrsel}$ 	    &	0 61&0 67 	&0 64 	&	0 72	            &0 65 	\\
POS-gram	    &	${(1,1),Log,attrsel}$ 	        &	0 62 	&	0 59	        &	0 61 	&	0 60	        &	0 64	\\
Linguistic	    &	${Phoneme,(1,2),Log,attrsel}$	&	0 61	&	\textbf{0 72}	        &	\textbf{0 66}	&	\textbf{0 70} 	        &	\textbf{0 68}	\\

\midrule[0.000001pt]
Majority baseline &&&&&&0 50 \\
\bottomrule
\end{longtable}}

\begin{table}[ht!]
  \caption{A list with 10 discriminating deceptive features for each dataset. The features are listed by decreasing estimate value as calculated by the logistic regression algorithm. In square brackets is the English translation.}
 {\begin{minipage}{25pc}
    \begin{tabular}{@{\extracolsep{\fill}}llllll}
    \hline
\textbf{Native English}&\textbf{CLiPS}&\textbf{EnglishIndia}&\textbf{Russian}&\textbf{SpanishMexico} &\textbf{Romanian}\\
word-gram&word-gram&word-gram&word-gram&word-gram & phoneme-gram+ling.
\\ \hline
how to	&	7	&then it	&	\selectlanguage{russian}проснулась	&	intercept	&	\#3rd person pron.	\\
	&	\textit{[7]}	& \textit{[then it]}		&	\textit{[woke up]}	&	 	&	\\
luxury	&	clichés	&intercept	&	\selectlanguage{russian}ним	&		&	\#1st person pron. (plural)	\\
	&	\textit{[clichés]}	& 	&	\textit{[him]}	&		&		\\
month	&	Potter	&	lie	&	intercept	& 	&	\#negations	\\
	&	\textit{[Potter]}	& \textit{[lie]}	&		&		&	
\\
cleaning	&	ober	&person he	&		&		&\#prepositions	\\
	&	\textit{[waiter]}	& \textit{[person he]}	&		&		&		
\\
i needed	&	menukaart	&very good	&		&		&	\#fricatives	\\
\textit{[I needed]}	&	\textit{[menu]}	&\textit{[very good]}	&		&		&	
\\
turned	&	vegetarisch	&	difficult	&		&		&	\#function words	\\
	&	\textit{[vegetarian]}	&\textit{[difficult]}		&		&		&	
\\
all i	&	cappuccino	& a punishment	&	&		&	\textipa{doR" esk} \textipa{s" a}	\\
\textit{[all I]}	&	\textit{[cappuccino]}	& \textit{[a punishm]}	&		&		&	\textit{[I want to]}	\\
seemed to	&	horrorfilm	&	she never	&		&		&	\#conjunctions	\\
	&	\textit{[horror movie]}	&	\textit{[she never]}	&		&		&	\\
intercept &	centrum	&	i would	&		&		&	\textipa{" " intelidZ" enta}	\\
	&	\textit{[centre]}	& \textit{[I would]}	&		&		&	\textit{[intelligence]}	\\
be staying	&	opslagruimte	& and not	& 	&		&	intercept	\\
	&	\textit{[storage area]}	& \textit{[and not]}	&		&		&		\\ \hline
    \end{tabular}
  \end{minipage}}
  \label{discriminative_deceptive}
\end{table}

\begin{table}[ht!]
  \caption{A list with 10 discriminating truthful features for each dataset. The features are listed by decreasing estimate value as calculated by the logistic regression algorithm. In square brackets is the English translation.}
 {\begin{minipage}{25pc}
    \begin{tabular}{@{\extracolsep{\fill}}llllll}
    \hline
\textbf{Native English}&\textbf{CLiPS}&\textbf{EnglishIndia}&\textbf{Russian}&\textbf{SpanishMexico} &\textbf{Romanian}\\
word-gram&word-gram&word-gram&word-gram&word-gram& phoneme-gram+ling.
\\ \hline
my best	&	fastfoodketen &he should	&\selectlanguage{russian}где-то	&	mi mejor 	&\textipa{kl" as}	\\
	& \textit{[fast food chain]}	& \textit{[he should]}	& \textit{[somewhere]}			& 	\textit{[my best]}		& 	\textit{[class]}	\\
	
rate	&	Harry	&help me
	&\selectlanguage{russian}в дверь 	&	en	&	\textipa{" un\textsuperscript{j}} \textipa{" alt}
	\\
	&	\textit{[Harry]}&	\textit{[help me]} &		\textit{[in the door]}	&		\textit{[in]}	&	\textit{[another]}	\\\

river	&	Parijs	&the girl	&	\selectlanguage{russian}очередной	&	&	\textipa{tS" e} \textipa{f" atSe}	\\
	&	\textit{[Paris]}	& \textit{[the girl]}	&	\textit{[regular]}	&		&	\textit{[what he is doing]}	
\\
michigan &	schitterend	&to him	&		\selectlanguage{russian}после чего	&	 &	\textipa{n" oj} \textipa{" Oamen\textsuperscript{j}" " \textsuperscript{j}}	\\
	\textit{[Michigan]}	&	\textit{[splendid]}	&	\textit{[to him]}	&	\textit{[then]}	&		&	\textit{[we people]}	
\\
returned	&	effecten	&	always there	&\selectlanguage{russian}вечером 	&	&	\#total pron.	\\
	&	\textit{[effects]}	& \textit{[always there]}&		\textit{[in the evening]}	&	&	
\\
elevator	&	Katniss	&in our	&	\selectlanguage{russian}мной	&		&	\#1st person pron.	\\
	&	\textit{[Katniss]}	&\textit{[in our]}	&	\textit{[me]}		&	&
\\
stayed here	&	sla	&are do 
	&&	&	omului	\\
	&	\textit{[lettuce]}	& \textit{[are doing]}		&		&	&	\textit{[human]}	\\
i believe	&	2011	&name	&	&		&\textipa{b" ut} \textipa{" in}	\\
	\textit{[I believe]}	&	\textit{[2011]}	&\textit{[name]}	&		&	&	\textit{[but in]}	\\
location	&	Woody	&els 	&		&		& \#nasals	\\
	&	\textit{[Woody]}	&  \textit{[else]} &		&		&		\\
2	&	broodjes	&so abort	&		&		&	\textipa{resp" ekt}	\\
	&	\textit{[sandwiches]}	&  \textit{[so abortion]}	&	&	&	\textit{[respect]}	\\ \hline
    \end{tabular}
  \end{minipage}}
  \label{discriminative_truthful}
\end{table}

Lastly, to get a visual insight over the above results we present the most valuable features for the configuration that achieved the best accuracy in the logistic regression experiments for all the examined datasets (see Tables~\ref{discriminative_deceptive} and ~\ref{discriminative_truthful}). The features are listed by decreasing estimate value. 
Most of the cases include morphological and semantic information that has been explicitly defined in linguistic cues (e.g., the use of pronouns as in ``my room'', tenses, spatial details, polarized words etc.). As a result, the combination of such n-gram features with linguistic cues do not work in synergy.
Moreover, notice the contribution of two features for discriminating deception in the SpanishMexico; the word ``mi mejor'' and the word ``en'' both attributed to the deceptive class. A similar behaviour with a small resulting feature set is also evident in the Russian dataset. 
	
\subsubsection{Discussion on features}\label{featuresdiscussion}
Among all the variations of n-grams tested in this work, word n-grams achieve  
the best results across almost all the datasets. The results for the other types of n-grams seems to be a little lower and to fluctuate in a per dataset basis. More content-based n-gram types such character-grams and phoneme-grams have an adequate performance while the other variations that bear more abstract and generalized linguistic information, such as POS n-grams and syntactic n-grams achieve lower performance. However, POS-gram seem to perform quite better than the syntactic n-grams.
The difference in accuracy decreases in cross domain experiments in which semantic information is more diverse, and as already discussed, linguistic indications of deception change from one domain to another. Lastly, stemming,  stopwords removal and lowercase conversion are generally beneficiary, so it is a preprocessing step that must be examined.
The experimental results show that the discriminative power of linguistic markers of deception is overly better than random baseline and the expected human performance (according to literature slightly better than chance, see Section~\ref{sec:bg}) especially in one domain scenarios  (see Tables~\ref{opspamresults} to~\ref{bluffresults}). More specifically, linguistic markers of deception are struggling in cross domain settings (see  Section~\ref{cross_domain_paragraph}). This confirms that linguistic markers of deception  vary considerably and are extremely sensitive even within the same culture, let
alone across different cultures (see Table~\ref{all-culture}). Different domains,
individual differences, even the way the texts were collected seem to influence
the behaviour of linguistic markers and indicate how complex the deception detection task is. In the native English case in which the employed feature set is richer and in general the linguistic markers are more well-studied, we can observe better results. This might  signal that there are opportunities for enhancement.

Lastly, the combination of linguistic features with n-gram variations does not enhance the performance in a decisive way in most of our experiments. N-grams and more often word-grams or phoneme-grams in an indirect way can capture information that has been explicitly encoded in the linguistic cues. However, there are cases when this combination can improve the performance of the classifier. In such cases the resulting feature space succeeds to blend content with the most valuable linguistic markers.

\subsection{BERT experiments}

In these experiments, we use BERT \citep{devlin-etal-2019-bert} with a task-specific linear classification layer on top,  using the sigmoid activation function,  as an alternative to the logistic regression classifiers of the previous experiments\footnote{For our experiments we used the python libraries tensorflow 2.2.0, keras 2.3.1, and the bert-for-tf2 0.14.4 implementation of google-research/bert, over an AMD Radeon VII card and the ROCm 3.7 platform.}. As already discussed in Section~\ref{embeddings}, BERT is already pretrained on a very large unlabelled corpus. Here it is further trained (`fine-tuned') jointly with the task-specific classifier on deception detection datasets to learn to predict if a text is deceptive or not. BERT produces context-aware embeddings for the tokens of the input text, and also an embedding for a special classification token ([CLS]), intended to represent the content of the entire input text. Here the input to the task-specific linear classifier is the embedding of the [CLS] token. We do not `freeze' any BERT layers during fine-tuning, i.e., the weights of all the neural layers of BERT are updated when fine-tuning on the deception detection datasets, which is the approach that typically produces the best results in most NLP tasks. We use \textit{categorical cross entropy} as the loss function during fine-tuning, and \textit{AdamW} as the optimizer \citep{adamw}. Finally, we exploit monolingual BERT models for each language (see  Table~\ref{bert}), as well as the multilingual multiBERT model. The BERT limitation of processing texts up to 512 wordpieces does not affect us, since the average length of the input texts of our experiments is  below this boundary (see Table~\ref{average}). However, due to batching and GPU memory restrictions,  the upper bound of the used text length was 200 wordpieces, so there is some loss of information due to text truncation, though it is limited overall. More specifically, the truncation affects 5.6 per cent of the total number of texts of all the datasets used in our experiments (506 texts out of a total of 8,971). The effect of truncation is more severe in the Bluff, OpSpam and Russian datasets, where 41 per cent (109 out of 267), 21 per cent (332 out of 1600) and  29 per cent (65 out of 226) of the texts were truncated, respectively; the average text length of the three datasets is 190, 148 and  160 wordpieces, respectively. In the other datasets, the percentage of truncated texts was much smaller (10 per cent or lower). We note that valuable signals may be lost when truncating long texts, and this is a limitation of our BERT experiments, especially those on Bluff and OpSpam,  where truncation was more frequent. For example, truthful texts may be longer, and truncating them may hide this signal, or vice versa. Deceptive parts of long documents may also be lost when truncating. In such cases, models capable of processing longer texts can be considered, such as hierarchical RNNs \citep{Chalkidis2019neural, Nishant2019} or multi instance learning as in \citep{Nishant2019}. No truncation was necessary in our logistic regression experiments, but long texts may still be a problem, at least in principle. For example, if only a few small parts of a long document are deceptive, features that average over the entire text of the document may not capture the deceptive parts. We leave a fuller investigation of this issue for future work.

In addition, we combined BERT with linguistic features. To this end, we concatenate the embedding of the [CLS] token with the linguistic features, and pass the resulting vector to the task-specific classifier. In this case, the classifier is a  multilayer perceptron with one hidden layer,
consisting of  128 neurons with ReLU activations. The MLP also includes \textit{layer normalization} \citep{ba2016layer} and a \textit{dropout} layer \citep{dropout} to avoid overfitting. 
Hyperparameters were tuned by random sampling 60 combinations of values and keeping the combination that gave the minimum validation loss. Early stopping with patience 4 was used on the validation loss to adjust the number of epochs (the max number of epochs was set to 20). The tuned hyperparameters were the following: \textit{learning rate} (1e-5, 1.5e-5, 2e-5, 2.5e-5, 3e-5, 3.5e-5, 4e-5), \textit{batch size} (16, 32), \textit{dropout rate} (0.0, 0.05, 0.1, 0.15, 0.2, 0.25, 0.3, 0.35, 0.4, 0.45) \textit{max token length} (125, 150, 175, 200, and average training text length in tokens), and the used \textit{randomness seeds} (12, 42, and a random number between 1-100).

Tables~\ref{bertresults} and ~\ref{bertcultureresults} present the results for these experiments. The former presents the results for each native English dataset, while the latter for the cross-culture datasets. For the US culture we used the unified dataset NativeEnglish. We explored both the BERT model alone and the BERT model augmented with the whole list of linguistic cues of deception studied in this work. For the native English cases we used the BERT model for the English language, while for the per culture experiments, we experimented with both the language monolingual models and the multilingual version of the BERT model.  
The dataset subscript declares the experimental setup, e.g., $_{bert+linguistic,en}$ uses the English language BERT model along with the linguistic cues.
Exactly what types of linguistic or world knowledge BERT-like models manage to capture (or not) and the extent to which they actually rely on each type of captured knowledge is the topic of much current research \citep{rogers2020primer}. It has been reported that the layers of BERT probably capture different types of linguistic information, like surface features at the bottom, syntactic features in the middle and semantic features at the top \citep{jawahar-etal-2019-bert}. Fine-tuning seems to allow retaining the most relevant types of information  to the end-task, in our case deception detection.

\begin{table}[ht!]
  \caption{Results on fine tuning BERT model for US datasets.}
 {\begin{minipage}{25pc}
    \begin{tabular}{@{\extracolsep{\fill}}lllll}
    \hline
Experiment&R&P&F1&Accu.\\ \hline 
OpSpam$_{bert,en}$&0 94	&0 89	&0 91
&\textbf{0 90} \\
OpSpam$_{bert+linguistic,en}$&0 84	&0 96	&0 90&\textbf{0 90} \\
Boulder$_{bert,en}$&0 70	&0 89	&0 79
&\textbf{0 67} \\
Boulder$_{bert+linguistic,en}$&0 70	&0 88	&0 78
&\textbf{0 67} \\
DeRev$_{bert,en}$&0 89	&1	&0 94 &0 94 \\  
DeRev$_{bert+linguistic,en}$&0 96&0 96&0 96 &\textbf{0 96} \\
EnglishUS$_{bert,en}$&0 59	&0 88	&0 71 &0 74 \\
EnglishUS$_{bert+linguistic,en}$&0 64	&0 87	&0 74 &\textbf{0 76}\\
Bluff$_{bert,en}$& 0 93 	&0 85	&0 89 &\textbf{0 83}  \\
Bluff$_{bert+lingustic,en}$&0 88	&0 82	&0 85 &0 77\\ 
    \hline
    \end{tabular}
  \end{minipage}}
  \label{bertresults}
\end{table}

\begin{table}[ht!]
  \caption{Per culture results for: a. the fine-tuned  BERT model  b. the fine-tuned  BERT model along with the linguistic features. Results are reported both for the monolingual and  the multilingual BERT models.}
 {\begin{minipage}{25pc}
    \begin{tabular}{@{\extracolsep{\fill}}lllll}
    \hline
Experiment&R&P&F1&Accu. \\ \hline

\multicolumn{5}{c}{\textit{Individualist}}\\
\midrule

\multicolumn{5}{l}{\textbf{USA-NativeEnglish}}\\
NativeEnglish$_{bert,en}$&0 79 	&\textbf{0 79}	&0 79
&0 75\\ 
NativeEnglish$_{bert+linguistic,en}$&\textbf{0 86}	&0 77	&\textbf{0 81}
&\textbf{0 77}\\
NativeEnglish$_{bert,multi}$&0 82	&0 75	&0 78
&0 73\\
NativeEnglish$_{bert+linguistic,multi}$&0 81	&0 76	&0 78
&0 74\\

\multicolumn{5}{l}{\textbf{Belgium-CLiPS}} \\
CLiPS$_{bert,nl}$&0 73	&0 78	&0 75
&0 74\\
CLiPS$_{bert+linguistic,nl}$&\textbf{0 77}	&0 78	&0 77
&0 77\\
CLiPS$_{bert,multi}$&0 74	&\textbf{0 86}	&\textbf{0 80}
&\textbf{0 80}\\
CLiPS$_{bert+linguistic,multi}$&0 68	&0 82	&0 75
&0 75\\\midrule

\multicolumn{5}{c}{\textit{Collectivist}}\\
\midrule

\multicolumn{5}{l}{\textbf{India-EnglishIndia}} \\
EnglishIndia$_{bert,en}$&0 39	&\textbf{0 70}	&0 50
&0 62\\ 
EnglishIndia$_{bert+linguistic,en}$&\textbf{0 66}	&0 63	&\textbf{0 64}
&\textbf{0 70}\\ 
EnglishIndia$_{bert,multi}$&0 64	&0 64	&\textbf{0 64}
&0 64\\ 
EnglishIndia$_{bert+linguistic,multi}$&0 19	&0 50	&0 27
&0 51\\
\multicolumn{5}{l}{\textbf{Russia-Russian}} \\
Russian$_{bert,ru}$&0 32	&\textbf{0 62}	&0 42
&\textbf{0 56}\\
Russian$_{bert+linguistic,ru}$&\textbf{0 64}	&0 48	&\textbf{0 55}
&0 48\\
Russian$_{bert,multi}$&\textbf{0 64} &0 44	&0 52
&0 42\\
Russian$_{bert+linguistic,multi}$&0 52	&0 43	&0 47
&0 42\\

\multicolumn{5}{l}{\textbf{Mexico-SpanishMexico}}\\
SpanishMexico$_{bert,es}$&0 42	&\textbf{0 76}	&0 54
&0 70\\
SpanishMexico$_{bert+linguistic,es}$&\textbf{0 65}	&0 65	&0 65
&0 70\\ 
SpanishMexico$_{bert,multi}$&0 39	&0 63	&0 48
&0 65\\  
SpanishMexico$_{bert+linguistic,multi}$&0 61	&0 70	&\textbf{0 66}
&\textbf{0 73}\\

\multicolumn{5}{l}{\textbf{Romania-Romanian}}\\
Romanian$_{bert,ro}$&\textbf{0 74}	&0 67 	&0 70
&0 68\\  
Romanian$_{bert+linguistic,ro}$&0 65&\textbf{0 83} &\textbf{0 73}
&0 69\\ 
Romanian$_{bert,multi}$&0 60	&0 62	&0 61
&0 61\\  
Romanian$_{bert+linguistic,multi}$&0 69	&0 71	&0 70
&\textbf{0 71}\\
    \hline
    \end{tabular}
  \end{minipage}}
  \label{bertcultureresults}
\end{table}

Overall, the experiments show similar, and in some cases improved results, compared to the logistic regression ones and the available related work (see Section~\ref{comparisonsection}). As shown in Table ~\ref{bertresults} this is the case for the OpSpam, Boulder, and EnglishUS datasets, while the performance drops a bit in the case of the DeRev dataset for the plain BERT model (the excellent 98 per cent accuracy drops to 94 per cent for the plain BERT model, rising again to 96 per cent for the combined BERT with the linguistic features). An interesting point is that for the Bluff dataset, the plain BERT model offers better performance to the logistic classifier (83 per cent accuracy compared to 75 per cent), which drops to 77 per cent when combined with the linguistic features. This is the only case where the addition of the linguistic features drops the performance of the classifier.
The reason might be that the plain BERT model possibly manages to capture humour, which is an internal feature of this dataset and a feature not captured by the linguistic features.

Regarding the per culture datasets shown in Table ~\ref{bertcultureresults} and compared to the logistic regression experiments, there are clear gains in the accuracy of most of the models
for the NativeEnglish, EnglishIndia and CLiPS datasets.
However, this is not the case for the SpanishMexico, and the Russian datasets.
Especially in the case of the peculiar Russian dataset, out of the four experimental setups, only the BERT alone setup with the dedicated Russian BERT model slightly surpassed the statistical random baseline of 50 per cent.
Recall that similar low performance is not only evident in our logistic experiments but also in the related work.
The low performance in the case of BERT, where there are no feature extraction steps that can  propagate misfires of the used tools or a problematic handling from our side,
showcases that this is  an intrinsically problematic collection.

A rather important finding is the contribution  of the linguistic features. The addition of the linguistic features to the BERT models
leads to better performance in many of the experiments, such as in the case of EnglishUS, DeRev, NativeEnglish, SpanishMexico, Romanian and EnglishIndia datasets.
This showcases their importance compared to the corresponding logistic regression experiments,
where the linguistic cues improved the n-grams approaches only in the case of the SpanishMexico and the Russian dataset.
The linguistic features seem to work better when combined with the BERT classifier, 
which might be the result of the model learning non-linear combinations of the features.
As already mentioned, in the case of the DeRev dataset the addition of the linguistics cues greatly improves the performance of the classifier,
leading to almost excellent performance.
Even though we have not made explicit experiments to identify which are the helpful linguistic cues in the case of BERT models,
we can speculate that they are phoneme related features, e.g., {\em \#fricatives, \#plosives, \#nasals}, and the punctuation feature.
These are significant features, which either the BERT models cannot capture or exploiting their explicit counts seems to be more effective (see Tables~\ref{u_test_us} and~\ref{u_test_culture} in the Appendix). 

Table~\ref{bertVSmultibert} provides a comparison between the monolingual BERT models  and multiBERT. In particular, monolingual BERT models seem to perform better, except in Dutch and Romanian.
Despite the lower performance of the multiBERT model, the difference is not prohibitive.

\begin{table}[ht!]
  \caption{Comparison between the monolingual BERT models and the multilingual model. We report the average accuracy of the monolingual BERT model among the BERT-only and the BERT+linguistic setups and for the multiBERT model respectively. 
  With bold font we mark the best accuracy. St. sign. stands for statistical significance. We performed a 1-tailed z-test with a 99 per cent confidence interval and $\alpha  =  0.01$.}
 {\begin{minipage}{25pc}
    \begin{tabular}{@{\extracolsep{\fill}}lcccc}
    \hline
    Dataset& Avg. accu. BERT&Avg. accu. multiBERT&St. sign.& 1-tailed probability\\
    \hline
NativeEnglish& \textbf{76 0}&73 5&yes&0 039\\
CLiPS        & 75 5&\textbf{77 5}&no & 0 115\\
EnglishIndia& \textbf{66 0}& 57 5&yes&0 001\\
Russian    & \textbf{52 0}&42 0 &no&0 016\\
SpanishMexico         &\textbf{70 0}&69 0 &no&0 388\\
Romanian         &68 5&\textbf{66 0}&no&0 133 \\\hline
    \end{tabular}
  \end{minipage}}
  \label{bertVSmultibert}
\end{table}

\subsubsection{Cross-language experiments}
In this section, we proceed with cross language experiments due to the adequate performance of the multiBERT  model.
The idea of the experiment is to fine-tune a BERT model over one language and test its performance
over another language, trying to exploit similarities in the morphological, semantic and syntactic information encoded in BERT layers, across cultures.
Our main focus is on cultures that are close in terms of the individualism dimension, thus could possibly share similar deceptive patterns that BERT can recognize. We are also interested in cross-cultural
experiments to evaluate to what extent BERT can distinguish between deceptive and truthful texts in a crosslingual setting. Finally, we have also added the EnglishUS dataset to experiment with same domain and alike collection procedure but cross language datasets (i.e., Romanian, SpanishMexico, EnglishIndia and EnglishUS).  We also performed experiments with the  NativeEnglish minus the EnglishUS collection to explore the effectiveness of a large training dataset to a different domain (EnglishUS) and to different cultures (Romanian, SpanishMexico, EnglishIndia).
For each experiment we trained a model over the 80 per cent of a language specific dataset, validated the model over the rest 20 per cent of the same dataset, and then tested the performance of the model over the other datasets. Notice that these experiments are not applicable for the NativeEnglish and EnglishUS datasets, since the former is a superset of the latter.

For most of the experiments the results are close to randomness. For example this is the case when Russian and Dutch (CLiPS) are used either as testing or training sets with any other language and when the 
combined NativeEnglish dataset is used for testing on any other language. For the Russian language this is quite expected given the performance in the monolingual experiments. However, on the Dutch dataset, the situation is different, since the fine-tuned BERT model manages to distinguish between deceptive and truthful texts in the monolingual setting but when the multiBERT is trained on the Dutch dataset, it doesn't perform well on the other datasets.

The Romanian, SpanishMexico, EnglishUS and EnglishIndia datasets that are part of the Cross-Cultural Deception dataset (see Section~\ref{comparisonsection}) show a different behaviour.
A model trained on one dataset offers an accuracy between 60 and 70 per cent on the other set using the multiBERT,
with SpanishMexico exhibiting the best performance when is is used as testing set for the EnglishUS trained model. This indicates that the domain is an important factor that alleviates the discrepancies in terms of culture and language in the crosslingual multiBERT setting. A reasonable explanation might be vocabulary memorization or lexical overlap, which occurs when word pieces are present during fine-tuning and in the language of the testing set. However, according to \citet{pires2019multilingual} multiBERT has the ability to learn even deeper multilingual representations.

Another important observation is the performance whenever the NativeEnglish is used as training set. The domain similarity is rather small in this case, since NativeEnglish is a largely diverse
dataset.
The results show that multiBERT can possibly reveal connections in a zero-shot transfer learning setting when the training size is quite adequate. This has been observed also in other tasks, like the multilingual and multicultural irony detection in the work of  \citet{multibert-irony}. In this case
instead of the multiBERT model, the authors applied an alignment of monolingual word embedding spaces in an unsupervised way.  Zero-shot transfer learning for specific tasks based on multiBERT is also the focus of other recent approaches
\citep{pires2019multilingual} \citep{libovick2019languageneutral}, that show  promising results. Removing the EnglishUS dataset from the NativeEnglish dataset reduces considerably the performance in the Romanian, SpanishMexico and EnglishIndia datasets, showcasing the importance of domain even for cross-lingual datasets. Notice though that for the SpanishMexico and the Romanian datasets the performance is greater than that of a random classifier, indicating cues of the zero-shot transfer connection hypothesis at least for this dataset. On the other hand, the random performance for the EnglishUS and EnglishIndia datasets, that have the same language with the trained model and which additionally belong to the same domain with the SpanishMexico and Romanian datasets, showcases that it is difficult to generalize.

\begin{table}[ht!]
    \caption{Comparison for the multilingual model when a model is trained over one language and then tested on another one. With bold are the accuracy values over 60 per cent. Rom. is Romanian, SpanMex. is SpanisMexico.}
 {\begin{minipage}{25pc}
    \begin{tabular}{|@{\extracolsep{\fill}}l|p{0.7cm}p{1.0cm}p{0.7cm}p{0.7cm}||p{0.7cm}p{0.7cm}||p{0.7cm} | p{1.8cm} |}
    \hline
    \textbf{Training}/\textbf{Testing}  & Rom&SpanMex&Rus& EnIn&CLiPS&NatEn&EnUS & NatEn $\setminus$ EnUS \\
    \hline
Rom          & - & \textbf{65 6} & 47 3  & 54 6 & 49 6 & 52 2& \textbf{61 5} & 50 7\\\hline
SpanMex     & \textbf{64 8} & -  & 49 5 & 57 8 & 50 6 & 52 8& \textbf{60 3} & 51 5\\\hline
Rus             &46 3 & 45 9 &  - & 49 0& 51 0& 46 2 & 49 6 & 45 6\\\hline
EnIn        &\textbf{61 0}& \textbf{67 0} & 50 0& - &50 2& 49 2& \textbf{68 0} & 53 7\\\hline\hline
CLiPS               & 44 4 & 47 1 &  52 2 & 48 5 & - & 54 0 & 48 6  & 54 8\\\hline
NatEn   & \textbf{60 5} & \textbf{67 0 }& 47 7 & \textbf{63 5 }& 50 2 &  - & N/A & N/A\\\hline\hline
EnUS    & \textbf{63 4} & \textbf{69 9} & 50 0  & \textbf{60 0} & 52 0 & N/A & - & 48 5 \\\hline
NatEn $\setminus$ EnUS  & 56  0 & 59 2& 49 5 & 50 3& 49 8 &  N/A & 51 0 &  -\\\hline
    \end{tabular}
  \end{minipage}}
  \label{crossLanguageMultiBert}
\end{table}

\subsection{Comparison with other works}\label{comparisonsection} 

Table~\ref{comparisonresults} provides an overall comparison between our best 
experimental setup and results, with those presented in other studies on the same 
corpora. The comparison was based on the accuracy scores reported in those studies. 
In addition, we report human accuracy whenever it is available. 
For comparison purposes, we set a \textit{p}-value of 0.01 and performed a 
1-tailed z-test evaluating if the differences between two proportions are statistically significant.
By comparing absolute numbers only, the comparison is not so straightforward and cannot easily lead
to conclusions, since the studies employed different model validation techniques and set difference research goals.

To the best of our knowledge, the only computational work that 
address cross-cultural  deception detection is the work of \citet{miha}. In that work, 
the authors build separate deception classifiers for each examined culture,
and report a performance ranging between 60-70 per cent.
Then they build cross-cultural classifiers by applying two alternative approaches. The first one was through the translation of unigrams and the second one by using equivalent LIWC semantic categories for each language. Both approaches resulted in lower performances. All the approaches were tested on the Cross-Cultural Deception dataset, which  was created by the authors \citep{perezrosas-mihalcea:2014:P14-2,miha}, and which we also used in this work (see Section~\ref{comparisonsection}). The treatment is different since each sub-domain dataset (death penalty, abortion, best friend) is separately examined. However, since average scores are also reported we compare this work with those scores. In addition, since the EnglishUS dataset has been extensively used in other works in the same way, we also report the average accuracy for these cases.

The comparison in Table~\ref{comparisonresults} shows that BERT outperforms other approaches in most of the cases. BERT's performance is mostly surpassed in the relatively smaller sized datasets, indicating the need for fine-tuning BERT 
over a large number of training samples.
In particular, BERT achieves state-of-the-art performance for the OpSpam dataset, that is the gold standard for opinion spam detection. In addition, for the CLiPS dataset, the BERT model outperforms the other models studied in this work, as well as another unigram approach in the bibliography \citep{L14-1001}. For the Cross-Cultural Deception dataset (see Section~\ref{datasetspanish}) BERT outruns other approaches that are based on feature engineering for the Romanian and the EnglishIndia  datasets. In the case of SpanishMexico dataset the combination of linguistic cues with word n-grams seems to have a strong discriminative power and in the EnglishUS dataset the combination of latent Dirichlet allocation topics (LDA) with a word-space model achieves the highest accuracy. 
Lastly, in comparison with human judgments, for the two datasets that we have numbers (i.e., OpSpam and Bluff), the automatic detection approaches significantly outperform human performance with respect to the accuracy measure.

\begin{table}[ht!]
  \caption{Comparison with other works on the same corpora. Bold values denote models studied in this work and the best scores.
  Accu. stands for 
Accuracy and 
St. Sign. marks cases where a statistically significant difference between this work 
and related work was found.}
 {\begin{minipage}{25pc}
    \begin{tabular}{@{\extracolsep{\fill}}p{11cm}ll}
    \hline
 \textbf{Work} & \textbf{Accu. (\%)}  & \textbf{St. Sign.} \\ \hline
\multicolumn{1}{c}{\emph{OpSpam}}{\textit{1,600 samples}}\\ 
\textbf{BERT} &   \textbf{0 90}&  - \\
\textbf{BERT+Linguistic} &   \textbf{0 90}&  - \\
BERT \citep{kennedy-etal-2019-fact} & 
\textbf{0 90} & no \\
RCNN \citep{ZHANG2018576} &0 88 & no\\
Psycholinguistic+word bigrams
\citep{Ott:2011:FDO:2002472.2002512,Ott13negativedeceptive} & 0 87 & no \\
\textit{Human performance} \citep{Ott:2011:FDO:2002472.2002512,Ott13negativedeceptive}&  
0 59& yes\\

\multicolumn{1}{c}{\emph{DeRev}}{\textit{236 samples}}\\ 
\textbf{Word n-gram} &   \textbf{1 00} &  - \\ 
LDA+word space model \citep{journals/soco/Hernandez-Castaneda17}&  0 95& 
yes \\
Various numeric features, e.g., length of reviews, frequency of n-grams etc. \citep{articlederev} &0 76& yes     \\
 
\multicolumn{1}{c}{\emph{EnglishUS}}{\textit{600 samples}}\\

\textbf{BERT+Linguistic} & 0 76 & - \\ 
Character 5-grams \citep{englishUScharacter}& 0 73 & no \\
LDA+word space model \citep{journals/soco/Hernandez-Castaneda17}& \textbf{0 85} & yes \\
LIWC \citep{perezrosas-mihalcea:2014:P14-2,miha}&  0 69& yes \\
Syntax+words \citep{Feng:2012:SSD:2390665.2390708} & 0 78 & no \\
Words \citep{DBLP:conf/acl/MihalceaS09}&  0 71& no \\

\multicolumn{1}{c}{\emph{Bluff}}{\textit{267 samples}}\\

\textbf{BERT} & \textbf{0 76} & - \\ 
\textit{Human performance}&  0 69& no \\

\multicolumn{1}{c}{\emph{CLiPS}}{\textit{1,298 samples}}\\ 
\textbf{BERT} &   \textbf{0 80} &  - \\
Unigrams \citep{L14-1001} & 0 72 & yes  \\

\multicolumn{1}{c}{\emph{EnglishIndia}}{\textit{600 samples}}\\

\\ 
\textbf{BERT+Linguistic} & \textbf{0 70} & - \\ 
LIWC \citep{perezrosas-mihalcea:2014:P14-2,miha}&  0 66& yes \\
 \multicolumn{1}{c}{\emph{Russian}}{\textit{226 samples}}\\ 
\textbf{Word n-gram} & 0 64& -\\
POS tags+POS tags bigrams features \citep{pisarevskaya-etal-2017-deception}& 0 57 &  no\\
Rocchio classification \citep{LitvinovaSLL17}& \textbf{0 68} &  no\\

 \multicolumn{1}{c}{\emph{SpanishMexico}}{\textit{346 samples}}\\ 
\textbf{Phoneme-gram} & \textbf{0 74} & - \\ 
LIWC \citep{perezrosas-mihalcea:2014:P14-2,miha}&  0 68& no \\

\multicolumn{1}{c}{\emph{Romanian}}{\textit{870 samples}}\\ 
\textbf{BERT+Linguistic} & \textbf{0 71} & - \\ 
LIWC \citep{miha}&  0 64& yes \\
\hline  
    \end{tabular}
  \end{minipage}}
  \label{comparisonresults}

\end{table}

\section{Conclusions}
\label{sec:conc}
This study explores the task of automated text-based
deception detection within cultures 
by taking into consideration cultural and language factors, as well as limitations in NLP tools and resources for the examined cases. Our aim is to add a larger scale computational approach in a series of recent inter-disciplinary works
that examine the connection between culture and deceptive language. 
Culture is a factor that is usually ignored in automatic deception detection approaches,
which simplistically assume the same deception patterns across cultures. 
To this end,  we experimented with datasets representing  six cultures, using countries as culture proxies (United States, Belgium, India, Russia, Mexico and Romania),  written in five languages (English, Dutch, Russian, Spanish and Romanian). The datasets cover diverse genres, ranging from reviews of products and services, to opinions in the form of short essays, and even transcripts from a radio game show. To the best of our knowledge, this is the first effort to examine in parallel and in a computational manner, multiple and diverse cultures for the highly demanding deception detection task in text.

We aimed at exploring to what extent conclusions drawn from the social psychology field about the connection of deception and culture can be confirmed in our study. The basic notion demonstrated by these studies is that specific linguistic cues to deception do not appear consistently across all cultures, e.g., they change direction, or are milder or stronger between truthful or deceptive texts. Our main focus was to investigate if these differences can be attributed to cultural norm differences and especially to the individualism/collectivism divide. The most closely related work is that of Taylor \citep{1890, Taylor2017CultureMC} from the field of social psychology, that studies the above considerations for four linguistic cues of deception namely negations, positive affect, pronouns usage and spatial details in texts from individualistic and collectivist cultures. Having as starting point Taylor's work, we performed a study with similar objectives over a larger feature set that we created, that also covers the previously mentioned  ones. 

The outcome of our statistical analysis demonstrates that indeed there are great differences in the usage of pronouns  between different cultural groups. In accordance with Taylor's work, people from individualistic cultures employ more third person  and less first person pronouns to distance themselves from the deceit when they are deceptive, whereas in the collectivism group this trend is milder, signalling the effort of the deceiver to distance the group from the deceit. 
Regarding the expression of sentiment in deceptive language across cultures, the original work of Taylor hypothesized that different cultures will use sentiment differently while deceiving,
a hypothesis that was not supported by the results of his research. The basis for this hypothesis is the observation that in high-context languages, which are related with collectivist cultures,  people tend to restrain their sentiment. Our experiments support the original hypothesis of Taylor, since we observe an increased usage of positive language in deceptive texts for individualistic cultures (mostly in the US datasets), which is not observed in more collectivist cultures.
In fact, by examining the statistical significant features and the resulting feature sets from the MLR analysis, we notice that generally, there are fewer  discriminating deception cues in the high-context cultures. This might be attributed to the fact that the bibliography overwhelmingly focuses on individualistic cultures, and to a lesser degree in collectivist cultures, leading to a smaller variation in deceptive cues for the latter. Additionally, it might indicate that during deception, high-context cultures use other communication channels on top of the verbal ones, a hypothesis that needs further research.
Moreover, in affirmation of the above considerations, we observed that the strongly distinguishing features are different for each culture. The most characteristic examples are the \emph{\#negations} for the EnglishIndia dataset and the phoneme-related features for the SpanishMexico and Romanian datasets (\emph{\#nasals} and \emph{\#fricatives}). Both types of features have been related to the implicit expression of sentiment in previous studies. However, there is a need for a more thorough analysis, in order for such observations to be understood and generalized in other cultures. 
In relation to spatial details differences, we found that in the cross-cultural deception task, the collectivist groups increased the spatial details vocabulary. The exact opposite holds for the individualist groups, who used more spatial details while being truthful. This result is in accordance with Taylor's work.

These findings can be analyzed in conjunction with our second research goal which was to investigate the existence of a universal feature set that is reliable enough to provide a satisfactory performance across cultures and languages. Our analysis showed the absence of such a feature set. On top of this, our experiments inside the same culture (United States of America) and over different genres, revealed how volatile and sensitive the deception cues are. The more characteristic example is the Bluff dataset in which deception and humour are employed at the same time and the examined linguistic features have the reversed direction. Furthermore, another variable in the examined datasets is the type of deception. The examined datasets contain multiple types such as falsifications, oppositions and exaggerations to name a few. In addition, the data collection extraction process  varies from user-generated content (e.g., posts in TripAdvisor, Amazon reviews), crowd-sourced workers, volunteers in controlled environments, and finally cases outside  computer-mediated communication (the transcriptions from the Bluff the Listener show). 
Despite this diversity, we have to note that some features seem to have a broader impact. 
This is the case for the length of texts (\emph{\#lemmas} and \emph{\#words} features), where  deceptive texts tend to be shorter. This was observed independently of the culture and the domain with only one exception, that of the Bluff dataset. This is in accordance with previous studies, attributing this behaviour to the reduction of  cognitive/memory load \citep{10.3389/fpsyg.2015.01965} during the deception act.  

Our third goal was to work towards the creation of culture/language-aware classifiers. We experimented with varying approaches and examined if we 
can employ specific models and approaches in a uniform manner across cultures and languages. We explored two classification schemes; logistic regression and fine-tuning BERT. 
Moreover, the experimentation with the logistic regression classifiers demonstrated the superiority of word and phoneme n-grams over all the others n-gram variations (character, POS and syntactic). Our findings show that the linguistic cues, even when combined  with n-grams, lag behind the single or combined n-gram features, whenever models are trained for a specific domain and language (although their performance surpasses the baselines). In more details, shallow features, like the various n-grams approaches, seem to be pretty important for capturing the domain of a dataset, while the linguistic features perform worse. This is the case at least for the native English datasets, where we conducted experiments over various genres and found that the shallow features perform better, even across-domains. On the other hand, the linguistic cues seem to be important for the collectivist cultures, especially when combined with swallow features (e.g., in Russian, SpanishMexico and Romanian datasets).
The fine tuning of the BERT models, although costly in terms of tuning the hyperparameters,
performed rather well. Particularly, in some datasets (the NativeEnglish, CLiPS, and EnglishIndia datasets) we report state-of-the-art performance. However, the most important conclusion is that the combination of BERT with linguistic markers of deception is beneficial, since it enhances the performance. This is probably due to the addition of linguistic information that BERT is unable to infer, such as phoneme related information. Indeed, phonemes play an important role in all individual parts of this study.
The experimentation with the multilingual embeddings of multiBERT, as a case of zero-shot transfer learning, showed promising results that can possibly be improved by incorporating culture specific knowledge or by taking advantage of cultural and language similarities for the least resourced languages. Finally, we observed the importance of domain specific deception cues across languages, which can be identified by multiBERT. Given the promising results of multiBERT, other recently introduced multilingual representations may be applied. Alternatives include, for example, MUSE \citep{chidambaram-etal-2019-learning,yang-etal-2020-multilingual}, LASER \citep{Artetxe2019LASER}, LaBSE \citep{feng2020languageagnostic}. XLM \citep{Lample2019XLM} and its XLM-R extension \citep{conneau-etal-2020-unsupervised} have been reported to obtain state-of-the-art performance in zero-shot cross-lingual transfer scenarios, making them appropriate for low resource languages \citep{Hu2020XTREME}.

Although this work is focused on deception detection from text using style-based features and without being concerned with a particular domain, we plan to consider additional features that have been used in other domains and other related work. Specifically, we aim to incorporate features used in discourse level analysis, such as rhetorical relationships \citep{rubin,Karimi2019discourse,Pisarevskaya:2019:AL:3308560.3316604}, other properties of deception like acceptability, believability, the reception \citep{Jankowski2018ResearchingFN} of a deceptive piece of text (e.g., number of likes or dislikes), and/or source-based features such as the credibility of the medium or author using stylometric approaches \citep{Potthast2018hyperpartisan,Baly2018Fakenews}. Such features are used extensively in fake news detection \citep{Zhou2020surveyfakenews}. We also plan to examine the correlation of such features with the perceiver’s culture \citep{seiter2002,Mealy2007acceptabilityecuador}.

We also plan to study deception detection under the 
prism of culture over other languages and cultures, e.g., Portuguese 
\citep{portuguese}, German \citep{10.1007/978-3-030-30760-8_25}, Arabic\footnote{\url{https://www.autoritas.net/APDA/task-description/}}, Italian \citep{DBLP:conf/lrec/FornaciariP12,capuozzo-etal-2020-decop}. We also are interested in exploring different contexts, e.g., fake news \citep{DBLP:journals/corr/abs-1708-07104}, 
modalities, e.g., spoken dialogues, as well as employing other state-of-the-art deep learning approaches, e.g., XLNet \citep{yang2019xlnet}, RoBERTa \citep{roberta}, and DistilBERT \citep{sanh2019distilbert}.

Additionally, we plan to extend the Bluff the Listener dataset with new episodes of this game show, in order  to further examine the linguistic cues of deception and humour and how they correlate,
and to enrich the community with relevant gold datasets for non-studied languages, e.g., Greek.
Moreover, we  plan to investigate the role of phonemes and its relation with the expression of sentiment, and incorporate and study phonemes embeddings  \citep{haque2019audiolinguistic}.
Finally, we will  apply and evaluate our models in real-life applications.
This will hopefully add more evidence to the generality of our conclusions, and eventually lead to further performance improvements and reliable practical applications.

\bibliographystyle{unsrtnat}
\bibliography{preprintCulture_Papantoniou}
\label{lastpage}

\appendix
\section{Mann-Whitney U test}
{\fontsize{9}{9}\selectfont
\begin{landscape}
\begin{longtable}[!ht!]{lrrrrrrr}
\caption{\textit{p}-value for linguistic cues for the native English datasets based on Mann-Whitney U test. The numbers in brackets denote the mean for truthful and the mean for deceptive texts respectively. With bold font \textit{p}-values \textless 0.01.}\\
\tiny  &&& &    &    \\  \hline
Linguistic cue       & OpSpam & Boulder & DeRev &  EnglishUS &Bluff   \\
\hline
avg. word length& 0 093	[4.533	4.509]&0 026	[4.43	4.403]&\bm{$\sim 0$}	[4.582	4.796]&0 674	[4.437	4.426]&0 034	[4.784	4.866]\\ 
  \#adj.  and  \#adv.& 0 337	[0.167	0.165]&0 122	[0.171	0.176]&\bm{$\sim 0$}	[0.164	0.141]&0 466	[0.155	0.153]&0 506	[0.126	0.125]\\ 
  \#articles& \textbf{0 006}	[0.12	0.117]&\textbf{0 001}	[0.121	0.116]&\bm{$\sim 0$}	[0.127	0.113]&0 923	[0.1	0.1]&0 302	[0.106	0.103]\\ 
  \#boosters& \bm{$\sim 0$}	[0.005	0.006]&0 094	[0.005	0.006]&0 087	[0.007	0.009]&0 27	[0.008	0.01]&0 184	[0.007	0.005]\\ 
  \#filled pauses& 0 965	[0.001	0]&0 97	[0.002	0.001]&0 91	[0.008	0]&$\sim 1$&$\sim 1$\\ 
  \#function words& \bm{$\sim 0$}	[0.279	0.29]&0 07	[0.293	0.296]&\bm{$\sim 0$}	[0.294	0.244]&0 021	[0.318	0.329]&\textbf{0 001}	[0.243	0.227]\\ 
  \#hedges& \bm{$\sim 0$}	[0.013	0.016]&0 329	[0.012	0.014]&0 027	[0.013	0.008]&0 729	[0.015	0.016]&0 512	[0.012	0.012]\\ 
  \#lemmas& 0 04	[89.237	84.71]&\bm{$\sim 0$}	[70.015	59.974]&\textbf{0 001}	[73.593	76.983]&\bm{$\sim 0$}	[51.163	42.283]&\bm{$\sim 0$}	[107.393	125.129]\\ 
  \#negations& 0 065	[0.017	0.016]&0 031	[0.017	0.02]&0 202	[0.015	0.011]&0 029	[0.023	0.029]&0 075	[0.013	0.011]\\ 
  \#prepositions& 0 43	[0.106	0.106]&0 502	[0.099	0.097]&\textbf{0 008}	[0.107	0.12]&\bm{$\sim 0$}	[0.111	0.093]&0 982	[0.111	0.111]\\ 
  \#punctuation marks& \bm{$\sim 0$}	[2.45	1.964]&\bm{$\sim 0$}	[2.095	1.92]&0 023	[2.597	2.817]&\bm{$\sim 0$}	[1.829	1.568]&0 735	[3.503	3.505]\\ 
  \#vague words& 0 564	[0.481	0.46]&0 593	[0.383	0.359]&0 475	[0.525	0.542]&0 658	[0.207	0.18]&0 285	[0.382	0.5]\\ 
  \#verbs& \bm{$\sim 0$}	[0.173	0.184]&0 014	[0.183	0.187]&0 092	[0.179	0.171]&\textbf{0 002}	[0.206	0.216]&0 412	[0.185	0.181]\\ 
  \#words& 0 294	[150.109	144.7]&\bm{$\sim 0$}	[114.206	93.446]&\textbf{0 003}	[126.11	122.051]&\bm{$\sim 0$}	[79.177	62.263]&\bm{$\sim 0$}	[169.393	201.331]\\ 
 \midrule \#fricatives& \textbf{0 001}	[0.135	0.138]&0 108	[0.139	0.138]&0 043	[0.134	0.139]&0 884	[0.144	0.145]&0 139	[0.134	0.132]\\ 
  \#nasals& \textbf{0 001}	[0.081	0.083]&0 313	[0.082	0.083]&0 204	[0.083	0.085]&0 369	[0.089	0.091]&0 468	[0.086	0.085]\\ 
  \#plosives& 0 165	[0.104	0.105]&0 621	[0.111	0.112]&0 958	[0.103	0.103]&0 876	[0.097	0.097]&0 84	[0.108	0.107]\\ 
 \midrule \#pronouns& \bm{$\sim 0$}	[0.066	0.076]&0 024	[0.076	0.08]&\textbf{0 002}	[0.077	0.057]&0 977	[0.093	0.095]&0 912	[0.057	0.057]\\ 
  \#1st person pron.& \bm{$\sim 0$}	[0.04	0.051]&0 076	[0.04	0.043]&\bm{$\sim 0$}	[0.037	0.017]&\bm{$\sim 0$}	[0.049	0.03]&0 024	[0.009	0.013]\\ 
  \#3rd person pron.& 0 427	[0.02	0.02]&0 038	[0.029	0.027]&0 055	[0.03	0.022]&\bm{$\sim 0$}	[0.036	0.052]&0 068	[0.03	0.034]\\ 
  \#1st person pron. (singular)& \bm{$\sim 0$}	[0.023	0.037]&0 055	[0.032	0.035]&\bm{$\sim 0$}	[0.035	0.013]&\bm{$\sim 0$}	[0.03	0.018]&0 055	[0.006	0.009]\\ 
  \#1st person pron. (plural)& \bm{$\sim 0$}	[0.017	0.014]&0 28	[0.008	0.007]&0 149	[0.002	0.004]&\textbf{0 004}	[0.02	0.012]&0 402	[0.003	0.004]\\ 
  \#demonstrative pron.& 0 484	[0.016	0.017]&0 02	[0.022	0.024]&0 331	[0.033	0.029]&0 612	[0.024	0.024]&\textbf{0 01}	[0.019	0.015]\\ 
  \#indefinite pron.& 0 07	[0.019	0.02]&0 577	[0.022	0.022]&0 458	[0.022	0.022]&\textbf{0 001}	[0.035	0.027]&0 672	[0.019	0.018]\\ 
 \midrule pos. SentiWordNet& 0 389	[0.382	0.374]&\textbf{0 005}	[0.392	0.42]&0 544	[0.384	0.359]&0 763	[0.401	0.408]&0 251	[0.316	0.306]\\ 
  neg. SentiWordNet& 0 172	[0.256	0.252]&0 745	[0.276	0.281]&0 356	[0.229	0.217]&0 045	[0.3	0.32]&0 374	[0.249	0.244]\\ 
  pos. MPQA& \textbf{0 001}	[0.08	0.088]&\bm{$\sim 0$}	[0.075	0.087]&0 048	[0.087	0.095]&0 283	[0.084	0.087]&0 27	[0.058	0.056]\\ 
  neg. MPQA& 0 032	[0.032	0.03]&0 442	[0.036	0.038]&0 937	[0.039	0.04]&\textbf{0 008}	[0.067	0.079]&0 222	[0.048	0.042]\\ 
  pos. FBS& 0 188	[0.06	0.063]&0 038	[0.057	0.063]&0 856	[0.058	0.056]&0 495	[0.04	0.045]&0 071	[0.029	0.032]\\ 
  neg. FBS& 0 716	[0.021	0.022]&0 688	[0.026	0.027]&0 669	[0.027	0.028]&0 054	[0.053	0.062]&0 023	[0.034	0.028]\\ 
  sentiment-ANEW& 0 056	[0.045	0.048]&0 861	[0.04	0.04]&0 076	[0.041	0.046]&0 904	[0.021	0.021]&\bm{$\sim 0$}	[19.606	23.958]\\ 
 \midrule mean sentence length& \bm{$\sim 0$}	[17.474	17.686]&\textbf{0 003}	[17.569	16.596]&\bm{$\sim 0$}	[19.186	22.119]&\bm{$\sim 0$}	[17.59	15.1]&0 799	[21.124	21.039]\\ 
  mean preverb length& \bm{$\sim 0$}	[5.077	5.652]&0 02	[5.586	5.281]&\bm{$\sim 0$}	[5.762	7.569]&0 374	[5.25	5.029]&0 277	[7.195	7.772]\\ 
  \#conjunction words& 0 073	[0.045	0.043]&0 902	[0.042	0.042]&0 153	[0.043	0.046]&0 164	[0.036	0.034]&0 06	[0.03	0.034]\\ 
  \#subordinate clauses& \bm{$\sim 0$}	[0.47	0.558]&0 952	[0.553	0.548]&0 012	[0.542	0.675]&\bm{$\sim 0$}	[0.742	0.606]&0 918	[0.75	0.758]\\ 
 \midrule \#exclusion words& \bm{$\sim 0$}	[0.007	0.006]&0 386	[0.006	0.006]&0 126	[0.007	0.004]&0 017	[0.005	0.003]&0 51	[0.005	0.005]\\ 
  \#modal verbs& \bm{$\sim 0$}	[0.015	0.018]&\textbf{0 002}	[0.016	0.021]&\textbf{0 001}	[0.021	0.025]&0 429	[0.044	0.048]&0 03	[0.021	0.015]\\ 
  \#motion verbs& 0 113	[0.103	0.097]&0 152	[0.088	0.083]&0 805	[0.209	0.189]&0 153	[0.047	0.039]&0 234	[0.088	0.102]\\ 
  \#spatial words& \bm{$\sim 0$}	[0.194	0.178]&0 878	[0.165	0.166]&0 287	[0.205	0.201]&0 867	[0.176	0.174]&0 483	[0.212	0.207]\\ 
  \#verbs in future tense& 0 745	[0.034	0.035]&0 673	[0.036	0.039]&\textbf{0 004}	[0.037	0.048]&0 744	[0.041	0.045]&0 709	[0.03	0.031]\\ 
  \#verbs in past tense& 0 077	[0.508	0.515]&\bm{$\sim 0$}	[0.403	0.355]&\bm{$\sim 0$}	[0.354	0.172]&\bm{$\sim 0$}	[0.169	0.127]&\bm{$\sim 0$}	[0.321	0.408]\\ 
  \#verbs in present tense& 0 1	[0.452	0.446]&\textbf{0 002}	[0.556	0.597]&\bm{$\sim 0$}	[0.603	0.773]&\textbf{0 001}	[0.786	0.826]&\bm{$\sim 0$}	[0.643	0.556]\\ 
 \hline
 \label{u_test_us}
\end{longtable}
\end{landscape}
}

{\fontsize{8}{8}\selectfont
\begin{landscape}
\begin{longtable}[ht!]{lrrrrrr}
\caption{\textit{p}-value for linguistic cues for datasets from individualistic cultures based on Mann-Whitney U test. The numbers in brackets denote the first number the mean for truthful and the second number the mean for deceptive texts. With bold font \textit{p}-values \textless 0.01. With N/A are marked those features that are not applicable for the specific language.} \\
Linguistic cue &  NativeEnglish   &    Dutch        & India & Russian & Mexico   & Romania \\
\hline
avg. word length& 0 052	[4.505	4.491]&\bm{$\sim 0$}	[8.577	8.694]&0 373	[4.466	4.445]&0 525	[4.339	4.3]&0 486	[4.149	4.134]&0 613	[4.151	4.167]\\ 
 \#adj.  and  \#adv.& 0 824	[0.164	0.164]&0 047	[0.339	0.347]&0 884	[0.143	0.145]&0 391	[0.052	0.047]&0 046	[0.126	0.139]&0 634	[0.145	0.146]\\ 
 \#articles& \bm{$\sim 0$}	[0.117	0.113]&0 022	[0.231	0.237]&0 823	[0.095	0.096]&N/A&0 839	[0.102	0.104]&0 181	[0.035	0.037]\\ 
 \#boosters& \textbf{0 002}	[0.006	0.007]&-&0 409	[0.011	0.013]&-&-&-\\ 
 \#filled pauses& 0 943	[0.002	0]&-&$\sim 1$&-&-&-\\ 
 \#function words& 0 044	[0.288	0.291]&0 815	[0.402	0.403]&0 1	[0.306	0.313]&0 62	[0.181	0.175]&0 339	[0.202	0.206]&0 339	[0.138	0.141]\\ 
 \#hedges& 0 032	[0.013	0.015]&-&0 27	[0.013	0.014]&-&-&-\\ 
 \#lemmas& \bm{$\sim 0$}	[77.553	71.217]&\textbf{0 01}	[102.835	98.21]&\textbf{0 001}	[48.613	43.307]&0 37	[105.77	109.15]&\bm{$\sim 0$}	[66.523	48.649]&\bm{$\sim 0$}	[62.301	49.874]\\ 
 \#negations& 0 985	[0.018	0.018]&0 04	[0.016	0.017]&\bm{$\sim 0$}	[0.02	0.026]&0 707	[0.018	0.019]&0 538	[0.021	0.022]&0 494	[0.024	0.026]\\ 
 \#prepositions& \textbf{0 002}	[0.105	0.102]&0 086	[0.165	0.16]&0 407	[0.108	0.106]&0 761	[0.201	0.198]&\textbf{0 007}	[0.124	0.112]&0 279	[0.086	0.084]\\ 
 \#punctuation marks& \bm{$\sim 0$}	[2.312	2.047]&\bm{$\sim 0$}	[2.664	2.421]&0 012	[1.532	1.432]&0 341	[2.384	2.387]&\bm{$\sim 0$}	[2.719	2.181]&\bm{$\sim 0$}	[2.267	1.992]\\ 
 \#vague words& 0 582	[0.407	0.389]&-&0 349	[0.24	0.157]&-&-&-\\ 
 \#verbs& \bm{$\sim 0$}	[0.182	0.188]&\textbf{0 003}	[0.281	0.291]&0 145	[0.199	0.205]&0 222	[0.173	0.17]&0 052	[0.189	0.196]&0 083	[0.173	0.177]\\ 
 \#words& \bm{$\sim 0$}	[127.933	115.155]&\bm{$\sim 0$}	[166.598	153.798]&\textbf{0 002}	[74.757	65.46]&0 339	[192.354	196.867]&\bm{$\sim 0$}	[106.43	72.328]&\bm{$\sim 0$}	[100.223	74.97]\\ 
\midrule \#fricatives& 0 574	[0.138	0.138]&0 214	[0.056	0.056]&0 094	[0.144	0.148]&0 565	[0.104	0.105]&0 973	[0.149	0.149]&\textbf{0 002}	[0.087	0.093]\\ 
 \#nasals& 0 012	[0.083	0.084]&0 081	[0.059	0.058]&0 491	[0.097	0.095]&0 683	[0.094	0.093]&\bm{$\sim 0$}	[0.111	0.104]&\bm{$\sim 0$}	[0.097	0.09]\\ 
 \#plosives& \bm{$\sim 0$}	[0.105	0.107]&0 083	[0.086	0.085]&0 839	[0.097	0.097]&0 181	[0.164	0.167]&0 693	[0.139	0.14]&0 575	[0.196	0.194]\\ 
\midrule \#pronouns& \bm{$\sim 0$}	[0.074	0.078]&0 2	[0.171	0.166]&0 839	[0.087	0.093]&0 575	[0.153	0.147]&0 624	[0.1	0.102]&0 198	[0.075	0.072]\\ 
 \#1st person pron.& 0 256	[0.04	0.04]&0 099	[0.044	0.041]&0 072	[0.043	0.038]&0 416	[0.057	0.054]&0 351	[0.06	0.057]&\bm{$\sim 0$}	[0.015	0.01]\\ 
 \#3rd person pron.& 0 076	[0.026	0.028]&0 067	[0.098	0.093]&\textbf{0 002}	[0.043	0.056]&0 367	[0.015	0.013]&0 192	[0.036	0.041]&\textbf{0 004}	[0.013	0.017]\\ 
 \#1st person pron. (singular)& \bm{$\sim 0$}	[0.027	0.031]&0 021	[0.043	0.039]&0 067	[0.03	0.028]&0 863	[0.027	0.028]&\bm{$\sim 0$}	[0.013	0.008]&\textbf{0 001}	[0.006	0.004]\\ 
 \#1st person pron. (plural)& \bm{$\sim 0$}	[0.013	0.009]&0 021	[0.002	0.004]&0 087	[0.013	0.009]&0 834	[0.012	0.011]&\textbf{0 001}	[0.012	0.007]&\bm{$\sim 0$}	[0.009	0.005]\\ 
 \#demonstrative pron.& 0 058	[0.02	0.021]&0 358	[0.01	0.009]&0 246	[0.019	0.018]&0 196	[0.006	0.006]&0 212	[0.003	0.003]&0 994	[0.007	0.007]\\ 
 \#indefinite pron.& 0 285	[0.023	0.022]&0 215	[0.007	0.008]&0 919	[0.033	0.034]&0 772	[0.001	0.001]&0 45	[0.011	0.012]&\textbf{0 004}	[0.011	0.009]\\ 
\midrule pos. SentiWordNet& 0 195	[0.385	0.393]&-&0 087	[0.418	0.451]&-&-&-\\ 
 neg. SentiWordNet& 0 558	[0.267	0.271]&-&0 875	[0.313	0.316]&-&-&-\\ 
 pos. MPQA& \bm{$\sim 0$}	[0.079	0.086]&-&0 968	[0.093	0.093]&-&-&-\\ 
 neg. MPQA& 0 791	[0.04	0.041]&-&0 571	[0.084	0.084]&-&-&-\\ 
 pos. FBS& \textbf{0 006}	[0.054	0.059]&-&0 172	[0.046	0.05]&-&-&-\\ 
 neg. FBS& 0 783	[0.029	0.03]&-&0 708	[0.066	0.067]&-&-&-\\ 
 sentiment-ANEW& 0 318	[0.039	0.04]&-&0 71	[0.021	0.024]&-&-&-\\ 
  pos. sentiment& -&0 035	[0.109	0.114]&-&0 744	[0.055	0.053]&0 141	[0.028	0.038]&0 063	[0.076	0.072]\\ 
 neg. sentiment& -&0 699	[0.118	0.116]&-&0 806	[0.02	0.019]&0 167	[0.039	0.036]&0 439	[0.09	0.094]\\

\midrule mean sentence length& 0 598	[17.814	17.343]&\bm{$\sim 0$}	[19.533	18.31]&0 205	[16.177	15.24]&0 64	[13.956	13.23]&\bm{$\sim 0$}	[39.134	29.471]&\bm{$\sim 0$}	[22.782	20.048]\\ 
 mean preverb length& \bm{$\sim 0$}	[5.394	5.655]&-&0 047	[6.001	5.299]&-&-&-\\ 
 \#conjunction words& 0 325	[0.042	0.041]&0 329	[0.11	0.108]&0 158	[0.032	0.029]&0 254	[0.066	0.063]&\bm{$\sim 0$}	[0.077	0.066]&\textbf{0 001}	[0.05	0.045]\\ 
 \#subordinate clauses& \textbf{0 001}	[0.556	0.579]&-&0 453	[0.476	0.506]&-&-&-\\ 
\midrule \#exclusion words& \bm{$\sim 0$}	[0.006	0.005]&-&0 602	[0.004	0.005]&-&-&-\\ 
 \#modal verbs& \bm{$\sim 0$}	[0.021	0.023]&-&0 2	[0.051	0.056]&-&-&-\\ 
 \#motion verbs& 0 041	[0.096	0.088]&0 212	[0.042	0.046]&0 831	[0.038	0.037]&0 46	[0.05	0.04]&-&-\\ 
 \#spatial words& \bm{$\sim 0$}	[0.185	0.175]&\textbf{0 001}	[0.102	0.109]&0 227	[0.21	0.202]&0 28	[0.094	0.088]&\textbf{0 001}	[0.06	0.071]&\bm{$\sim 0$}	[0.228	0.271]\\ 
 \#verbs in future tense& 0 26	[0.035	0.038]&-&0 626	[0.054	0.06]&0 814	[0.203	0.202]&0 653	[0.032	0.038]&-\\ 
 \#verbs in past tense& \bm{$\sim 0$}	[0.404	0.374]&0 02	[0.27	0.293]&0 691	[0.161	0.157]&0 746	[0.667	0.671]&\bm{$\sim 0$}	[0.126	0.072]&0 789	[0.107	0.115]\\ 
 \#verbs in present tense& \textbf{0 001}	[0.555	0.582]&\bm{$\sim 0$}	[0.552	0.517]&0 556	[0.781	0.777]&0 452	[0.149	0.142]&\bm{$\sim 0$}	[0.841	0.89]&0 021	[0.711	0.734]\\

 \hline
 \label{u_test_culture}
\end{longtable}
\end{landscape}
}

\begin{table}[ht!]
  \caption{Results per culture of logistic regression experiments on various feature types, including combinations of pairs. The accuracy measure is reported and the bold font marks the pair with the best achieved performance. Best n-gram row indicates the best accuracy for the no-paired configuration.}
 {\begin{minipage}{25pc}
    \begin{tabular}{@{\extracolsep{\fill}}lllllll}
    \hline
&\textbf{US}&\textbf{Dutch}&\textbf{India}&\textbf{Russia}&\textbf{Mexico} &\textbf{Romania}\\
\\ \hline
\multicolumn{7}{c}{\emph{Single feature types}}\\\hline
Linguistic &	0 62	& 0 60	&	0 54	& 0 50	& 0 60 	& 0 64 \\
Phoneme &	0 65	&	0 73	&	0 60	&	0 64	&	\textbf{0 74}	&	0 66	\\
Character &	0 69	&	0 73	&	0 61	&	0 50	&	0 63	&	0 62	\\
Word &	\textbf{0 72}	&	\textbf{0 78}	&	0 61	&	\textbf{0 64}	&	\textbf{0 74}	&	0 65	\\
POS &	0 64	&	0 50	&	0 60	&	0 61	&	0 63	&	0 64	\\
SN &	0 67	&	-	&	0 58	&	-	& - 	&	-	\\\hline
\multicolumn{7}{c}{\emph{Pairs of n-grams}}\\\hline
Phoneme+Word	&	0 70	&	\textbf{0 78}	&	\textbf{0 66}	&	0 55	&	\textbf{0 74}	&	0 63	\\
Phoneme+POS	&	0 68	&	0 76	&	0 64	&	0 46	&	\textbf{0 74}	&	\textbf{0 70}	\\
Phoneme+SN	&	0 68	&	-	&	0 60	&	-	&	-	&	-	\\
Character+Phoneme	&	0 66	&	0 75	&	0 62	&	0 60	&	0 55	&	0 65	\\
Character+Word	&	\textbf{0 72}	&	\textbf{0 78}	&	0 63	&	0 57	&	0 62	&	0 63	\\
Character+POS	&	0 67	&	0 73	&	0 61	&	0 46	&	0 58	&	0 67	\\
Character+SN 	&	0 69	&	-	&	0 63	&	-	&	-	&	-	\\
Word+POS	&	\textbf{0 72}	&	\textbf{0 78}	&	0 64	&	0 60	&	\textbf{0 74}	&	0 63	\\
Word+SN 	&	\textbf{0 72}	&	-	&	0 63	&	-	&	-	&	-	\\
POS+SN	&	0 68	&	-	&	0 51	&	-	&	-	&	-	\\ \hline
\multicolumn{7}{c}{\emph{Best linguistic + n-gram model}}\\\hline
Linguistic+ Best n-gram &	\textbf{0 72}	& 0 76	&	0 56	&	0 57	& 0 63	&	0 68	\\ \hline

\hline
    \end{tabular}
  \end{minipage}}
  \label{ngrampairs}
\end{table}
\end{document}